\documentclass[10pt,journal,compsoc]{IEEEtran}

\usepackage{amsmath,amssymb,amsfonts}

\usepackage{amsthm}

\usepackage{mathtools}

\usepackage{symbolDef}
\let\trace\undefined

\DeclareMathOperator{\trace}{trace}

\newcommand{\sctbw}{\widetilde{\bar{w}}}

\newcommand{\schts}{\widehat{\scts}}

\usepackage{algorithm}
\usepackage{algpseudocode}

\usepackage{xcolor}

\theoremstyle{plain}
\newtheorem{theorem}{Theorem}
\newtheorem{lemma}{Lemma}
\newtheorem{proposition}{Proposition}
\newtheorem{corollary}{Corollary}
\newtheorem{assumption}{Assumption}

\DeclareMathOperator{\snr}{SNR}
\DeclareMathOperator{\cond}{cond}
\DeclareMathOperator{\mmse}{MMSE}

\theoremstyle{definition}
\newtheorem{definition}{Definition}

\theoremstyle{remark}

\usepackage{booktabs}
\usepackage{multirow}

\usepackage{float} %
\usepackage{graphicx}
\usepackage{adjustbox}
\usepackage{array}
\graphicspath{{figures/}}

\usepackage[nocompress]{cite}

\ifCLASSINFOpdf
\else
\fi

\usepackage[hidelinks]{hyperref}

\begin{document}
\bstctlcite{IEEEexample:BSTcontrol}

\title{Graph Signal Generation via Graph-Aware Diffusion Models}

\author{Vimal~K~B, Sergio~Rozada, Antonio~G.~Marques, Elvin~Isufi,
        and~Hadi~J.~Rad%
\IEEEcompsocitemizethanks{%
\IEEEcompsocthanksitem V.~K~B, E.~Isufi and H.~J.~Rad are with Delft University of Technology, Delft, The Netherlands.
\IEEEcompsocthanksitem S.~Rozada and A.~G.~Marques are with King Juan Carlos University, Madrid, Spain.\protect\vspace{\baselineskip}}%
\thanks{This work was partially supported by the Spanish AEI grant PID2022-136887NB-I00 and PID2025-170000NB-I00, the Community of Madrid via the ELLIS Madrid Unit and grants TEC-2024/COM-89, TU Delft AI Labs programme, NWO OTP GraSPA proposal \#19497, NWO VENI proposal 222.032, the TU Delft - Shell.ai TKI project, and by the SURE-AI Centre grant \#357482, Research Council of Norway.}}

\markboth{IEEE Transactions on Pattern Analysis and Machine Intelligence,~Vol.~XX, No.~X, Month~2026}%
{K~B \MakeLowercase{\textit{et al.}}: Graph Signal Generation via Graph-Aware Diffusion Models}

\IEEEtitleabstractindextext{%
\begin{abstract}
Many signals are defined on the nodes of a graph, such as traffic flow at road sensors or user purchases of related products. They are often partly observed, and learning their distribution enables imputing missing values and forecasting future ones. Diffusion models learn such distributions, but their forward noise is independent across nodes and does not depend on the graph. We introduce graph-aware diffusion (GAD), whose forward process is a heat diffusion on the graph driven by Gaussian noise. In the graph Fourier basis, the forward transition is a scalar Gaussian channel at each frequency. With a shared diffusion rate, high graph frequencies are attenuated faster than low ones. We propose Mode-Adaptive SNR Scheduling (MASS), which assigns each graph frequency its own diffusion rate and keeps all of them at the same SNR. For graph-stationary signals, MASS equalizes the normalized error of the optimal linear denoiser across frequencies. We bound reverse-SDE stiffness as a function of the (shifted) Laplacian's condition number, with a logarithmic upper bound for MASS and a linear lower bound for shared rates. GAD with MASS improves recommendation accuracy by 5--19\%, reduces traffic-imputation RMSE by up to 7.8\% with half as many reverse steps, and improves forecasting MAE and RMSE using only 3 reverse steps.
\end{abstract}

}

\maketitle

\IEEEraisesectionheading{\section{Introduction}\label{sec:introduction}}

\IEEEPARstart{D}{iffusion} models learn a data distribution by gradually corrupting the data with noise and training a neural network, the denoiser, to reverse this corruption~\cite{sohldickstein2015deepunsupervisedlearningusing,ho2020denoisingdiffusionprobabilisticmodels,song2021scorebasedgenerativemodelingstochastic}. Many signals of interest live on the nodes of a graph~\cite{Shuman_2013,ortega2018graph,bronstein2017geometric,isufi2024graph,gama2020graphs,corso2024graph}, and diffusion models have recently been applied to graph-signal tasks that need plausible samples rather than a single estimate, such as spatio-temporal imputation of traffic readings~\cite{tashiro2021csdi,liu2023pristi,mao2026fence}, probabilistic forecasting~\cite{wen2024diffstg}, recommendation from a few user interactions~\cite{wang2023diffrec,lee2025cdiff4rec}, and resource allocation in wireless systems~\cite{uslu2025generative, uslu2026graph}. In these tasks, the observations alone do not determine the signal distribution, and the graph supplies part of the prior by encoding dependencies among nodes.

Most of these models use the graph only in the denoiser or in a conditioning network~\cite{tashiro2021csdi,liu2023pristi,wen2024diffstg,wang2023diffrec,lee2025cdiff4rec,uslu2026graphgeneric, uslu2026generative}. Their forward process corrupts every node independently, so their prior is an isotropic Gaussian and the denoiser must learn the interplay between the graph topology and the data, which is costly when data are scarce or the graph is large. Frequency-dependent corruption has been studied mainly for images, either by blurring, which attenuates the high frequencies before the low ones~\cite{rissanen2023generative,hoogeboom2023blurring,daras2022soft}, or by setting a different noise variance for each frequency~\cite{falck2024fourier,scimeca2024learning}. These methods rely on the Fourier basis of the regular pixel grid, which does not exist on graphs with irregular connections. On any graph, the eigenvectors of the graph Laplacian can play the role of Fourier modes and its eigenvalues act as graph frequencies~\cite{Shuman_2013}. These provide a direct basis for defining a forward process and its diffusion rates.

We propose graph-aware diffusion (GAD), whose forward process is heat diffusion on the graph driven by Gaussian noise~\cite{rozada2025graphawarediffusionsignalgeneration}. Heat diffusion smooths the signal by attenuating each graph frequency exponentially, the higher frequencies faster. A diffusion rate sets how fast this corruption proceeds. The process has closed-form Gaussian transitions and converges to a Gaussian Markov random field (GMRF) defined by the graph Laplacian~\cite{rue2005gaussian}, which we use as the prior. Since this prior already encodes the dependencies between connected nodes, the denoiser does not have to learn them from data, only how the data deviate from them, a task suited to graph filters and graph neural networks~\cite{chen2021graph}.

The noise schedule sets the signal-to-noise ratio (SNR) of the corrupted signal at each diffusion time, i.e., the power of the surviving signal relative to that of the added noise. The SNR affects how much of the clean signal the denoiser can recover~\cite{kingma2021variational,karras2022elucidating,chen2023importance}. In isotropic diffusion, every direction is attenuated equally, and the SNRs differ only through the power spectrum of the data~\cite{falck2024fourier}. In GAD, the SNR of each graph frequency depends on its clean power (set by the data), its prior power (set by the graph), and its attenuation, which decays exponentially with the graph frequency. With a shared diffusion rate, the SNRs separate exponentially in the frequency gap as the integrated rate grows. We propose Mode-Adaptive SNR Scheduling (MASS), which sets the diffusion rate of each graph frequency from its clean power and its prior power, so that all graph frequencies share the same SNR at every diffusion time.

A preliminary version of this work~\cite{rozada2025graphawarediffusionsignalgeneration} introduced the GAD forward process with a shared polynomial diffusion rate (FCPS) and a limited evaluation. This journal version develops the spectral theory of the process, introduces frequency-dependent diffusion rates with guarantees on denoising error and reverse-SDE stiffness, and substantially broadens the evaluation. Our contributions are as follows.

\begin{itemize}
\item[\textbf{C1}] \textbf{Spectral analysis of GAD}
(Sections~\ref{sec:graph_aware_diffusion} and~\ref{sec:denoising_consequence}).
GAD has a scalar Gaussian forward channel with closed-form SNR at each graph frequency. Under graph stationarity, we characterize the optimal linear denoising error and the fixed-time training rate at each graph frequency.

\item[\textbf{C2}] \textbf{Mode-Adaptive SNR Scheduling}
(Section~\ref{sec:mode_adaptive_snr}).
MASS gives each graph frequency its own diffusion rate so that all share the same SNR, without changing the GMRF prior. Under graph stationarity, the optimal linear denoiser then has the same normalized error at every graph frequency. The stiffness of the reverse SDE admits a logarithmic upper bound with MASS and a linear lower bound with a shared rate, in terms of the condition number of the shifted Laplacian.

\item[\textbf{C3}] \textbf{Empirical validation}
(Section~\ref{sec:numerical_analysis}).
On synthetic graph signals, MASS reduces the stiffness diagnostic for the diffusion rate by factors of $4.88{\times}10^{2}$ to $1.49{\times}10^{5}$ and improves sample quality at a fixed number of reverse steps. On downstream tasks, it improves recommendation metrics by $5$--$19\%$, reduces imputation RMSE by up to $7.8\%$ with half the reverse steps, and lowers forecasting MAE and RMSE with only $3$ reverse steps.\footnote{The code for our experiments is available at
\url{https://github.com/vimalkb7/SpectralAnalysisGAD}.}
\end{itemize}

\section{Context and Background}

\subsection{Related Work}
\label{sec:related_work}

We review prior work on the forward process, the noise schedule, and the sampler. Most existing diffusion models for graph signals use isotropic forward noise, and the graph is used only by the denoiser or the conditioning variables~\cite{tashiro2021csdi,liu2023pristi,wen2024diffstg,wang2023diffrec,lee2025cdiff4rec,uslu2026graphgeneric,uslu2026graph, schmidt2026stability}.

Structured forward processes either shape the noise covariance or make the drift graph-aware~\cite{park2024equivariant}. In the first case, the covariance is fixed in advance, built from the data domain, or estimated from the conditioning signal~\cite{voleti2022nonisotropic,phillips2022spectral,curreli2025nonisotropic,huang2024blue,lee2022priorgrad}, but the drift does not propagate the signal along the edges, and the terminal covariance is chosen rather than reached as the invariant law of the dynamics. In the second case, the forward process blurs images~\cite{rissanen2023generative,hoogeboom2023blurring} or filters signals over a graph~\cite{choi2023bspm,zhu2024giffcf,xia2025sdiff,yang2025topological,wyrwal2026topological}, and attenuates each frequency at a rate proportional to its eigenvalue. Under a diffusion rate shared by all frequencies, the SNRs of the high and low frequencies then separate, and the effect of this separation on training and sampling has not been analyzed for generative modeling. GAD belongs to this second group. Its GMRF prior encodes the graph, so the denoiser does not have to learn it, and MASS controls how this prior interacts with the data by setting the diffusion rate of each graph frequency from its clean power and its prior power.

Noise schedules set the SNR of each component of the signal through the shape of the schedule~\cite{nichol2021improved}, the terminal condition~\cite{lin2024common}, the eigenvalues of a graph~\cite{xia2025sdiff}, or a learned network~\cite{kingma2021variational,sahoo2024mulan}. In all cases, the SNR of a component is a by-product of these choices, and different components take very different SNRs at the same diffusion time. For linear denoising of a single component, the optimal estimator approximately inverts the attenuation at high SNR and approaches the prior mean at low SNR~\cite{hang2023minsnr,kingma2023understanding}. Reweighting the denoising error over diffusion time rescales the contribution of each time but leaves the SNR of each component at that time unchanged~\cite{hang2023minsnr}. MASS instead sets the SNR of every graph frequency directly. Table~\ref{tab:related_work_positioning} summarizes these axes.

\begin{table*}[t]
\centering
\caption{Positioning of representative forward processes. Entries for the bridge and flow-matching methods refer to their reference process.}
\label{tab:related_work_positioning}
\renewcommand{\arraystretch}{1.12}
\begin{adjustbox}{max width=\textwidth}
\begin{tabular}{lccccccc}
\toprule
& \multicolumn{5}{c}{Corruption design}
& \multicolumn{2}{c}{Induced learning problem} \\
\cmidrule(lr){2-6}\cmidrule(lr){7-8}
Method
& \shortstack{Graph in forward\\drift/mean}
& \shortstack{Frequency-dependent\\covariance}
& \shortstack{Shared\\diffusion rate}
& \shortstack{SNR set per\\frequency}
& \shortstack{Graph-aware\\terminal prior}
& \shortstack{Frequency-wise SNR\\characterized}
& \shortstack{Frequency-wise denoising error \\and training time constant} \\
\midrule
Isotropic graph-signal diffusion \cite{tashiro2021csdi,liu2023pristi,wen2024diffstg,wang2023diffrec,lee2025cdiff4rec,uslu2026graphgeneric,uslu2026graph}
& $\times$ & $\times$ & $\checkmark$ & $\times$ & $\times$ & $\times$ & $\times$ \\
\midrule
Non-isotropic DDPM \cite{voleti2022nonisotropic}
& $\times$ & $\checkmark$ & $\checkmark$ & $\times$ & $\times$ & $\times$ & $\times$ \\
Spectral diffusion processes \cite{phillips2022spectral}
& $\times$ & $\checkmark$ & $\checkmark$ & $\times$ & $\times$ & $\times$ & $\times$ \\
SkeletonDiffusion \cite{curreli2025nonisotropic}
& $\times$ & $\checkmark$ & $\checkmark$ & $\times$ & $\checkmark$ & $\times$ & $\times$ \\
Blue-noise diffusion \cite{huang2024blue}
& $\times$ & $\checkmark$ & $\checkmark$ & $\times$ & $\times$ & $\times$ & $\times$ \\
PriorGrad \cite{lee2022priorgrad}
& $\times$ & $\checkmark$ & $\checkmark$ & $\times$ & $\times$ & $\times$ & $\times$ \\
\midrule
Inverse heat dissipation \cite{rissanen2023generative}
& $\checkmark$ & $\times$ & $\checkmark$ & $\times$ & $\times$ & $\times$ & $\times$ \\
Blurring diffusion \cite{hoogeboom2023blurring}
& $\checkmark$ & $\checkmark$ & $\checkmark$ & $\times$ & $\times$ & $\checkmark$ & $\times$ \\
Topological SB and FM \cite{yang2025topological,wyrwal2026topological}
& $\checkmark$ & $\checkmark$ & $\checkmark$ & $\times$ & $\times$ & $\times$ & $\times$ \\
S-Diff \cite{xia2025sdiff}
& $\checkmark$ & $\checkmark$ & $\checkmark$ & $\times$ & $\times$ & $\checkmark$ & $\times$ \\
\midrule
MuLAN \cite{sahoo2024mulan}
& $\times$ & $\checkmark$ & $\times$ & $\times$ & $\times$ & $\checkmark$ & $\times$ \\
GAD with a shared diffusion rate \cite{rozada2025graphawarediffusionsignalgeneration}
& $\checkmark$ & $\checkmark$ & $\checkmark$ & $\times$ & $\checkmark$ & $\times$ & $\times$ \\
\textbf{GAD with MASS (ours)}
& $\checkmark$ & $\checkmark$ & $\times$ & $\checkmark$ & $\checkmark$ & $\checkmark$ & $\checkmark$ \\
\bottomrule
\end{tabular}
\end{adjustbox}
\end{table*}

Diffusion models also fit different frequency components of the data at different rates~\cite{rahaman2019spectral,wang2024analytical,kadkhodaie2024generalization}. For a spectral linear denoiser, the fixed-time training constant of a component is inversely proportional to its corrupted power. In sampling, all graph frequencies share one integration grid, and stiffness in the reverse drift can restrict the step size of explicit solvers, as for ill-conditioned scores across scales on images~\cite{guth2022wavelet}. Fast samplers reduce the number of steps without changing the forward process~\cite{song2021denoising,liu2022pseudo,lu2022dpmsolverfastodesolver,lu2022dpmpp,zhao2023unipc,salimans2022progressivedistillationfastsampling,song2023consistencymodels,zhang2023fast,zhang2023gddim}, and graph-signal models adopt them directly~\cite{wen2024diffstg,wang2023diffrec,xia2025sdiff}. MASS instead changes the forward process itself, so that all graph frequencies share the same reverse-SDE gain.

\subsection{Preliminaries}
\label{sec:foundations}

\noindent\textbf{Graph signal processing.}
Let $\graph=(\stV,\stE)$ be an undirected graph on $n=|\stV|$ nodes, with weighted adjacency matrix $\mtA\in\fdR_{\geq 0}^{n\times n}$. A graph signal $\vcx\in\fdR^{n}$ assigns the value $x_i$ to node $i$. We encode the topology through the Laplacian $\mtL=\mtD-\mtA$, with degree matrix $\mtD=\diag(\mtA\vcOnes)$. Since $\mtL$ is symmetric positive semidefinite, it admits the eigendecomposition
\begin{equation}
    \mtL=\mtV\mtLambda\mtV^{\Tr},
    \qquad
    \mtLambda=\diag(\sclambda_0=0,\ldots,\sclambda_{n-1}),
    \label{eq:laplacian_eigendecomposition}
\end{equation}
with $\mtV=[\vcv_0,\ldots,\vcv_{n-1}]$ the eigenvectors and $\mtLambda$ the respective eigenvalues. The columns of $\mtV$ form an orthonormal graph Fourier basis, with graph Fourier transform (GFT) and inverse GFT $\vctx=\mtV^{\Tr}\vcx$ and $\vcx=\mtV\vctx$, respectively \cite{ortega2018graph}. We write $\sctx_k=\vcv_k^{\Tr}\vcx$ for the $k$th GFT coefficient of $\vcx$. For Laplacians, it holds that $\vcx^{\Tr}\mtL\vcx=\sum_{(i,j)\in\stE}A_{i,j}(x_i-x_j)^2$ and hence $\sclambda_k=\vcv_k^{\Tr}\mtL\vcv_k=\sum_{(i,j)\in\stE}A_{i,j}([\vcv_k]_i-[\vcv_k]_j)^2$. That is, eigenvectors associated with small eigenvalues vary slowly across connected nodes, and their eigenvalues are the \emph{low graph frequencies}, while eigenvectors associated with large eigenvalues oscillate between neighbors, and their eigenvalues are the \emph{high graph frequencies}~\cite{Shuman_2013}. In turn, the $k$th GFT coefficient $\sctx_k$ captures the signal contribution along the $k$th graph frequency $\sclambda_k$.

\noindent\textbf{Graph filters.}
A \emph{graph convolutional filter} of order $K$ is a matrix polynomial of $\mtL$~\cite{isufi2024graph},
\begin{equation}
\label{eq:graph_filter}
\vcy
=
h_{\vctheta}(\mtL)\vcx,
\qquad
h_{\vctheta}(\mtL)=\sum_{m=0}^{K}\theta_m \mtL^m,
\end{equation}
with coefficients $\vctheta=[\theta_0,\ldots,\theta_K]^{\Tr}$. Since $\mtL^m$ couples $m$-hop neighborhoods, \eqref{eq:graph_filter} propagates the input signal $\vcx$ over the graph with $\theta_m$ weighting each neighborhood. As $\mtL^m=\mtV\mtLambda^m\mtV^{\Tr}$, the filter acts pointwise on graph frequencies,
\begin{equation}
\label{eq:filter_frequency_response}
\scty_k
=
h_{\vctheta}(\sclambda_k)\,\sctx_k,
\qquad
h_{\vctheta}(\sclambda)=\sum_{m=0}^{K}\theta_m\sclambda^m ,
\end{equation}
where $\scty_k$ is the $k$th GFT coefficient of the output $\vcy$ and $h_{\vctheta}(\sclambda)$ is the filter frequency response. %
More generally, given a scalar function $h:\fdR\to\fdR$, a \emph{spectral graph filter} is the function of the Laplacian,
\begin{equation}
    h(\mtL)
    =
    \mtV\diag\!\bigl(h(\sclambda_0),\ldots,h(\sclambda_{n-1})\bigr)\mtV^{\Tr}.
    \label{eq:spectral_graph_filter}
\end{equation}

\noindent\textbf{Graph neural networks.}
A graph neural network (GNN) is defined by the propagation rule at layer $\ell = 0, \ldots, N_L-1$
\begin{equation}
    \mtX^{(\ell+1)}
    =
    \fnphi\!\left(
        \sum_{m=0}^{K_\ell}
        \mtL^m\mtX^{(\ell)}\mtW_m^{(\ell)}
    \right),\quad \mtX^{(0)} = \mtX
    \label{eq:message_passing_gnn}
\end{equation}
where $\mtX^{(\ell)}$ collects the node features at layer $\ell$, $\mtW_m^{(\ell)}$ are learnable feature-mixing matrices, $K_\ell$ is the filter order, and $\fnphi$ is the entrywise nonlinearity. Before the activation, the graph filter bank acts separately on each graph frequency, mixing features through $\{\mtW_m^{(\ell)}\}_m$. The nodewise activation can then mix graph frequencies.

\noindent\textbf{Graph heat diffusion.}
Heat diffusion on a graph $\graph$ is defined by the Laplacian-based
differential equation~\cite{chungspectral,smola2003kernels,hammond2009waveletsgraphsspectralgraph,thanou2016learningheatdiffusiongraphs}
\begin{equation}
    \frac{d \vcu_t}{dt}
    =
    -c\,\mtL\vcu_t,
    \qquad
    \vcu_0=\vcx,
    \label{eq:graph_heat_equation}
\end{equation}
where $\vcu_t\in\fdR^n$ is a time-varying graph signal and $c>0$ sets the diffusion rate (speed). %
The solution of \eqref{eq:graph_heat_equation} reads as
\begin{equation*}
\vcu_t
=
e^{-ct\mtL}\vcx
=
\psi_t(\mtL)\vcx,
\qquad
\psi_t(\sclambda)=e^{-ct\sclambda},
\end{equation*}
where $e^{-ct\mtL}$ is the graph heat kernel~\cite{thanou2016learningheatdiffusiongraphs}, i.e., a spectral graph filter \eqref{eq:spectral_graph_filter} that acts on the GFT coefficients as $\sctu_{t,k}=e^{-ct\sclambda_k}\sctx_k$. The frequency response $\psi_t(\sclambda)$ is a low-pass filter whose bandwidth shrinks with $t$. In the steady state, $t\to\infty$, we have $e^{-ct\sclambda_k}\to0$ for every eigenvalue $\sclambda_k>0$ while for the null eigenvalue we have $e^{-ct\sclambda_0}=1$, so, for a connected graph, the output converges to the average of the input signal, $\vcu_t\to\vcv_0\vcv_0^{\Tr}\vcx$.

\noindent\textbf{Score-based generative modeling.}
Score-based diffusion models define a forward stochastic process that maps the data distribution $\vcx_0\sim\pi_{\mathrm{data}}$ to a tractable prior $\vcx_T\sim\pi_T$ over a horizon $[0,T]$, and learn a reverse process that maps samples of $\pi_T$ back to $\pi_{\mathrm{data}}$~\cite{sohldickstein2015deepunsupervisedlearningusing,song2021scorebasedgenerativemodelingstochastic}. In continuous time, the forward stochastic differential equation (SDE) reads as
\begin{equation}
    d\vcx_t
    =
    \mu(\vcx_t,t)\,dt
    +
    \beta(t)\,d\vcw_t,
    \label{eq:general_forward_sde}
\end{equation}
where $\vcx_t$ is the state at diffusion time $t$ with density $\pi_t$, $\vcw_t$ is a standard Brownian motion, $\mu$ is the drift, and $\beta$ the diffusion coefficient. Under standard assumptions~\cite{anderson1982reverse}, the reverse-time SDE takes the form
\begin{equation}
    d\vcx_t
    =
    \left[
    \mu(\vcx_t,t)
    -
    \beta(t)^2\nabla_{\vcx_t}\log \pi_t(\vcx_t)
    \right]dt
    +
    \beta(t)\,d\vcbw_t,
    \label{eq:general_reverse_sde}
\end{equation}
with $\vcbw_t$ a reverse-time Brownian motion. The score $\nabla_{\vcx_t}\log\pi_t$ is unknown and is learned from data, commonly by denoising~\cite{vincent2011connection}, where a network predicts from $\vcx_t$ the clean signal, the injected noise, or the score itself~\cite{ho2020denoisingdiffusionprobabilisticmodels,song2021scorebasedgenerativemodelingstochastic}. Sampling starts from the terminal prior and integrates \eqref{eq:general_reverse_sde} backwards.

\section{Graph-Aware Diffusion}
\label{sec:graph_aware_diffusion}

In the forward SDE \eqref{eq:gad_forward_sde_1}, GAD uses the graph heat flow \eqref{eq:graph_heat_equation} with a shifted Laplacian and adds Brownian noise. To generate graph signals, GAD starts from a Gaussian prior determined by the graph and solves the reverse SDE \eqref{eq:gad_reverse_sde} with a learned score.

\subsection{Forward and Reverse Processes}
\label{subsec:gad_processes}

GAD drives the heat flow with the \emph{shifted Laplacian}
\begin{equation}
    \mtL_{\scgamma}=\mtL+\scgamma\mtI,
    \qquad
    \scgamma>0,
    \qquad
    \sclambda_{\scgamma,k}=\sclambda_k+\scgamma,
    \label{eq:shifted_laplacian}
\end{equation}
in place of $\mtL$. The specialized forward SDE reads as
\begin{equation}
    d\vcx_t
    =
    -c_t\mtL_{\scgamma}\vcx_t\,dt
    +
    \sqrt{2c_t}\,\scsigma\,d\vcw_t,
    \label{eq:gad_forward_sde_1}
\end{equation}
with diffusion rate $c_t\geq0$ and noise scale $\scsigma>0$. The diffusion rate is shared by all graph frequencies, and choosing it is the noise schedule of GAD. Relative to
\eqref{eq:general_forward_sde}, the drift is 
$\mu(\vcx,t)=-c_t\mtL_{\scgamma}\vcx$ and the diffusion coefficient is $\beta(t)=\sqrt{2c_t}\,\scsigma$. In the graph Fourier basis, the \textit{drift} of the $k$th GFT coefficient is $-c_t\sclambda_{\scgamma,k}\sctx_{t,k}$. Since the Laplacian matrix has a zero eigenvalue, the contraction is weakest at $\sclambda_0=0$, where the drift is $-c_t\scgamma\,\sctx_{t,0}$. Without the shift term $\scgamma\mtI$, the component associated with the zero eigenvalue of the Laplacian evolves as a time-changed Brownian motion. Its noise variance $2\scsigma^{2}\scbc_t$ grows linearly with the integrated rate defined next, precluding a stationary law for \eqref{eq:gad_forward_sde_1}. Next, we define the integrated rate
\begin{equation}
   \scbc_t=\int_0^t c_s\,ds,
   \qquad
   \scbc_0=0,
   \label{eq:integrated_schedule}
\end{equation}
which we take strictly increasing, with $c_t>0$ for $t>0$. Since the drift is linear and the noise additive, the transition of \eqref{eq:gad_forward_sde_1} is Gaussian~\cite{sarkka2019applied},
\begin{equation}
\begin{gathered}
    \vcx_t\mid\vcx_0
    \sim
    \gaussDist\!\left(\mtH_t\vcx_0,\mtSigma_t\right),
    \\[2pt]
    \mtH_t
    =
    \exp\!\left(-\scbc_t\mtL_{\scgamma}\right),
    \qquad
    \mtSigma_t
    =
    \scsigma^2
    \left(\mtI-\mtH_t^2\right)
    \mtL_{\scgamma}^{-1}.
\end{gathered}
\label{eq:gad_forward_marginal}
\end{equation}
Equivalently, the corrupted signal is $\vcx_t=\mtH_t\vcx_0+\vce_t$, where $\vce_t\sim\gaussDist(\vcZeros,\mtSigma_t)$, the attenuation operator $\mtH_t$ attenuates the clean signal, and $\vce_t$ is the accumulated forward noise. As the integrated rate grows without bound, $\scbc_t\to\infty$, the attenuation operator vanishes, $\mtH_t\to\mtZeros$, and the covariance converges to $\mtSigma_t\to\scsigma^2\mtL_{\scgamma}^{-1}$. Thus, the limiting distribution is
\begin{equation}
    \vcx_\infty
    \sim
    \gaussDist\!\left(
        \vcZeros,\,
        \scsigma^2\mtL_{\scgamma}^{-1}
    \right),
    \label{eq:gad_stationary_prior}
\end{equation}
which is a Laplacian-based Gaussian Markov random field. We refer to it as the GMRF prior.
At a finite horizon $T$, the attenuation operator $\mtH_T$ maps the clean signal to the conditional mean, $\mathbb{E}[\vcx_T\mid\vcx_0]=\mtH_T\vcx_0$. Its squared operator norm, $\|\mtH_T\|_2^2=e^{-2\scgamma\scbc_T}$, is the worst-case squared attenuation of $\mtH_T$, determined by the slowest-decaying GFT coefficient, associated with the smallest Laplacian eigenvalue, $\sclambda_0=0$.

\begin{definition}[$\scdelta$-complete horizon]
\label{def:delta_complete}
For a tolerance $\scdelta\in(0,1)$, the horizon $T$ is \emph{$\scdelta$-complete} if the forward mean retains at most a fraction $\scdelta$ of the clean-signal power at any graph frequency
\begin{equation}
    \|\mtH_T\|_2^{2}
    =
    e^{-2\scgamma\scbc_T}
    \leq
    \scdelta
    \iff
    \scbc_T
    \geq
    \frac{\log(1/\scdelta)}{2\scgamma} .
    \label{eq:delta_complete}
\end{equation}
\end{definition}
\noindent
The shift $\scgamma$ sets both the largest prior variance, $\scsigma^{2}/\scgamma$, attained at the lowest graph frequency, and the slowest decay of the forward mean, $e^{-\scgamma\scbc_t}$. A smaller $\scgamma$ therefore preserves more low-frequency variability in the prior but requires a larger integrated rate to reach $\scdelta$-completeness.

The reverse process requires the score of the marginal density of $\vcx_t$, $\fns_t(\vcx)=\nabla_{\vcx}\log\pi_t(\vcx)$. Substituting the drift and the diffusion coefficient of \eqref{eq:gad_forward_sde_1} into \eqref{eq:general_reverse_sde} gives the reverse-time SDE
\begin{equation}
    d\vcx_t
    =
    \left[
        -c_t\mtL_{\scgamma}\vcx_t
        -
        2c_t\scsigma^2
        \fns_t(\vcx_t)
    \right]dt
    +
    \sqrt{2c_t}\,\scsigma\,d\vcbw_t,
    \label{eq:gad_reverse_sde}
\end{equation}
with $\vcbw_t$ the reverse-time Brownian motion of \eqref{eq:general_reverse_sde}. The score of the transition density is linear in the corrupted signal, $\nabla_{\vcx_t}\log\pi(\vcx_t\mid\vcx_0)=-\mtSigma_t^{-1}(\vcx_t-\mtH_t\vcx_0)$, and the score of the marginal is its conditional expectation given $\vcx_t$. Taking that expectation yields Tweedie's formula~\cite{Efron01122011},
\begin{equation}
    \fns_t(\vcx_t)
    =
    -\mtSigma_t^{-1}
    \left( \vcx_t-
        \mtH_t\,
        \xp[\vcx_0\mid\vcx_t,t]             
    \right),
    \label{eq:gad_tweedie}
\end{equation}
which expresses the score through the posterior mean of the clean graph signal. GAD estimates that posterior mean by a graph-aware denoising regression. More specifically, let $f_{\Theta}(\vcx_t,\mtL,t)$ be a denoiser with parameters $\Theta$, and let $\pi_{\mathrm{tr}}$ be the uniform distribution on the horizon $[0,T]$. Training minimizes the denoising error, i.e., the mean squared error between the clean signal and the denoiser output, over diffusion time and pairs $(\vcx_0,\vcx_t)$ that the forward process generates at that time,
\begin{equation}
    \mathcal{L}_{\mathrm{den}}(\Theta)
    =
    \xp_{t\sim\pi_{\mathrm{tr}}}
    \!\left[
        \xp_{
            \substack{
                \vcx_0\sim\pi_{\mathrm{data}} \\
                \vcx_t\sim\pi(\cdot\mid\vcx_0,t)
            }
        }
        \!\left[
            \left\|
                \vcx_0
                -
                f_{\Theta}(\vcx_t,\mtL,t)
            \right\|_2^2
        \right]
    \right],
    \label{eq:gad_mmse_param}
\end{equation}
where $\pi_{\mathrm{data}}$ is the data distribution and $\pi(\cdot\mid\vcx_0,t)$ the transition \eqref{eq:gad_forward_marginal}. At each fixed time $t$, the posterior mean $\xp[\vcx_0\mid\vcx_t,t]$ minimizes the conditional denoising risk in \eqref{eq:gad_mmse_param} over all square-integrable functions of $\vcx_t$. Denoising is therefore a minimum mean squared error (MMSE) estimation problem, and the error of the posterior mean is the MMSE. Substituting the trained denoiser $f_{\Theta^\star}$ for that posterior mean in \eqref{eq:gad_tweedie} gives the plug-in score, and with it the learned reverse process \eqref{eq:gad_reverse_sde}.

\subsection{Spectral Characterization}
\label{subsec:gad_spectral}

Since the attenuation operator $\mtH_t$ and covariance $\mtSigma_t$ are functions of $\mtL_{\scgamma}$, the forward transition, the linear part of the reverse drift, and the coefficients of Tweedie's formula share its eigenbasis. Expressing the dynamics in this basis gives one equation per graph frequency
and makes explicit, for each GFT coefficient, the interplay between its graph frequency, the spectral content of the data, and the shared diffusion rate.

\smallskip
\noindent\textbf{Forward.} Let the GFTs of the corrupted signal, the clean signal and the forward noise be $\vctx_t=\mtV^{\Tr}\vcx_t$, $\vctx_0=\mtV^{\Tr}\vcx_0$ and $\vcte_t=\mtV^{\Tr}\vce_t$, with respective $k$th coefficients $\sctx_{t,k}$, $\sctx_{0,k}$ and $\scte_{t,k}$. The GFT of the driving noise, $\vctw_t=\mtV^{\Tr}\vcw_t$, is a standard Brownian motion as $\mtV$ is orthogonal. Using these GFTs, the forward SDE \eqref{eq:gad_forward_sde_1} decouples into $n$ time-inhomogeneous Ornstein--Uhlenbeck equations,
\begin{equation}
    d\sctx_{t,k}
    =
    -c_t\sclambda_{\scgamma,k}\sctx_{t,k}\,dt
    +
    \sqrt{2c_t}\,\scsigma\,d\sctw_{t,k},
    ~
    k=0,...,n-1 .
    \label{eq:gad_spectral_forward_mode}
\end{equation}
Conditioned on $\vcx_0$, their solutions are
\begin{equation}
    \sctx_{t,k}
    =
    a_{t,k}\sctx_{0,k}
    +
    \scte_{t,k},
    \qquad
    \scte_{t,k}
    \sim
    \gaussDist(0,\scnu_{t,k}),
    \label{eq:gad_channel_view}
\end{equation}
with attenuation $a_{t,k}$ and noise variance $\scnu_{t,k}$ given by
\begin{equation}
    a_{t,k}
    =
    \exp\!\left(
        -\sclambda_{\scgamma,k}\scbc_t
    \right),
    \qquad
    \scnu_{t,k}
    =
    \frac{\scsigma^2}{\sclambda_{\scgamma,k}}
    \left(1-a_{t,k}^2\right).
    \label{eq:gad_mode_attenuation_variance}
\end{equation}
The attenuation $a_{t,k}$ and the noise variance $\scnu_{t,k}$ are the eigenvalues of $\mtH_t$ and $\mtSigma_t$ on $\vcv_k$, respectively. The noise coefficients $\scte_{t,k}$ are independent across graph frequencies and independent of $\vcx_0$. The covariance between two distinct corrupted GFT coefficients $\sctx_{t,j}$ and $\sctx_{t,k}$ equals $a_{t,j}a_{t,k}\operatorname{Cov}(\sctx_{0,j},\sctx_{0,k})$, and it vanishes if and only if the clean coefficients are uncorrelated.

Equation~\eqref{eq:gad_channel_view} is the graph counterpart of blurring diffusion~\cite{hoogeboom2023blurring}, in which the attenuation and the noise variance are a free per-frequency choice with a white terminal law. Here, they follow from the Laplacian-based SDE.

The attenuation and the noise variance \eqref{eq:gad_mode_attenuation_variance} state how much of the clean coefficient survives and how much noise replaces it. To compare two graph frequencies, we combine them into the ratio of the retained signal power to the injected noise power, which needs the power at each graph frequency in the clean signal and in the prior. We define the clean power at graph frequency $\sclambda_k$ as the second order moment of the $k$th GFT coefficient, $p_k=\xp[\sctx_{0,k}^{\,2}]$, and we call $\{p_k\}_{k=0}^{n-1}$ the power spectrum of the data. Its prior power is the second order moment under the GMRF prior, $q_k=\xp[\sctx_{\infty,k}^{\,2}]=\scsigma^{2}/\sclambda_{\scgamma,k}$, and its power ratio is $g_k=p_k/q_k=\sclambda_{\scgamma,k}p_k/\scsigma^{2}$. The ratio $g_k$ increases with graph frequency when $p_k$ decays more slowly than $1/\sclambda_{\scgamma,k}$.

\begin{definition}[Frequency-wise $\snr$]
\label{def:modewise_snr}
For $t\in(0,T]$, the frequency-wise SNR at graph frequency $\sclambda_k$ is the power retained from the clean coefficient divided by the variance of the injected noise. Leveraging  \eqref{eq:gad_mode_attenuation_variance} and $g_k=\sclambda_{\scgamma,k}p_k/\scsigma^{2}$, it holds that, under a shared diffusion rate, the SNR is
\begin{equation}
    \snr_{t,k}
    =
    \frac{a_{t,k}^{2}\,p_k}{\scnu_{t,k}}
    =
    \frac{g_k}{\exp(2\sclambda_{\scgamma,k}\scbc_t)-1} .
    \label{eq:modewise_snr_definition}
\end{equation}
\end{definition}

\noindent
Definition~\ref{def:modewise_snr} assigns to each graph frequency the scalar SNR that variational diffusion models attach to a whole noise schedule~\cite{kingma2021variational}. The integrated rate is one scalar for all graph frequencies, and $\snr_{t,k}$ depends on it only through the exponent $\sclambda_{\scgamma,k}\scbc_t$. %
For two graph frequencies $\sclambda_j$ and $\sclambda_k$ and diffusion time $t>0$, the ratio of their SNRs factorizes as
\begin{equation}
    \frac{\snr_{t,j}}{\snr_{t,k}}
    =
    \underbrace{
        \frac{p_j}{p_k}
    }_{\text{clean signal}}
    \cdot
    \underbrace{
        \frac{\sclambda_{\scgamma,j}}{\sclambda_{\scgamma,k}}
        \,
        \frac{\exp(2\sclambda_{\scgamma,k}\scbc_t)-1}
             {\exp(2\sclambda_{\scgamma,j}\scbc_t)-1}
    }_{\text{graph--rate interaction}} .
    \label{eq:snr_ratio}
\end{equation}
The first factor is the power imbalance present in the clean signal, and no diffusion rate affects it. The second factor tends to one as $\scbc_t\to0$ and, for $\sclambda_{\scgamma,j}>\sclambda_{\scgamma,k}$, decays as $\exp(-2(\sclambda_{\scgamma,j}-\sclambda_{\scgamma,k})\scbc_t)$ once $\scbc_t$ is large. A shared diffusion rate therefore separates the SNRs of two graph frequencies exponentially in their gap, and the separation widens for as long as the integrated rate grows. We call this separation the spectral imbalance, and we measure it at diffusion time $t$ by the spread $\max_k\snr_{t,k}/\min_k\snr_{t,k}$. The left panel of Fig.~\ref{fig:forward_spectral_diagnostics} shows the SNRs of different graph frequencies diverging along the horizon.

\smallskip
\noindent\textbf{Reverse.} Let the $k$th GFT coefficients of the score and of the reverse-time Brownian motion be respectively $\scts_{t,k}(\vcx_t)=[\mtV^{\Tr}\fns_t(\vcx_t)]_k$ and $\sctbw_{t,k}=[\mtV^{\Tr}\vcbw_t]_k$. Projecting the reverse SDE \eqref{eq:gad_reverse_sde} onto the graph Fourier basis yields
\begin{equation}
    d\sctx_{t,k}
    =
    \left[
        -c_t\sclambda_{\scgamma,k}\sctx_{t,k}
        -
        2c_t\scsigma^{2}\,\scts_{t,k}(\vcx_t)
    \right]dt
    +
    \sqrt{2c_t}\,\scsigma\,d\sctbw_{t,k}.
    \label{eq:gad_spectral_reverse_mode}
\end{equation}
Likewise, projecting Tweedie's formula \eqref{eq:gad_tweedie} gives the $k$th GFT coefficient of the score,
\begin{equation}
    \scts_{t,k}(\vcx_t)
    =
    \frac{
        a_{t,k}\,\xp[\sctx_{0,k}\mid\vcx_t,t]
        -
        \sctx_{t,k}
    }{\scnu_{t,k}} .
    \label{eq:gad_exact_spectral_score}
\end{equation}
The plug-in score $\schts_{t,k}$ replaces the posterior mean in \eqref{eq:gad_exact_spectral_score} by the $k$th GFT coefficient of the trained denoiser, $[\mtV^{\Tr}f_{\Theta^\star}(\vcx_t,\mtL,t)]_k$. The attenuation $a_{t,k}$ and the noise variance $\scnu_{t,k}$ therefore appear in two places. First, they fix how much of the clean coefficient is retained in $\sctx_{t,k}$. Second, they convert the denoiser output into the score $\scts_{t,k}$, which the reverse SDE \eqref{eq:gad_spectral_reverse_mode} turns into the drift of that coefficient. The separation of the forward process across graph frequencies does not carry over to the reverse. The posterior mean $\xp[\sctx_{0,k}\mid\vcx_t,t]$ conditions on the whole vector $\vcx_t$, and it can depend on other GFT coefficients when the clean coefficients are statistically dependent.

\section{Spectral Analysis of Denoising Regression}
\label{sec:denoising_consequence}

In Section~\ref{sec:graph_aware_diffusion}, we showed that a shared diffusion rate places different graph frequencies at different SNRs at the same diffusion time $t$. This matters because GAD learns the score only through the denoiser. By \eqref{eq:gad_exact_spectral_score}, the error of the plug-in score at $\sclambda_k$ is $a_{t,k}/\scnu_{t,k}$ times the error of the denoiser output with respect to the posterior mean at that frequency, and errors in the score propagate to the reverse drift and from there to the generated samples. The frequency-wise denoising error therefore affects how accurately each graph frequency is generated. In this section, we characterize this error through three quantities: the error attainable at each graph frequency, the time constant at which training reaches that error, and the diffusion times at which the frequency contributes to training. We first consider linear denoisers in Section~\ref{subsec:fixed_time_spectral_regression}. Their minimum error, the linear MMSE, i.e., the MMSE restricted to linear denoisers, is in closed form and coincides with the MMSE for Gaussian data. Section~\ref{sec:nonlinear_gnn_denoising} relates it to the error of a GNN denoiser. Proofs are in Appendix~\ref{app:denoising_regression}.

Our analysis relies on the following assumptions. 

\begin{assumption}[Distinct Laplacian eigenvalues]
\label{ass:simple_spectrum}
The eigenvalues of $\mtL$ are distinct, i.e., $0=\sclambda_0<\sclambda_1<\cdots<\sclambda_{n-1}$.
\end{assumption}

\begin{assumption}[Graph wide-sense stationarity~\cite{marques2017stationary,perraudin2017stationary}]
\label{ass:graph_wss}
The covariance of the clean graph signal is diagonal in the graph Fourier basis of \eqref{eq:laplacian_eigendecomposition},
\begin{equation}
    \mtP
    =
    \operatorname{Cov}(\vcx_0)
    =
    \mtV\diag(p_0,\ldots,p_{n-1})\mtV^{\Tr},
    \label{eq:clean_spectral_power}
\end{equation}
with $p_k=\xp[\sctx_{0,k}^{\,2}]$ the clean power at $\sclambda_k$.
\end{assumption}

Assumption~\ref{ass:simple_spectrum} concerns the graph.  We aim to assign an error and a training time constant to each graph frequency, which requires each of them to identify a single eigenvector. When $\mtL$ has repeated eigenvalues, any rotation of the eigenvectors within the eigenspace is also a valid basis and the per-frequency quantities would change with it. Assumption~\ref{ass:graph_wss} concerns the data. The forward process corrupts each GFT coefficient with noise that is uncorrelated across graph frequencies. For the denoising problem to separate across graph frequencies as well, we need the clean coefficients to be uncorrelated.
This implies that the $n$ powers $p_k$ fully describe the second-order statistics of $\vcx_0$. Since the error of a linear denoiser depends on $(\vcx_0,\vcx_t)$ only through second moments, Assumption~\ref{ass:graph_wss} reduces the regression to $n$ scalar problems, one per GFT coefficient. Without loss of generality, we take $\xp[\vcx_0]=\vcZeros$, as the training data are centered. Throughout, we fix $t\in(0,T]$ and consider the fixed-time denoising error
\begin{equation}
    \stL_t(f)
    =
    \xp\!\left[\|\vcx_0-f(\vcx_t)\|_2^2\,\middle|\,t\right],
    \label{eq:fixed_time_risk}
\end{equation}
which \eqref{eq:gad_mmse_param} averages over the diffusion time.

\subsection{Linear Denoisers}
\label{subsec:fixed_time_spectral_regression}

Consider the linear denoiser $f_{\mtW}(\vcx_t)=\mtW\vcx_t$ with $\mtW\in\fdR^{n\times n}$. This class contains the graph convolutional filter \eqref{eq:graph_filter}, the spectral graph filter \eqref{eq:spectral_graph_filter}, and the GNN \eqref{eq:message_passing_gnn} with identity nonlinearity. We characterize three quantities per graph frequency: (i) the error of the optimal linear denoiser, (ii) the rate at which gradient descent converges to it, and (iii) the diffusion times at which it contributes to the fixed-time error $\stL_t$. All three depend on the second moment of the $k$th corrupted GFT coefficient,
\begin{equation}
    \scrho_{t,k}
    =
    \xp[\sctx_{t,k}^{\,2}]
    =
    a_{t,k}^{2}p_k+\scnu_{t,k}
    =
    a_{t,k}^{2}p_k
    +
    \bigl(1-a_{t,k}^{2}\bigr)q_k
    >0 ,
    \label{eq:effective_variance_convex_combination}
\end{equation}
which we call the \textit{corrupted power}. It is a convex combination of the clean power $p_k$ and the prior power $q_k$.

\smallskip
\noindent\textbf{Optimal linear denoiser.}
Under Assumption~\ref{ass:graph_wss}, the corrupted GFT coefficients are uncorrelated across graph frequencies, and the minimizer of $\stL_t(f_{\mtW})$ is the spectral graph filter
\begin{equation}
    f_{\vch_t}(\vcx_t)
    =
    \mtV\diag(\vch_t)\mtV^{\Tr}\vcx_t
    \label{eq:spectral_filter_denoiser}
\end{equation}
with frequency response
\begin{equation}
    h_{t,k}^{\star}
    =
    \frac{a_{t,k}p_k}{\scrho_{t,k}}
    =
    \frac{1}{a_{t,k}}
    \frac{\snr_{t,k}}{1+\snr_{t,k}},
    \quad
    k=0,\ldots,n-1
    \label{eq:optimal_modewise_gain}
\end{equation}
(Lemma~\ref{lem:optimal_spectral_response}). This is the Wiener filter of the scalar Gaussian model for $\sctx_{t,k}$, which inverts the attenuation $a_{t,k}$ and scales the result by a factor that decreases with the SNR. Each clean coefficient $\sctx_{0,k}$ is estimated from its own corrupted coefficient $\sctx_{t,k}$, and the error of this denoiser is formalized next.

\begin{proposition}[Frequency-wise normalized error]
\label{thm:modewise_denoising_difficulty}
Let Assumptions~\ref{ass:simple_spectrum} and~\ref{ass:graph_wss} hold, let $t\in(0,T]$, and $p_k>0$. Then, at $\sclambda_k$ the optimal linear denoiser \eqref{eq:spectral_filter_denoiser}--\eqref{eq:optimal_modewise_gain} yields the normalized error
\begin{equation}
    d_{t,k}
    =
    \frac{
        \xp\!\left[
            (\sctx_{0,k}-h_{t,k}^{\star}\sctx_{t,k})^{2}
            \,\middle|\,t
        \right]
    }{p_k}
    =
    \frac{1}{1+\snr_{t,k}} .
    \label{eq:modewise_difficulty}
\end{equation}
\end{proposition}

\noindent
Summing over graph frequencies gives the linear MMSE, i.e., the minimum fixed-time error over linear denoisers,
\begin{equation}
    \stB_t
    =
    \min_{\mtW}\stL_t(f_{\mtW})
    =
    \sum_{k=0}^{n-1}
    p_k\,d_{t,k} .
    \label{eq:benchmark_risk}
\end{equation}
Both the normalized error $d_{t,k}$ and the linear MMSE $\stB_t$ depend on $a_{t,k}$ and $\scnu_{t,k}$ only through $\snr_{t,k}$. The normalized error $d_{t,k}$ goes from zero for $\snr_{t,k}\gg1$ to one for $\snr_{t,k}\ll1$. Because of the spectral imbalance \eqref{eq:snr_ratio}, once $\scbc_t$ is large enough for the exponential factor to dominate the power ratio $p_j/p_k$, a shared diffusion rate keeps the highest graph frequencies at the lowest SNR, so they contribute $p_kd_{t,k}\approx p_k$ to $\stB_t$. Reducing $\stB_t$ requires a higher SNR at these graph frequencies, which a shared diffusion rate can achieve only by reducing $\scbc_t$. This raises the SNR at all graph frequencies at that time. Keeping $\scbc_T$ fixed then shifts part of the corruption to later times.

\smallskip
\noindent\textbf{Training time constant.}
Under a finite number of training iterations, a graph frequency whose frequency response is still far from $h_{t,k}^{\star}$ still has a larger error, so we want to characterize how fast it converges. Since the minimizer is a spectral graph filter, we train the $n\times 1$ frequency response $\vch_t$ within this class using gradient descent with learning rate $\eta>0$. 
At fixed diffusion time $t$, we study the gradient flow
$d\vch_t/d\sctau=-\eta\nabla_{\vch_t}\stL_t(f_{\vch_t})$,
where $\sctau$ is optimization time and $\eta>0$.

\begin{proposition}[Frequency-wise training time constant]
\label{thm:optimization_spectrum}
Let Assumptions~\ref{ass:simple_spectrum} and~\ref{ass:graph_wss} hold, let $t\in(0,T]$, and let the frequency response $\vch_t(\sctau)$ follow the gradient flow introduced above from any initialization $\vch_t(0)$. Then, the frequency response at each graph frequency converges to $h_{t,k}^{\star}$ in \eqref{eq:optimal_modewise_gain} independently of the others,
\begin{equation}
    h_{t,k}(\sctau)-h_{t,k}^{\star}
    =
    e^{-\sctau/\sctau_{t,k}}
    \bigl(
        h_{t,k}(0)-h_{t,k}^{\star}
    \bigr),
    \quad
    \sctau_{t,k}
    =
    \bigl(2\eta\,\scrho_{t,k}\bigr)^{-1} ,
    \label{eq:spectral_training_dynamics}
\end{equation}
\noindent
where $\sctau_{t,k}$ is the training time constant at $\sclambda_k$.
\end{proposition}

\noindent
The training time constant at $\sclambda_k$ depends only on the corrupted power $\scrho_{t,k}$, hence the graph frequencies are learned in decreasing order of that power. This is the spectral bias of diffusion training~\cite{wang2024analytical}, and under GAD the diffusion rate sets it through the attenuation $a_{t,k}$ in $\scrho_{t,k}$. Training all graph frequencies to the same accuracy requires waiting for the slowest one, which takes $\max_k\scrho_{t,k}/\min_k\scrho_{t,k}$ times longer than the fastest. We call this ratio the spread of the training time constants. %
As $a_{t,k}\to0$ for all $k$, the corrupted powers approach the prior powers $q_k$, and the spread tends to the condition number of the shifted Laplacian, $\cond(\mtL_{\scgamma})=(\sclambda_{n-1}+\scgamma)/\scgamma$. We call this limit the terminal spread, with the highest graph frequency being the slowest. This limit depends only on $\mtL_{\scgamma}$.

\smallskip
\noindent\textbf{Informative graph frequencies.}
Propositions~\ref{thm:modewise_denoising_difficulty} and~\ref{thm:optimization_spectrum} hold at a fixed diffusion time, whereas training samples $t$ over the whole horizon. By Proposition~\ref{thm:modewise_denoising_difficulty}, a graph frequency with $\snr_{t,k}\gg1$ has $d_{t,k}\approx0$ and one with $\snr_{t,k}\ll1$ has $d_{t,k}\approx1$. The optimal linear response approximately inverts the attenuation at high SNR and approaches the prior-mean estimate at low SNR~\cite{hang2023minsnr,kingma2023understanding}. We therefore investigate which graph frequencies a shared diffusion rate keeps at a moderate SNR, and when.

\begin{definition}[$\sceps$-informative graph frequency]
\label{def:informative_mode}
Let $\sceps\in(0,1)$ be a tolerance. Graph frequency $\sclambda_k$ is \emph{$\sceps$-informative} at diffusion time $t$ if $\snr_{t,k}\in[\sceps,\sceps^{-1}]$.
\end{definition}

\noindent
We call the set of $\sceps$-informative graph frequencies at time $t$, $\{\sclambda_k:\snr_{t,k}\in[\sceps,\sceps^{-1}]\}$, the \textit{informative band}. Graph frequencies with $\snr_{t,k}>\sceps^{-1}$ are preserved at time $t$, and those with $\snr_{t,k}<\sceps$ are lost. The value of $\snr_{t,k}$ decreases strictly in $\scbc_t$, so graph frequency $\sclambda_k$ is informative on a single interval of integrated-rate values, of length at most $\log(1/\sceps)/\sclambda_{\scgamma,k}$ (Lemma~\ref{lem:informative_window}). If $p_k$ is nonincreasing with graph frequency, these intervals start earlier and are shorter at higher frequencies. Their probability under the training law depends on both $\scbc_t$ and $\pi_{\mathrm{tr}}$. The next proposition bounds the width of this band.

\begin{proposition}[Informative band]
\label{prop:informative_band}
Let the diffusion rate be shared, let $p_k>0$ for all $k$, and let $g_{\min}=\min_kg_k$ and $g_{\max}=\max_kg_k$. Then, any two graph frequencies $\sclambda_j$ and $\sclambda_k$ that are $\sceps$-informative at the same diffusion time satisfy
\begin{equation}
    \frac{\sclambda_{\scgamma,j}}{\sclambda_{\scgamma,k}}
    \;\leq\;
    \frac{\log\bigl(1+g_{\max}\sceps^{-1}\bigr)}
         {\log\bigl(1+g_{\min}\sceps\bigr)} .
    \label{eq:informative_band_ratio}
\end{equation}
\end{proposition}

\noindent 
The bound constrains the band when it is smaller than $\cond(\mtL_{\scgamma})$. With a shared diffusion rate, all graph frequencies can be informative at the same time, but distinct frequencies cannot maintain equal SNRs over a time interval. This motivates our design in Section~\ref{sec:mode_adaptive_snr}.

\subsection{GNN Denoiser}
\label{sec:nonlinear_gnn_denoising}

In practice, GAD trains the time-conditioned GNN $f_{\Theta}(\vcx_t,\mtL,t)$ of \eqref{eq:message_passing_gnn}, whose output at each graph frequency can depend on the whole vector $\vctx_t$. We analyze its fixed-time error $\stL_t(f_{\Theta})$ by comparing the GNN with two estimators: the posterior mean $\fnm_t(\vcx)=\xp[\vcx_0\mid\vcx_t=\vcx,t]$, whose error is the MMSE, and the optimal linear denoiser $h_{t,k}^{\star}$, whose error is the linear MMSE $\stB_t$.

\begin{proposition}[GNN error decomposition]
\label{thm:gnn_risk_decomposition}
Let Assumptions~\ref{ass:simple_spectrum} and~\ref{ass:graph_wss} hold, let $t\in(0,T]$, and let $m_{t,k}=[\mtV^{\Tr}\fnm_t(\vcx_t)]_k$ and $\widehat{\sctx}_{0,k}=[\mtV^{\Tr}f_{\Theta}(\vcx_t,\mtL,t)]_k$ be the $k$th GFT coefficients of the posterior mean and of the GNN output. Then, the fixed-time error $\stL_t(f_{\Theta})$ of the GNN decomposes as
\begin{equation}
    \stL_t(f_{\Theta})
    =
    \stB_t
    -
    \scDelta_{\mathrm{PM},t}
    +
    \scOmega_{\Theta,t},
    \label{eq:gnn_risk_decomposition}
\end{equation}
where $\stB_t$ is the linear MMSE,
\begin{equation*}
    \scDelta_{\mathrm{PM},t}
    =
    \sum_{k=0}^{n-1}
    \scDelta_{t,k},
    \qquad
    \scDelta_{t,k}
    =
    \xp\!\left[
        (m_{t,k}-h_{t,k}^{\star}\sctx_{t,k})^{2}\,\middle|\,t
    \right]
    \geq0
\end{equation*}
is the gain of the posterior mean over the optimal linear denoiser $h_{t,k}^{\star}$, with $\scDelta_{t,k}$ its frequency-wise term, and
\begin{equation*}
    \scOmega_{\Theta,t}
    =
    \sum_{k=0}^{n-1}
    \xp\!\left[
        (m_{t,k}-\widehat{\sctx}_{0,k})^{2}\,\middle|\,t
    \right]
    \geq0
\end{equation*}
is the error of the GNN with respect to the posterior mean.
\end{proposition}

\noindent
The posterior-mean gain $\scDelta_{\mathrm{PM},t}$ measures the improvement over the optimal linear denoiser. It depends on the data law and the corruption, vanishing for Gaussian data under graph stationarity. Let $m^{\mathrm{sc}}_{t,k}=\xp[\sctx_{0,k}\mid\sctx_{t,k},t]$ be the posterior mean given only the $k$th corrupted coefficient. Then
\begin{equation*}
\begin{aligned}
    \scDelta_{t,k}
    = \,
    \xp[(m_{t,k}-m^{\mathrm{sc}}_{t,k})^{2}\mid t]
        \, + \,
    \xp[(m^{\mathrm{sc}}_{t,k}-h^{\star}_{t,k}\sctx_{t,k})^{2}\mid t].
\end{aligned}
\end{equation*}
The first term is the gain from observing the other corrupted GFT coefficients, and the second is the nonlinear gain from the $k$th coefficient alone. The excess error $\scOmega_{\Theta,t}$ includes approximation and optimization error and is the only term that depends on the GNN parameters. This decomposition follows from orthogonality (Lemma~\ref{lem:bayes_orthogonality}), which also gives the second moment of the posterior mean at each graph frequency.

\begin{corollary}[Second moment of the posterior mean]
\label{prop:bayes_target_energy}
Let Assumptions~\ref{ass:simple_spectrum} and~\ref{ass:graph_wss} hold, let $t\in(0,T]$, and let $p_k>0$. Then, the second moment of the posterior mean $m_{t,k}$ at $\sclambda_k$ is
\begin{equation}
    \xp\!\left[
        m_{t,k}^{2}
        \,\middle|\,t
    \right]
    =
    p_k
    \frac{\snr_{t,k}}{1+\snr_{t,k}}
    +
    \scDelta_{t,k},
    \label{eq:bayes_target_energy}
\end{equation}
where $\scDelta_{t,k}$ is the frequency-wise gain of Proposition~\ref{thm:gnn_risk_decomposition}.
\end{corollary}

\noindent
The first term in \eqref{eq:bayes_target_energy} is the energy
of the optimal linear estimate from $\sctx_{t,k}$. The second,
$\scDelta_{t,k}$, combines the gain from observing the corrupted
coefficients $\sctx_{t,j}$ with $j\neq k$ and the gain from
estimating $\sctx_{0,k}$ nonlinearly from $\sctx_{t,k}$ alone.

Proposition~\ref{thm:optimization_spectrum} does not carry over directly to the GNN. At a fixed diffusion time, off-diagonal entries of its tangent kernel can arise from the nonlinearity $\fnphi$ and from parameters shared across graph frequencies (Appendix~\ref{app:gnn_ntk}). The shared parameters also couple different diffusion times during training.

For low-pass data under a shared diffusion rate, the high graph frequencies are informative over shorter ranges of integrated rates. Near the prior, their fixed-time linear training constants are the largest.

\newcommand{\revtemp}[1]{\textcolor{magenta}{#1}}
\section{Mode-Adaptive SNR Scheduling}
\label{sec:mode_adaptive_snr}

In this section, we design diffusion rates that remove the spectral imbalance identified in Section~\ref{sec:denoising_consequence}. Three quantities from that section depend on the diffusion rate only through the frequency-wise SNR: the normalized linear error $d_{t,k}$, whether a frequency is informative, and the energy of its optimal linear estimate as a fraction of clean power. Equalizing the SNR equalizes all three. A shared diffusion rate cannot maintain this equality over a time interval for distinct eigenvalues. We propose Mode-Adaptive SNR Scheduling (MASS), which assigns each graph frequency its own diffusion rate and imposes a common target SNR on all of them. A frequency-dependent diffusion rate does not change the process that each GFT coefficient follows. It only reparametrizes the diffusion time of that coefficient through its integrated rate $\scbc_{t,k}$. As a result, the GMRF prior is preserved, and MASS synchronizes the graph frequencies along the forward process rather than replacing it. Section~\ref{subsec:mass_schedule} derives the diffusion rates, Section~\ref{subsec:mass_effects} analyzes their effect on training and sampling, and Section~\ref{subsec:mass_algorithms} gives the resulting algorithms and their cost. Proofs are in Appendix~\ref{app:mas_snr_sampling_theory}.

\subsection{MASS Diffusion Rates}
\label{subsec:mass_schedule}

A diffusion rate per graph frequency requires a single eigenvector for each of them, which Assumption~\ref{ass:simple_spectrum} provides. Let $c_{t,k}\geq0$ be the diffusion rate at $\sclambda_k$ and $\scbc_{t,k}=\int_0^tc_{s,k}\,ds$ its integrated rate. We collect them in the graph operator $\mtC_t=\mtV\diag(c_{t,0},\ldots,c_{t,n-1})\mtV^{\Tr}$ and replace the scalar $c_t$ of the forward SDE \eqref{eq:gad_forward_sde_1} by this operator,
\begin{equation}
    d\vcx_t
    =
    -\mtC_t\mtL_{\scgamma}\vcx_t\,dt
    +
    \sqrt{2}\,\scsigma\,
    \mtC_t^{1/2}d\vcw_t .
    \label{eq:mass_forward_sde}
\end{equation}
Since $\mtC_t$ and $\mtL_{\scgamma}$ are diagonal in the same basis, each GFT coefficient still follows \eqref{eq:gad_channel_view}, with $\scbc_t$ replaced by $\scbc_{t,k}$ in $a_{t,k}$ and $\scnu_{t,k}$. A frequency-dependent diffusion rate changes how quickly each GFT coefficient evolves, but not its stationary distribution. The GMRF prior therefore remains
\eqref{eq:gad_stationary_prior}.
Let $r_t>0$ be a continuously differentiable, decreasing target SNR on $(0,T]$, satisfying $\lim_{t\to0^+}r_t=+\infty$ and $r_T=r_{\min}>0$. We exclude $t=0$, where no noise has been injected and the SNR is undefined. We require the SNR at every graph frequency to follow this target,
\begin{equation}
    \snr_{t,k}
    =
    r_t,
    \qquad
    k=0,\ldots,n-1 .
    \label{eq:mas_snr_constraint}
\end{equation}

\begin{theorem}[MASS diffusion rates]
\label{thm:mass_forward_channel}
Let Assumption~\ref{ass:simple_spectrum} hold, let $p_k>0$ for all $k$, and let the target SNR $r_t>0$ be continuously differentiable and decreasing on $(0,T]$, with $\lim_{t\to0^+}r_t=+\infty$. Then, the constraint \eqref{eq:mas_snr_constraint} uniquely determines the
frequency-wise integrated rates
\begin{equation}
    \scbc_{t,k}
    =
    \frac{1}{
        2\sclambda_{\scgamma,k}
    }
    \log\!\left(
        1+\frac{g_k}{r_t}
    \right).
    \label{eq:mas_snr_cumulative_schedule}
\end{equation}
Each integrated rate is nonnegative and nondecreasing in $t$, and $\scbc_{t,k}\to 0$ as $t\to0^+$ (equivalently, as $r_t\to\infty$). The corresponding diffusion rate is
\begin{equation}
    c_{t,k}
    =
    -\frac{1}{2\sclambda_{\scgamma,k}}
    \frac{g_k}{r_t+g_k}
    \frac{d}{dt}\log r_t
    \;\geq\;0 .
    \label{eq:mas_snr_rate_schedule}
\end{equation}
\end{theorem}

\noindent
The rate in \eqref{eq:mas_snr_rate_schedule} is approximately proportional to $p_k$ when $r_t\gg g_k$ and approximately inversely proportional to $\sclambda_{\scgamma,k}$ when $r_t\ll g_k$. Substituting \eqref{eq:mas_snr_cumulative_schedule} into \eqref{eq:gad_mode_attenuation_variance} and using $q_kg_k=p_k$ gives, under MASS,
\begin{align}
    a_{t,k}^{2}
    =
    \frac{r_t}{r_t+g_k},~~
    \scnu_{t,k}
&=
    \frac{p_k}{r_t+g_k},
    ~
    \scrho_{t,k}
    =
    (1+r_t)\,\scnu_{t,k} .
    \label{eq:mas_snr_channel}
\end{align}
The attenuation $a_{t,k}$ and the normalized noise variance $\scnu_{t,k}/p_k$ depend on the graph frequency only through the power ratio $g_k=p_k/q_k$, which compares the power of the data at $\sclambda_k$ with the power that the GMRF prior already assigns to it. Graph frequencies at which the data carry more power than the prior require more attenuation to reach the common SNR $r_t$, and those well explained by the prior require less. MASS thus sets the corruption of each graph frequency from the interplay between the graph and the data.

\smallskip
\noindent\textbf{Target SNR.}
Since $r_t$ diverges at $t=0$, we run the process on $[t_0,T]$ with finite endpoints $r_{t_0}=r_{\max}$ and $r_{T}=r_{\min}$. From here on, the horizon is $[t_0,T]$, and the training law $\pi_{\mathrm{tr}}$ is supported on it. We set tolerances on the fraction of clean power retained by each GFT coefficient, requiring $a_{t_0,k}^{2}\geq1-\scdelta_0$ and $a_{T,k}^{2}\leq\scdelta$ for every $k$, with a start tolerance $\scdelta_0\in(0,1)$ and a tolerance $\scdelta\in(0,1)$. The latter is the $\scdelta$-completeness of Definition~\ref{def:delta_complete} applied at each graph frequency. By \eqref{eq:mas_snr_channel}, these conditions hold if and only if $r_{\max}\geq g_{\max}(1-\scdelta_0)/\scdelta_0$ and $r_{\min}\leq g_{\min}\scdelta/(1-\scdelta)$, with $g_{\max}=\max_kg_k$ and $g_{\min}=\min_kg_k$, hence the target SNR must span
\begin{equation}
    \frac{r_{\max}}{r_{\min}}
    \;\geq\;
    \frac{1-\scdelta_0}{\scdelta_0}\,
    \frac{1-\scdelta}{\scdelta}\,
    \frac{g_{\max}}{g_{\min}} .
    \label{eq:mass_dynamic_range}
\end{equation}
The target SNR depends on the graph only through this span. Between the endpoints, $r_t$ can be chosen freely, and we let the log-SNR decrease linearly in time,
\begin{equation}
    \log r_t
    =
    (1-u_t)\log r_{\max}
    +
    u_t\log r_{\min},
    \label{eq:mas_snr_log_trajectory}
\end{equation}
where $u_t$ increases from $u_{t_0}=0$ to $u_{T}=1$. Finally, MASS needs the clean powers $p_k$. We center the $N$ training signals, estimate
\begin{equation}
    \widehat p_k
    =
    \frac{1}{N}
    \sum_{i=1}^{N}
    \left(
        \vcv_k^{\Tr}\vcx_0^{(i)}
    \right)^2,
    \qquad
    \widehat p_k
    \leftarrow
    \max(\widehat p_k,p_{\min}),
    \label{eq:mas_snr_power_estimate}
\end{equation}
with $p_{\min}>0$, and add the mean back after sampling. The floor keeps the estimated power ratios positive. Using the floored powers $\widehat p_k$, the SNR attained is $\snr_{t,k}=r_t\,p_k/\widehat p_k$.

\subsection{Effect on Training and Sampling}
\label{subsec:mass_effects}

\noindent\textbf{Training.}
With $\snr_{t,k}=r_t$ for all $k$, Proposition~\ref{thm:modewise_denoising_difficulty} gives the same normalized error $d_{t,k}=1/(1+r_t)$ at every graph frequency, and the linear MMSE $\stB_t$ becomes $\xp\|\vcx_0\|_2^{2}/(1+r_t)$, which depends on the data only through the total power $\sum_kp_k$ and does not depend on the Laplacian. Every graph frequency is $\sceps$-informative on the same set of diffusion times $\{t:\sceps\leq r_t\leq\sceps^{-1}\}$, so the informative band then covers all of them. The energy of the optimal linear estimate is the same fraction $r_t/(1+r_t)$ of the clean power at every graph frequency. Under MASS, the posterior-mean gain depends on diffusion time only through $r_t$.

\smallskip
\noindent\textbf{Sampling.}
The spectral imbalance can also affect the stiffness of the reverse SDE. To study step-size restrictions for explicit solvers~\cite{dahlquist1963special,hairer1996solving}, we write the reverse drift in the GFT basis. Let $\tilde{f}_{\Theta}(\vctx,t)=\mtV^{\Tr}f_{\Theta}(\mtV\vctx,\mtL,t)$ be the denoiser in this basis, with $k$th coefficient $\tilde{f}_{\Theta,k}(\vctx,t)$, and let $\schts_{t,k}(\vctx)=(a_{t,k}\tilde{f}_{\Theta,k}(\vctx,t)-\sctx_k)/\scnu_{t,k}$ be the plug-in score obtained by replacing the posterior mean in $\scts_{t,k}$ with $\tilde{f}_{\Theta,k}(\vctx,t)$. Substituting the latter into the reverse SDE \eqref{eq:gad_spectral_reverse_mode} with diffusion rate $c_{t,k}$, and using $\scsigma^{2}/\scnu_{t,k}=\sclambda_{\scgamma,k}/(1-a_{t,k}^{2})$ from \eqref{eq:gad_mode_attenuation_variance}, gives the reverse drift of the $k$th GFT coefficient,
\begin{equation}
    \sctb_{t,k}(\vctx)
    =
    \underbrace{
        c_{t,k}
        \frac{\sclambda_{\scgamma,k}}{1-a_{t,k}^{2}}
    }_{\text{reverse-time gain}}
    \underbrace{
        \Bigl[
            (1+a_{t,k}^{2})\sctx_k
            -
            2a_{t,k}
            \tilde{f}_{\Theta,k}(\vctx,t)
        \Bigr]
    }_{\text{denoising direction}} .
    \label{eq:generic_reverse_drift}
\end{equation}
The reverse-time gain scales the magnitude of the drift of the $k$th GFT coefficient, and the denoising direction determines its sign and its dependence on the denoiser output. For a spectral graph filter with frequency response $h_{t,k}$, we have $\tilde{f}_{\Theta,k}(\vctx,t)=h_{t,k}\sctx_k$, and the drift reduces to
\begin{equation}
    \sctb_{t,k}(\vctx)
    =
    \sczeta_{t,k}\sctx_k,
    \qquad
    \sczeta_{t,k}
    =
    c_{t,k}\sclambda_{\scgamma,k}
    \frac{
        1+a_{t,k}^{2}
        -
        2a_{t,k}h_{t,k}
    }{
        1-a_{t,k}^{2}
    } .
    \label{eq:sampling_kappa_b_definition}
\end{equation}
We call $\sczeta_{t,k}$ the \textit{local stiffness} at $\sclambda_k$. For a nonlinear denoiser, the GFT coefficients are coupled. The drift $\vctb_t$, which collects \eqref{eq:generic_reverse_drift} over $k$, depends on the whole vector $\vctx$, and the eigenvalues of its Jacobian $\nabla_{\vctx}\vctb_t$ govern local linear stability. We summarize the scale of this Jacobian by
\begin{equation}
    L_S
    =
    \sup_{t\in[t_0,T]}
    \sup_{\vctx\in\stR_t}
    \left\|
        \nabla_{\vctx}
        \vctb_t(\vctx)
    \right\|_2 ,
    \label{eq:reverse_stiffness_proxy}
\end{equation}
where $\stR_t\subset\fdR^n$ is any set containing the vectors the sampler can reach at time $t$, and we call $L_S$ the reverse-time stiffness. For the spectral graph filter, the Jacobian is $\diag(\sczeta_{t,0},\ldots,\sczeta_{t,n-1})$ and $L_S=\sup_{t\in[t_0,T]}\max_k|\sczeta_{t,k}|$. For an explicit reverse-SDE solver, the local linear stability test uses the eigenvalues of the drift Jacobian scaled by $-\Delta t$. For a given time parametrization, we use $L_S$ as a proxy for the step-size restriction.

MASS makes the reverse-time gain identical at every graph frequency. Substituting \eqref{eq:mas_snr_rate_schedule} and \eqref{eq:mas_snr_channel} gives $c_{t,k}\sclambda_{\scgamma,k}/(1-a_{t,k}^{2})=-\tfrac12\,d\log r_t/dt$, and the supremum of the common gain is $M/2$, where $M=\sup_{t\in[t_0,T]}|d\log r_t/dt|$. Among the target SNRs \eqref{eq:mas_snr_log_trajectory} that join $r_{\max}$ to $r_{\min}$, a linear $u_t$ has the smallest log-SNR slope, $M=\log(r_{\max}/r_{\min})/(T-t_0)$, since a supremum is at least an average. The Jacobian also depends on the sensitivity of the denoiser to $\vctx$, which can grow with the spectral radius through the powers of $\mtL$ used by the GNN. We isolate this sensitivity.

\smallskip
\begin{assumption}[Bounded attenuated Jacobian]
\label{ass:local_bounded_jacobian}
There exists a constant $B<\infty$, independent of the eigenvalues of $\mtL_{\scgamma}$, such that
$\bigl\|\diag(\vca_t)\nabla_{\vctx}\tilde{f}_{\Theta}(\vctx,t)\bigr\|_2\leq B$
for all $\vctx\in\stR_t$ and $t\in[t_0,T]$.
\end{assumption}

\noindent
For the spectral filter $h_{t,k}^{\star}$, which is the optimal linear denoiser under Assumption~\ref{ass:graph_wss}, the matrix in Assumption~\ref{ass:local_bounded_jacobian} is diagonal with entries $a_{t,k}h_{t,k}^{\star}=\snr_{t,k}/(1+\snr_{t,k})<1$, hence one can take $B=1$ for this filter. For a trained GNN, $B$ depends on the trained weights, the attenuations and the set $\stR_t$.

\begin{theorem}[MASS stiffness bound]
\label{thm:mas_snr_stiffnessRe}
Let Assumptions~\ref{ass:simple_spectrum} and~\ref{ass:local_bounded_jacobian} hold, and let $L_S^{\mathrm{MASS}}$ be the reverse-time stiffness \eqref{eq:reverse_stiffness_proxy} under the MASS diffusion rates \eqref{eq:mas_snr_cumulative_schedule}. Then,
\begin{equation}
    L_S^{\mathrm{MASS}}
    \leq
    M(1+B),
    \qquad
    M
    =
    \sup_{t\in[t_0,T]}
    \left|\frac{d}{dt}\log r_t\right| ,
    \label{eq:mas_stiffness_bound}
\end{equation}
where $M$ is the log-SNR slope and $B$ is the constant in Assumption~\ref{ass:local_bounded_jacobian}. 

If, in addition, the endpoints $r_{\max}$ and $r_{\min}$ satisfy \eqref{eq:mass_dynamic_range} with equality for $\scdelta_0=\scdelta\in(0,1/2)$, the target SNR follows \eqref{eq:mas_snr_log_trajectory} with the linear ramp $u_t=(t-t_0)/(T-t_0)$, and
$\max_kp_k\leq\kappa\min_kp_k$ for some $\kappa\geq1$, then
\begin{equation}
    \!L_S^{\mathrm{MASS}}
    \leq
    \frac{1+B}{T-t_0}
    \!\Big[
        2\log\frac{1\!-\!\scdelta}{\scdelta}
        +\log\kappa
        +\log\cond(\mtL_{\scgamma})\!
    \Big]\!.
    \label{eq:mass_rate_explicit}
\end{equation}
\end{theorem}

\noindent
For fixed $B$, the bound depends on the graph through the log-SNR slope $M$. For comparison, under a shared, nondecreasing diffusion rate, a $\scdelta$-complete horizon with $\snr_{T,n-1}\leq 1/3$, and the optimal linear denoiser $h_{t,k}^{\star}$, the reverse-time stiffness satisfies $L_S^{\mathrm{shared}}\geq\log(1/\scdelta)\cond(\mtL_{\scgamma})/(4T)$ (Proposition~\ref{prop:scalar_stiffness} in Appendix~\ref{app:mas_snr_sampling_theory}). For fixed $\scdelta$, horizon, $B$ and $\kappa$, the two bounds scale as
\begin{equation}
    L_S^{\mathrm{MASS}}
    =
    \bigOh\!\left(\log\cond(\mtL_{\scgamma})\right),
    ~
    L_S^{\mathrm{shared}}
    =
    \Omega\!\left(\cond(\mtL_{\scgamma})\right).
    \label{eq:stiffness_scaling_contrast}
\end{equation}
Under a shared diffusion rate, $\scdelta$-completeness imposes an integrated rate at the lowest graph frequency, which contributes to the stiffness at the highest frequency. Under MASS, the attenuation tolerances instead determine the required span $r_{\max}/r_{\min}$.

\subsection{Algorithms and Complexity}
\label{subsec:mass_algorithms}

Algorithm~\ref{alg:train_mas_snr} trains through the closed-form MASS attenuation $a_{t,k}$ and noise variance $\scnu_{t,k}$ \eqref{eq:mas_snr_channel}, and Algorithm~\ref{alg:sampling_mas_snr} integrates the reverse SDE in the graph Fourier domain by Euler--Maruyama with exactly integrated noise variance at each step. Both use the floored estimate $\widehat p_k$ and $\widehat g_k=\sclambda_{\scgamma,k}\widehat p_k/\scsigma^{2}$. The sampler uses a diagonal Gaussian approximation to the terminal marginal and returns $\vcx_{t_0}$.

Table~\ref{tab:mass_complexity} reports the computational cost of GAD-MASS, where $\mathcal{C}^{\mathrm{train}}_{\Theta}$ and $\mathcal{C}^{\mathrm{eval}}_{\Theta}$ denote the cost of one training pass and of one denoiser evaluation, respectively. Two operations are performed only once. The first is the eigendecomposition of $\mtL_{\scgamma}$, which GAD requires anyway. The second is the estimation of the power spectrum $\widehat p_k$, which costs $\bigOh(Nn^{2})$ and is dominated by the eigendecomposition whenever $N\lesssim n$. After this, MASS only adds the frequency-wise attenuation $a_{t,k}$ and noise variance $\scnu_{t,k}$ at each diffusion time. This is an elementwise operation of cost $\bigOh(n)$, which is negligible compared with the $\bigOh(n^{2})$ GFT and inverse GFT that the algorithms perform at every step. We finally note that MASS adds no trainable parameters.

\begin{algorithm}[t]
\caption{Training with GAD-MASS}
\label{alg:train_mas_snr}
\small
\begin{algorithmic}[1]

\Require Dataset $\stD$, Laplacian $\mtL$ with graph Fourier basis $\mtV$ and
shifted eigenvalues $\{\sclambda_{\scgamma,k}\}$, noise scale $\scsigma$,
power floor $p_{\min}$, target SNR $r_t$ on $[t_0,T]$, diffusion-time
law $\pi_{\mathrm{tr}}$ on $[t_0,T]$, denoiser $f_{\Theta}$, learning
rate $\eta$, minibatch size $N_{\mathrm b}$.
\Ensure Trained parameters $\Theta^\star$.

\State Estimate $\widehat p_k$ from \eqref{eq:mas_snr_power_estimate} and set
$\widehat g_k\gets\sclambda_{\scgamma,k}\widehat p_k/\scsigma^{2}$ for all $k$.

\While{not converged}

    \State Draw minibatch
    $\{\vcx_0^{(b)}\}_{b=1}^{N_{\mathrm b}}\sim\stD$
    and $t\sim\pi_{\mathrm{tr}}$.

    \State $a_{t,k}\gets\sqrt{\dfrac{r_t}{r_t+\widehat g_k}}$,
    \quad
    $\scnu_{t,k}\gets\dfrac{\widehat p_k}{r_t+\widehat g_k}$,
    \quad for all $k$.

    \State $\vctx_0^{(b)}\gets\mtV^{\Tr}\vcx_0^{(b)}$,
    \quad
    $\vcz^{(b)}\sim\gaussDist(\vcZeros,\mtI)$.

    \State $\sctx_{t,k}^{(b)}
    \gets
    a_{t,k}\sctx_{0,k}^{(b)}
    +\sqrt{\scnu_{t,k}}\,\scz_k^{(b)}$ for all $k$,
    \quad
    $\vcx_t^{(b)}\gets\mtV\vctx_t^{(b)}$.

    \State $\vchx_0^{(b)}
    \gets
    f_{\Theta}(\vcx_t^{(b)},\mtL,t)$.

    \State $\widehat{\mathcal L}_{\mathrm{den}}(\Theta)
    \gets
    \dfrac{1}{N_{\mathrm b}}
    \sum_{b=1}^{N_{\mathrm b}}
    \|\vcx_0^{(b)}-\vchx_0^{(b)}\|_2^2$.

    \State $\Theta\gets
    \Theta-\eta\nabla_{\Theta}
    \widehat{\mathcal L}_{\mathrm{den}}(\Theta)$.

\EndWhile

\State \Return $\Theta^\star\gets\Theta$

\end{algorithmic}
\end{algorithm}

\begin{algorithm}[t]
\caption{Sampling with GAD-MASS}
\label{alg:sampling_mas_snr}
\small
\begin{algorithmic}[1]

\Require Laplacian $\mtL$ with graph Fourier basis $\mtV$, MASS parameters
$\{\widehat p_k,\widehat g_k\}$, noise scale $\scsigma$, target SNR
$r_t$, reverse grid $t_0<\cdots<t_{N_S}=T$ with $t_0>0$, trained denoiser
$f_{\Theta^\star}$.
\Ensure Sample $\vcx_{t_0}$, produced in $N_S$ reverse steps and $N_S$
denoiser evaluations.

\State Compute $\scbc_{t_i,k}$ from
\eqref{eq:mas_snr_cumulative_schedule} for all $i$ and $k$.

\State Sample $\sctx_{T,k}\sim\gaussDist(0,\scrho_{T,k})$ for all $k$,
with $\scrho_{T,k}$ from \eqref{eq:mas_snr_channel}.

\For{$i=N_S,\ldots,1$}

    \State $t\gets t_i$, \quad $t^-\gets t_{i-1}$, \quad
    $\Delta t_i\gets t-t^-$

    \State $a_{t,k}\gets
    \sqrt{\dfrac{r_t}{r_t+\widehat g_k}}$,
    \quad
    $\Delta\scbc_{i,k}\gets
    \scbc_{t,k}-\scbc_{t^-,k}$,
    \quad for all $k$

    \State $\vchx_0\gets
    f_{\Theta^\star}(\mtV\vctx_t,\mtL,t)$,
    \quad
    $\widehat{\vctx}_0\gets\mtV^{\Tr}\vchx_0$

    \State $\sctb_{t,k}\gets
    -\dfrac{1}{2}
    \left.\dfrac{d}{ds}\log r_s\right|_{s=t}
    \left[
    (1+a_{t,k}^{2})\sctx_{t,k}
    -2a_{t,k}\widehat{\sctx}_{0,k}
    \right]$ for all $k$

    \State $\vcz_i\sim\gaussDist(\vcZeros,\mtI)$

    \State $\sctx_{t^-,k}\gets
    \sctx_{t,k}
    -\sctb_{t,k}\Delta t_i
    +\scsigma\sqrt{2\Delta\scbc_{i,k}}\,\scz_{i,k}$ for all $k$

\EndFor

\State \Return $\vcx_{t_0}\gets\mtV\vctx_{t_0}$

\end{algorithmic}
\end{algorithm}

\begin{table}[t]
\centering
\caption{Computational cost of GAD-MASS.}
\label{tab:mass_complexity}
\small
\setlength{\tabcolsep}{4pt}
\begin{tabular}{ll}
\toprule
Operation & Cost \\
\midrule
Eigendecomposition of $\mtL_{\scgamma}$ (once)
    & $\bigOh(n^{3})$ \\
Power spectrum $\widehat p_k$ (once)
    & $\bigOh(Nn^{2})$ \\
Attenuation and noise variance (per $t$)
    & $\bigOh(n)$ \\
Training iteration
    & $\bigOh(N_{\mathrm{b}}n^{2})
       +\mathcal{C}^{\mathrm{train}}_{\Theta}$ \\
Reverse step
    & $\bigOh(n^{2})+\mathcal{C}^{\mathrm{eval}}_{\Theta}$ \\
\bottomrule
\end{tabular}
\end{table}

\section{Numerical Analysis}
\label{sec:numerical_analysis}

In this section, we test our theoretical results on synthetic data in Section~\ref{subsec:experimental_protocol} and evaluate GAD with MASS on three real-data case studies, namely collaborative recommendation, traffic imputation and probabilistic spatio-temporal forecasting, in Section~\ref{subsec:downstream_graph_diffusion_tasks}. Overall, GAD with MASS matches or improves the state of the art in all three tasks with a fraction of the reverse steps.

\subsection{Synthetic Experiments}
\label{subsec:experimental_protocol}

We test Theorem~\ref{thm:mass_forward_channel}, Proposition~\ref{prop:scalar_stiffness} and Theorem~\ref{thm:mas_snr_stiffnessRe} on a synthetic graph and synthetic data.

\smallskip
\noindent\textbf{Setup.}
We use a random geometric $8$-nearest-neighbor graph with $n=128$ nodes and Gaussian edge weights. The Gaussian weights make the eigenvalues distinct, as Assumption~\ref{ass:simple_spectrum} requires. We draw two Gaussian datasets of $10{,}000$ training, $2{,}000$ validation and $2{,}000$ test signals.
\begin{itemize}
\item The stationary signal satisfies Assumption~\ref{ass:graph_wss} and tests MASS where the theory is exact. Its GFT coefficients are independent, $\sctx_{0,k}\sim\gaussDist(0,p_k)$, with $p_k\propto(0.05+\sclambda_k/\sclambda_{n-1})^{-3/2}+24.8$ scaled to unit mean power. The power-law term concentrates the power at the low graph frequencies, and the offset $24.8$ raises the share of the highest quarter of the graph frequencies to $20\%$, so that the metrics are sensitive to the graph frequencies that a shared diffusion rate loses first \eqref{eq:snr_ratio}. The power spectrum uses the eigenvalues of $\mtL$, not of $\mtL_{\scgamma}$, so that a change of the shift does not change the data. It is much flatter than the prior power $\scsigma^{2}/\sclambda_{\scgamma,k}$, hence $g_k$ varies across graph frequencies and MASS does not reduce to an equal attenuation \eqref{eq:mas_snr_channel}.

\item The block-covariance signal has four groups of $32$ nodes with unit variance and a correlation of $0.8$ within a group. Its covariance is specified in the vertex domain, hence it does not commute with $\mtL_{\scgamma}$ and violates Assumption~\ref{ass:graph_wss}. 
\end{itemize}

\smallskip
\noindent\textbf{Forward processes.}
We compare MASS \eqref{eq:mas_snr_cumulative_schedule} with three baselines. FCPS, the floor-constrained polynomial schedule of our preliminary work~\cite{rozada2025graphawarediffusionsignalgeneration}, is the shared GAD diffusion rate that MASS replaces, $\scbc_t=c_{\min}t+(\scbc_T-c_{\min}T)(t/T)^{\alpha+1}$. Cosine is the shared GAD diffusion rate whose integrated rate follows the logarithm of the signal fraction of the cosine schedule~\cite{nichol2021improved}, with offset $s_{\mathrm c}$ and endpoint truncation $t_{\mathrm c}$. Both have a nondecreasing diffusion rate $c_t$, as Proposition~\ref{prop:scalar_stiffness} assumes. The isotropic variance-preserving (VP) baseline~\cite{song2021scorebasedgenerativemodelingstochastic}, the forward process of the downstream pipelines, has the cosine signal fraction in every direction and no Laplacian in its forward process, and keeps the graph-aware denoiser. Every forward process retains at most a fraction $\scdelta=10^{-3}$ of the clean power at $T$ (Definition~\ref{def:delta_complete}), and the shared diffusion rates use $\scbc_T=\log(1/\scdelta)/(2\scgamma)$. MASS sets $r_{\max}=g_{\max}(1-\scdelta_0)/\scdelta_0$ and $r_{\min}=g_{\min}\scdelta/(1-\scdelta)$, the endpoint conditions of Section~\ref{subsec:mass_schedule} at equality, and uses the cosine ramp $u_t=\tfrac12\{1-\cos(\pi t/T)\}$ in its target SNR \eqref{eq:mas_snr_log_trajectory}.

\smallskip
\noindent\textbf{Tuning.}
We tune each forward process separately, because the shift acts differently on each. Per dataset, we train $12$ configurations of each graph-aware forward process, three shifts $\scgamma\in\{10^{-3},3{\times}10^{-3},10^{-2}\}$ times four settings of its shape parameters, namely $\scdelta_0$ for MASS, $\alpha$ and $c_{\min}$ for FCPS, and $t_{\mathrm c}$ for Cosine. The isotropic VP baseline has no shift and uses $12$ settings of $s_{\mathrm c}$ and $t_{\mathrm c}$. We select the configuration with the lowest validation sliced Wasserstein distance (SWD) at $10$ DPM-Solver++2M steps, without using the graph metrics, and score it once on the test split. The grid stops at $10^{-2}$, where the shift already moves $\sclambda_1$ by $26\%$ and brings the forward process closer to the isotropic VP baseline. Both datasets have unit RMS per node, and we set $\scsigma$ to the RMS of the centered training signals, i.e., $\scsigma=1$, except in the ablation.

\smallskip
\noindent\textbf{Denoiser and evaluation.}
The denoiser is a time-conditioned ChebNet~\cite{defferrard2016convolutional} with four residual blocks of Chebyshev order three and $64$ hidden units that predicts the clean signal. It trains for $40$ epochs, and we keep the checkpoint with the lowest validation denoising error. Within a dataset, all forward processes share the graph, the splits, the denoiser, its initialization, and the minibatch and noise draws of training. Our samplers step on a grid uniform in $t$ that stops at $t_0=10^{-3}$, where $r_{t_0}$ lies within $0.01\%$ of $r_{\max}$. The cosine ramp makes each step advance the log-SNR least at both ends of the horizon, where short steps reduce the discretization error~\cite{karras2022elucidating}. We report mean and standard deviation over five paired seeds of initialization, training and sampling. In all tables, bold and underline mark the best and second-best methods, and a method evaluated at several numbers of reverse steps counts once, with its best entry. SWD compares $1{,}024$ generated with $1{,}024$ test signals, averaged over five draws of $128$ random projections.

\begin{figure*}[t]
\centering
\includegraphics[height=3.45cm]{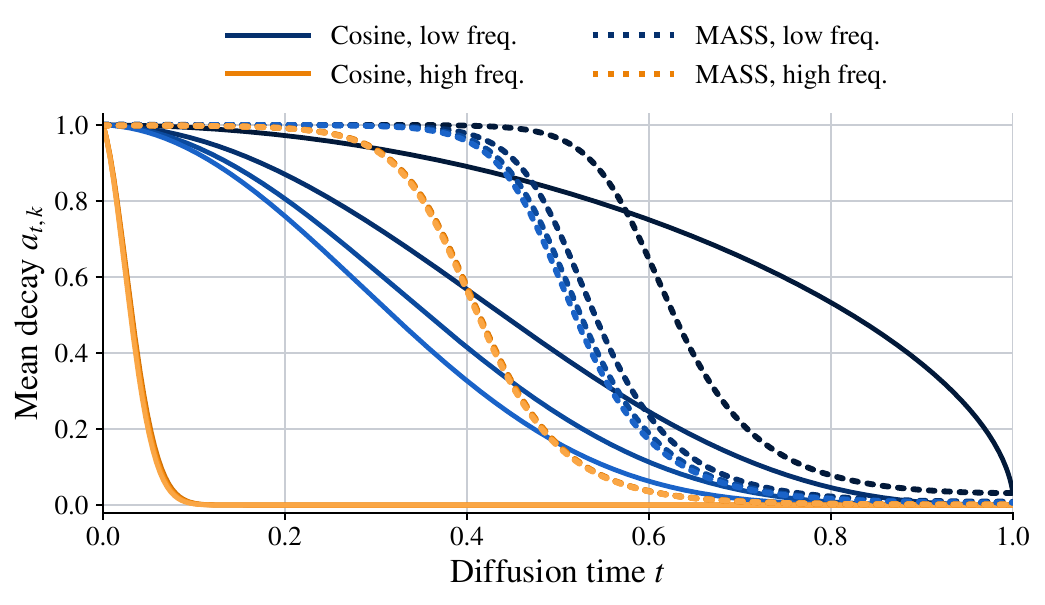}
\hfill
\includegraphics[height=3.45cm]{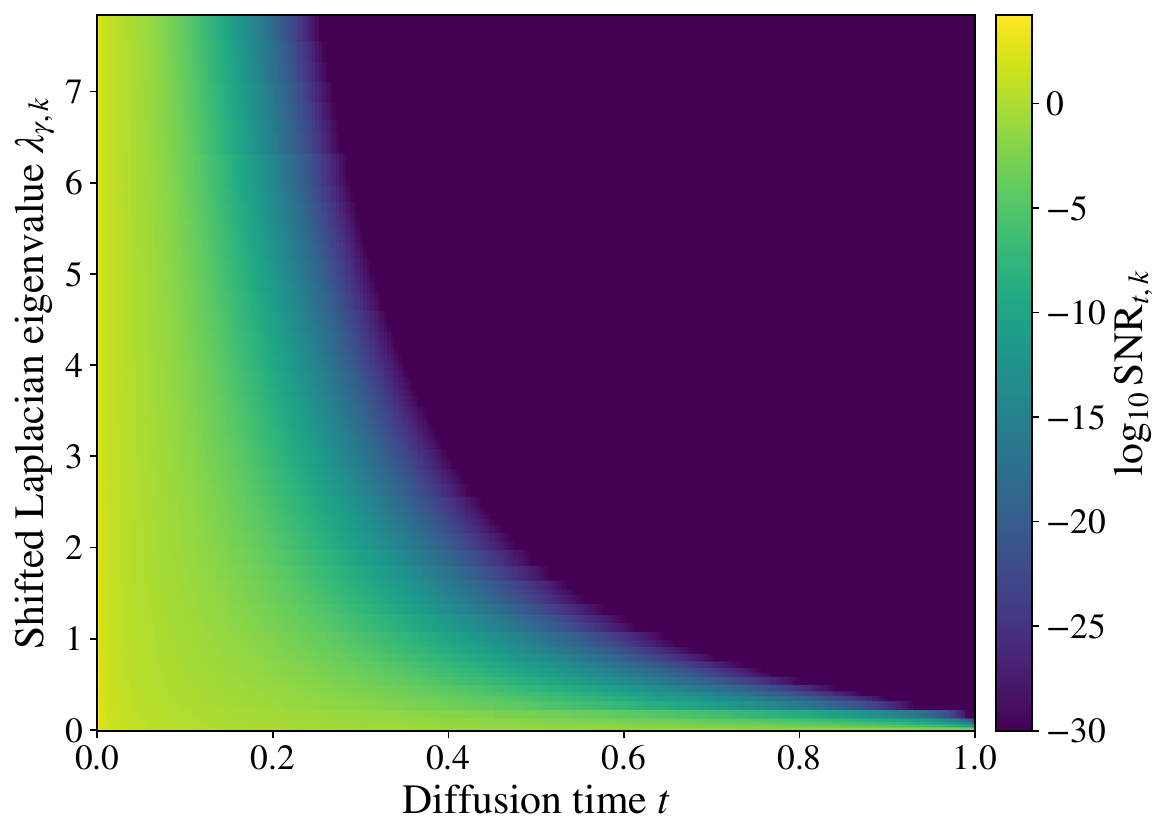}
\hfill
\includegraphics[height=3.45cm]{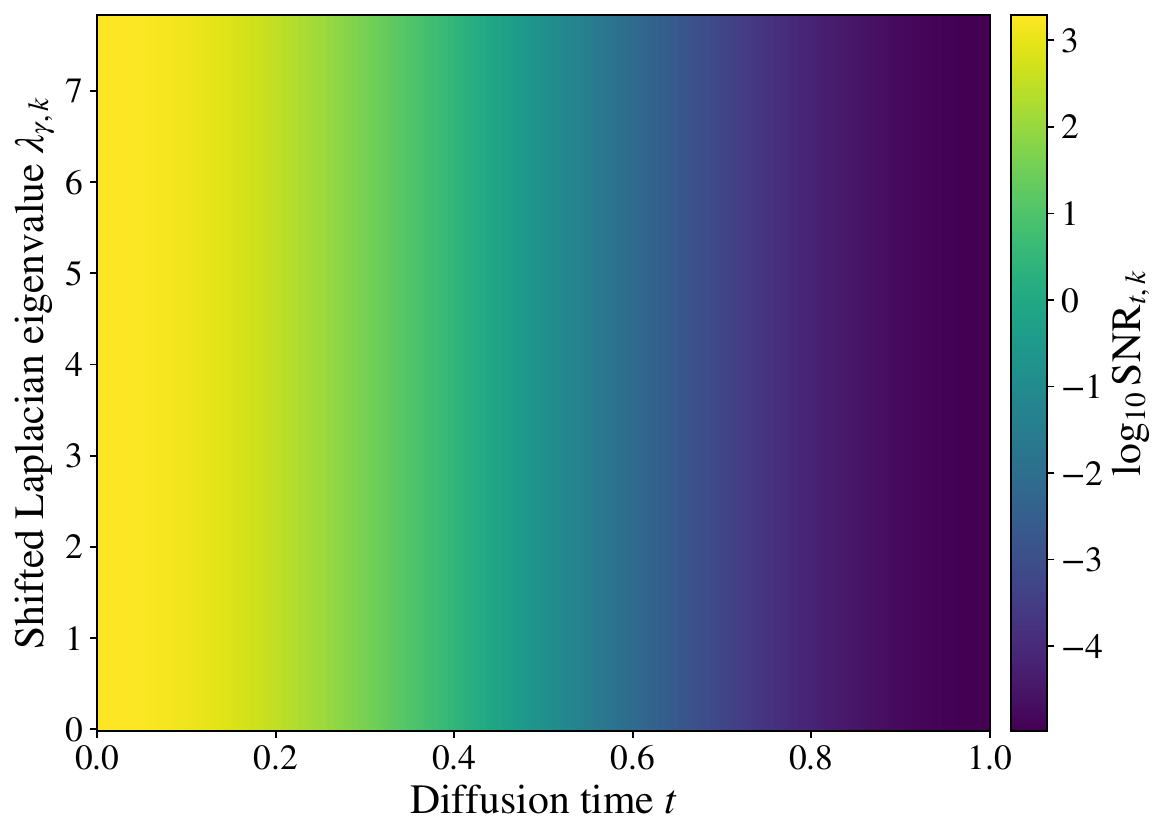}

\vspace{0.45em}

\includegraphics[height=3.45cm]{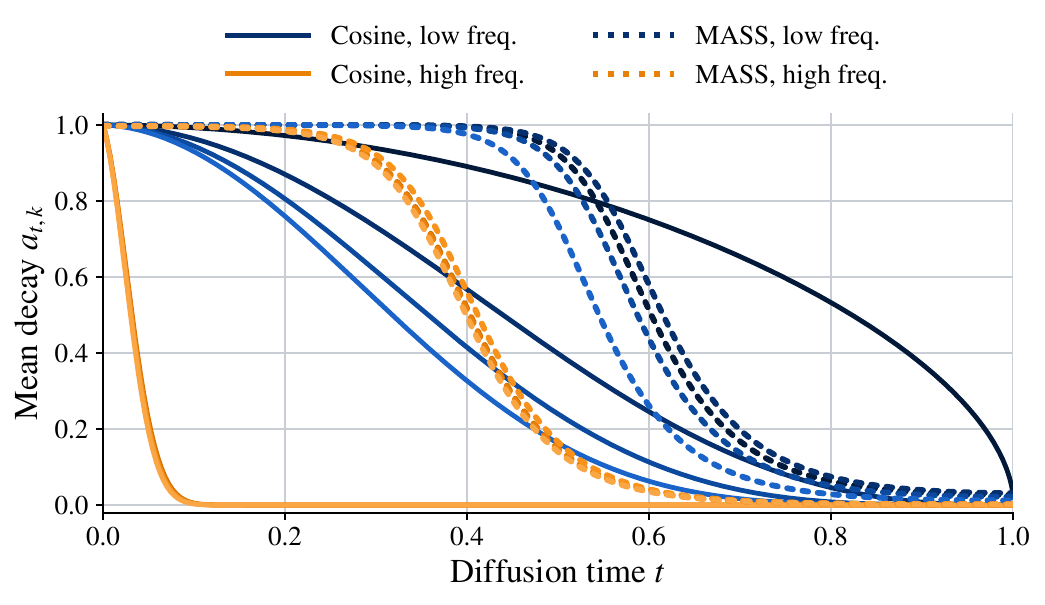}
\hfill
\includegraphics[height=3.45cm]{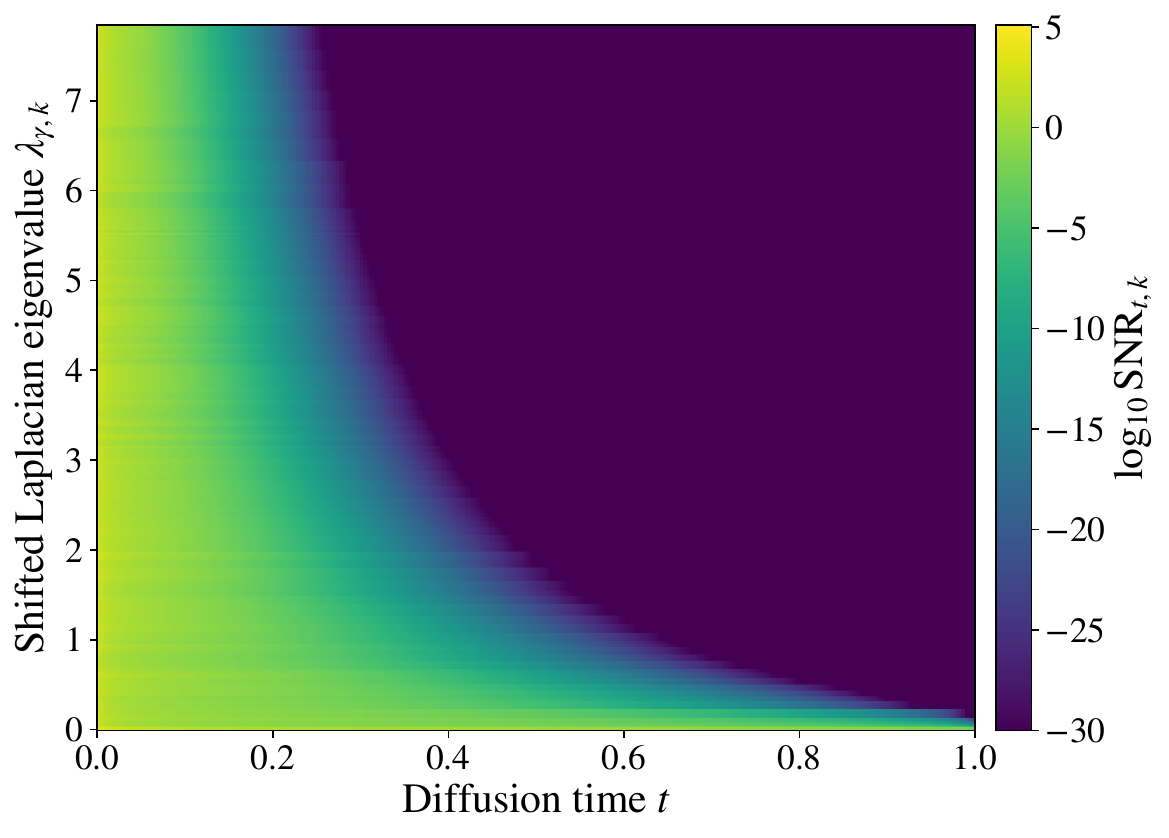}
\hfill
\includegraphics[height=3.45cm]{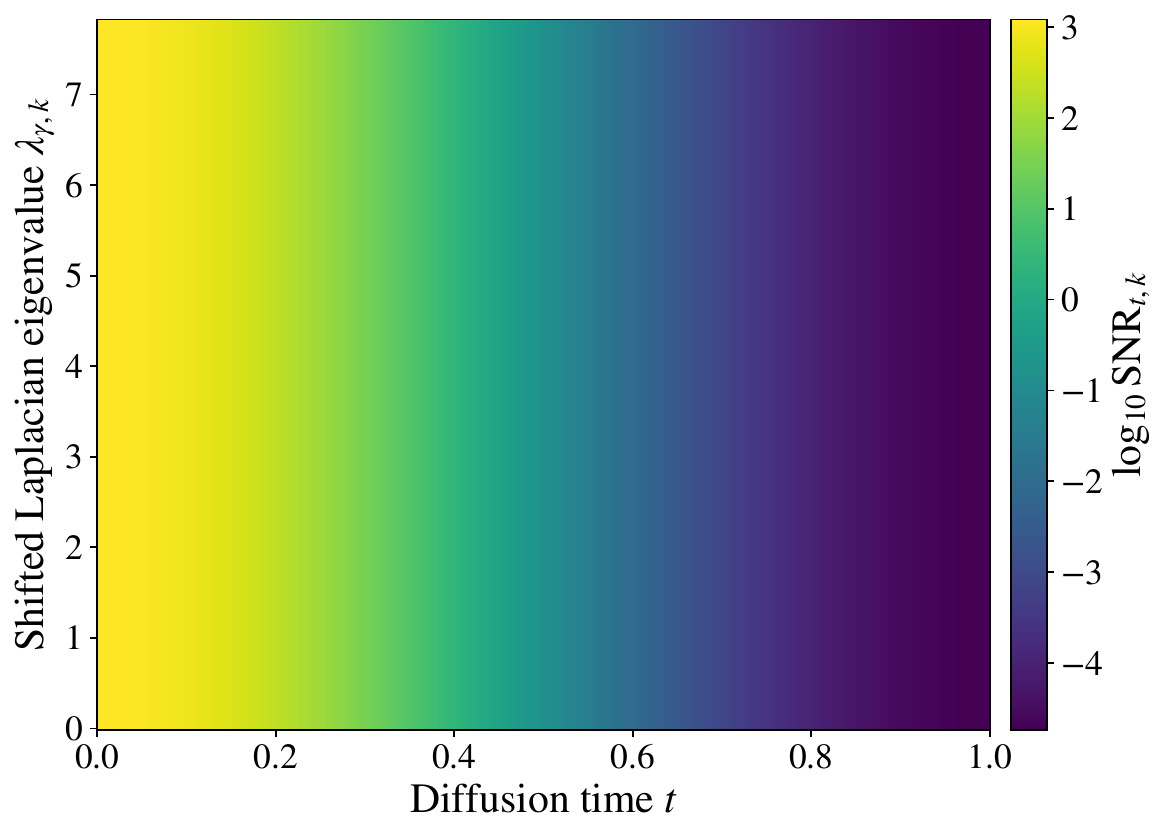}
\caption{\textbf{Frequency-wise attenuation and SNR under a shared and a frequency-dependent diffusion rate}, for the forward processes selected in Table~\ref{tab:reverse_sampling_efficiency}. Top row stationary, bottom row block covariance. Left: attenuation $a_{t,k}$ of the four lowest (blue) and four highest (orange) graph frequencies under Cosine (solid) and MASS (dotted). Center and right: $\log_{10}\snr_{t,k}$ over diffusion time and graph frequency under Cosine, clipped at $-30$, and under MASS, on its own color range.}
\label{fig:forward_spectral_diagnostics}
\end{figure*}

\begin{table*}[t]
\centering
\caption{\textbf{Stiffness induced by the diffusion rate.} $L_S^{(0)}$ from \eqref{eq:schedule_only_stiffness_diagnostic}, and its MASS bound $M$ (Theorem~\ref{thm:mas_snr_stiffnessRe}, $B=0$). $^{*}$ and $^{\dagger}$ mark the shifts selected for MASS and for the shared diffusion rates.}
\label{tab:forward_snr_stiffness_all}
\scriptsize
\setlength{\tabcolsep}{4.5pt}
\renewcommand{\arraystretch}{1.08}
\begin{tabular}{llccccccc}
\toprule
& & & \multicolumn{3}{c}{$L_S^{(0)}\downarrow$} & & \multicolumn{2}{c}{Reduction $L_S^{(0)}/L_S^{(0),\mathrm{MASS}}\uparrow$} \\
\cmidrule(lr){4-6}\cmidrule(lr){8-9}
Dataset & $\scgamma$ & $\cond(\mtL_{\scgamma})$ & FCPS & Cosine & MASS & $M$ & FCPS & Cosine \\
\midrule
\multirow{3}{*}{Stationary}
& $10^{-3}$ & $7687$ & $\underline{1.33\times10^{5}}$ & $4.10\times10^{6}$ & $\mathbf{31.25}$ & $31.64$ & $4.25\times10^{3}$ & $1.31\times10^{5}$ \\
& $3{\times}10^{-3}{}^{*}$ & $2563$ & $\underline{4.43\times10^{4}}$ & $1.37\times10^{6}$ & $\mathbf{29.36}$ & $29.91$ & $1.51\times10^{3}$ & $4.67\times10^{4}$ \\
& $10^{-2}{}^{\dagger}$ & $770$ & $\underline{1.33\times10^{4}}$ & $4.11\times10^{5}$ & $\mathbf{27.24}$ & $28.02$ & $4.88\times10^{2}$ & $1.51\times10^{4}$ \\
\midrule
\multirow{3}{*}{Block covariance}
& $10^{-3}{}^{*}$ & $7687$ & $\underline{2.39\times10^{5}}$ & $4.10\times10^{6}$ & $\mathbf{27.54}$ & $28.29$ & $8.68\times10^{3}$ & $1.49\times10^{5}$ \\
& $3{\times}10^{-3}$ & $2563$ & $\underline{7.97\times10^{4}}$ & $1.37\times10^{6}$ & $\mathbf{27.46}$ & $28.22$ & $2.90\times10^{3}$ & $4.99\times10^{4}$ \\
& $10^{-2}{}^{\dagger}$ & $770$ & $\underline{2.39\times10^{4}}$ & $4.11\times10^{5}$ & $\mathbf{27.18}$ & $27.98$ & $8.80\times10^{2}$ & $1.51\times10^{4}$ \\
\bottomrule
\end{tabular}
\end{table*}

\begin{table*}[t]
\centering
\caption{\textbf{Sampling at a fixed number of reverse steps.} SWD on the test split. \textbf{(a)} Selection point, $10$ DPM-Solver++2M steps. \textbf{(b)} FCPS and MASS at $25$ steps under two solvers.}
\label{tab:reverse_sampling_efficiency}
\scriptsize
\setlength{\tabcolsep}{4.5pt}
\renewcommand{\arraystretch}{1.08}
\begin{tabular}{llrcc}
\toprule
Solver & Forward process & Steps & Stationary SWD$\downarrow$ & Block covariance SWD$\downarrow$ \\
\midrule
\multicolumn{5}{l}{\textit{(a) Selection point}}\\
\addlinespace[2pt]
DPM-Solver++2M & Isotropic VP & 10 & $\underline{0.0767 \pm 0.0102}$ & $0.1887 \pm 0.0288$ \\
 & FCPS & 10 & $0.0785 \pm 0.0266$ & $\underline{0.1836 \pm 0.0202}$ \\
 & Cosine & 10 & $0.2843 \pm 0.0114$ & $0.2966 \pm 0.0219$ \\
 & \textbf{MASS (ours)} & 10 & $\mathbf{0.0630 \pm 0.0037}$ & $\mathbf{0.1611 \pm 0.0095}$ \\
\cmidrule(lr){2-5}
 & \textit{Improvement over best baseline} & -- & $17.9\%$ & $12.3\%$ \\
\midrule
\multicolumn{5}{l}{\textit{(b) Forward process vs.\ solver: FCPS vs.\ MASS at 25 steps}}\\
\addlinespace[2pt]
DDIM & FCPS & 25 & $\underline{0.1568 \pm 0.0170}$ & $\underline{0.1821 \pm 0.0242}$ \\
 & MASS (ours) & 25 & $\mathbf{0.1067 \pm 0.0097}$ & $\mathbf{0.1684 \pm 0.0096}$ \\
\addlinespace[2pt]
DPM-Solver++2M & FCPS & 25 & $\underline{0.0771 \pm 0.0203}$ & $\underline{0.1719 \pm 0.0099}$ \\
 & MASS (ours) & 25 & $\mathbf{0.0654 \pm 0.0048}$ & $\mathbf{0.1682 \pm 0.0109}$ \\
\bottomrule
\end{tabular}
\end{table*}

\begin{table*}[t]
\centering
\caption{\textbf{Generation quality across graph frequencies.} Errors of the samples of Table~\ref{tab:reverse_sampling_efficiency}(a) against the test signals.}
\label{tab:synthetic_mass_sampling_efficiency}
\scriptsize
\setlength{\tabcolsep}{5.5pt}
\renewcommand{\arraystretch}{1.12}
\begin{tabular}{llrrrr}
\toprule
Dataset & Forward process & Cov.\ err.\ $\downarrow$
& Power-spectrum err.\ $\downarrow$ & Band-ratio err.\ $\downarrow$ & Smoothness err.\ $\downarrow$ \\
\midrule
\multirow{3}{*}{Stationary}
& Isotropic VP & $\underline{0.4787 \pm 0.0166}$ & $\underline{0.1322 \pm 0.0458}$ & $\underline{0.0305 \pm 0.0267}$ & $\underline{57.92 \pm 25.88}$ \\
& Cosine & $0.8434 \pm 0.0132$ & $0.8079 \pm 0.0066$ & $1.0359 \pm 0.0043$ & $354.52 \pm 1.84$ \\
& \textbf{MASS (ours)} & $\mathbf{0.4354 \pm 0.0119}$ & $\mathbf{0.0800 \pm 0.0125}$ & $\mathbf{0.0201 \pm 0.0138}$ & $\mathbf{23.93 \pm 9.66}$ \\
\midrule
\multirow{3}{*}{Block covariance}
& Isotropic VP & $0.8727 \pm 0.0102$ & $\underline{0.4894 \pm 0.0463}$ & $\underline{0.0929 \pm 0.0402}$ & $\mathbf{104.79 \pm 34.58}$ \\
& Cosine & $\underline{0.8599 \pm 0.0248}$ & $0.7608 \pm 0.0697$ & $0.9014 \pm 0.0312$ & $322.71 \pm 4.21$ \\
& \textbf{MASS (ours)} & $\mathbf{0.8302 \pm 0.0051}$ & $\mathbf{0.2562 \pm 0.0221}$ & $\mathbf{0.0497 \pm 0.0222}$ & $\underline{116.10 \pm 2.63}$ \\
\bottomrule
\end{tabular}
\end{table*}

\begin{figure*}[t]
\centering
\includegraphics[width=0.485\textwidth]{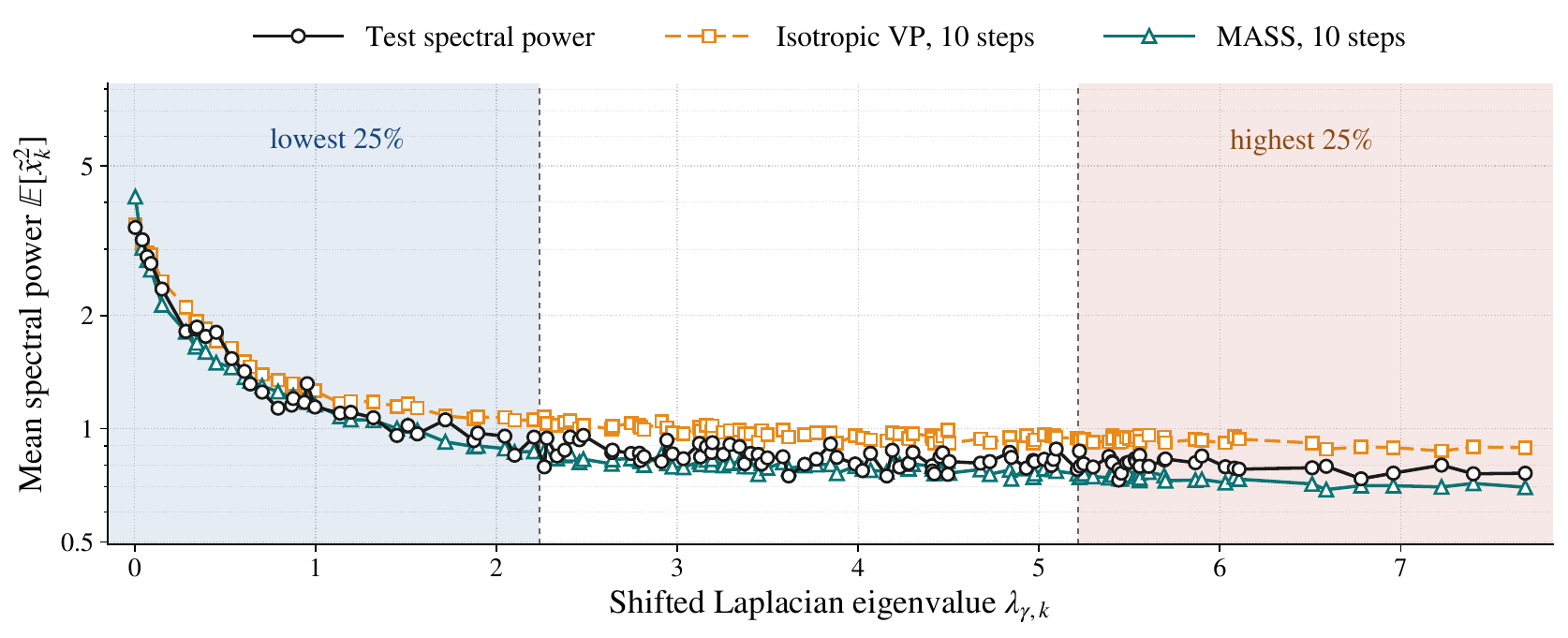}
\hfill
\includegraphics[width=0.485\textwidth]{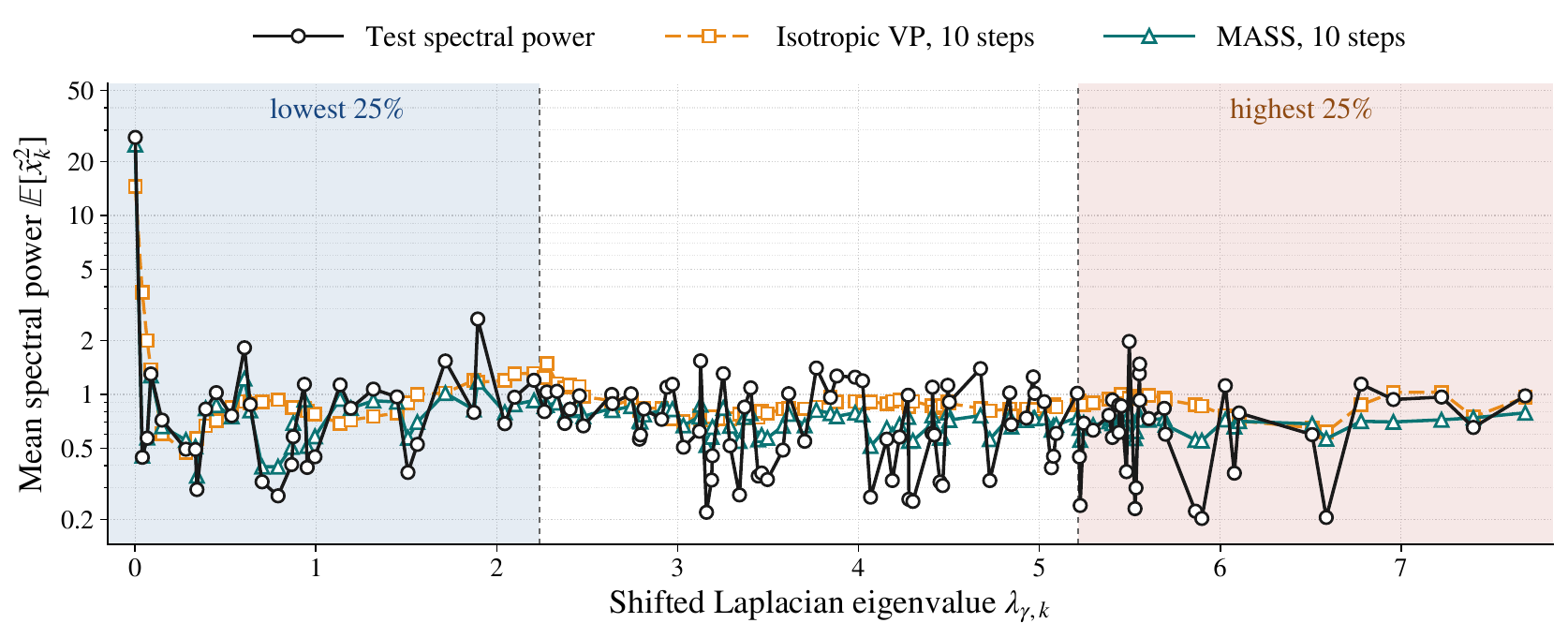}
\caption{\textbf{Frequency-wise power of generated signals.} Frequency-wise power $\xp[\sctx_k^{2}]$ against $\sclambda_{\scgamma,k}$ for the test signals, the isotropic VP baseline and MASS at $10$ DPM-Solver++2M steps, the setting of Table~\ref{tab:synthetic_mass_sampling_efficiency}. Left: stationary. Right: block covariance. Shading marks the lowest and highest quarter of the graph frequencies.}
\label{fig:spectral_power_matching}
\end{figure*}

\smallskip
\noindent\textbf{Stiffness induced by the diffusion rate.}
The reverse-time stiffness $L_S$ depends on the trained denoiser. To isolate the part that the diffusion rate creates, we set the frequency response of the denoiser to zero in the local stiffness \eqref{eq:sampling_kappa_b_definition},
\begin{equation}
    L_S^{(0)}
    =
    \sup_{t,k}
    \left.|\sczeta_{t,k}|\right|_{h_{t,k}=0},
    \label{eq:schedule_only_stiffness_diagnostic}
\end{equation}
which needs no training. We evaluate all diffusion rates on the same operator at each shift. At $h_{t,k}=0$, the theory gives $L_S^{(0),\mathrm{shared}}\geq c_T\sclambda_{\scgamma,\max}$ and $L_S^{(0),\mathrm{MASS}}\leq M$. For the cosine ramp, $M=\tfrac{\pi}{2}\log(r_{\max}/r_{\min})/T$, a factor $\pi/2$ above that of the linear ramp, which keeps the logarithmic dependence on $\cond(\mtL_{\scgamma})$.

Table~\ref{tab:forward_snr_stiffness_all} confirms both bounds. Under the shared diffusion rates, $L_S^{(0)}\approx c_T\sclambda_{\scgamma,\max}$, and under MASS it lies within $3\%$ below $M$, which makes MASS between $490$ and $1.5{\times}10^{5}$ times less stiff. Over the tenfold range of $\cond(\mtL_{\scgamma})$, the stiffness of the shared diffusion rates grows tenfold, whereas that of MASS grows by at most $15\%$, since it depends on the shift only through the span $r_{\max}/r_{\min}$.
Fig.~\ref{fig:forward_spectral_diagnostics} shows the attenuation and the SNR behind this reduction. Under Cosine, the four highest graph frequencies are lost within the first tenth of the horizon, whereas the lowest keeps $a_{t,k}>0.8$ beyond its middle, and the informative band of Proposition~\ref{prop:informative_band} shifts from high to low graph frequencies. Under MASS, all eight graph frequencies decay in the middle of the horizon, and every graph frequency follows the common target SNR $r_t$.

\smallskip
\noindent\textbf{Sampling at a fixed number of reverse steps.}
We compare sample quality at fixed step counts using DPM-Solver++2M and DDIM, which solve the probability-flow ODE. At the selection point of Table~\ref{tab:reverse_sampling_efficiency}(a), $10$ steps of the second-order DPM-Solver++2M~\cite{lu2022dpmpp}, MASS improves on the best baseline by $17.9\%$ and $12.3\%$, and beats every baseline in all five seeds on the stationary data and in at least four of five on the block-covariance data. Panel~(b) tests whether a better solver can replace a better forward process, by comparing MASS with FCPS, the strongest graph-aware baseline, at $25$ steps of the first-order DDIM~\cite{song2021denoising} and of DPM-Solver++2M. MASS remains best under both solvers, and under DDIM the SWD of FCPS is $1.47$ and $1.08$ times that of MASS. Forward-process design and fast solvers are therefore complementary.

\smallskip
\noindent\textbf{Generation quality across graph frequencies.}
The spectral imbalance affects how the energy of a sample is distributed over the graph frequencies. We measure this on the samples of Table~\ref{tab:reverse_sampling_efficiency}(a) with four errors against the test signals. The covariance error is the relative Frobenius error of the node covariance, the power-spectrum error is the relative $\ell_1$ error of the frequency-wise power $\xp[\sctx_k^{2}]$, the band-ratio error is the $\ell_1$ distance between the mean energy fractions of the lowest, middle and highest third of the graph frequencies, and the smoothness error is the Wasserstein-1 distance between the distributions of $\vcx^{\Tr}\mtL\vcx$. In Table~\ref{tab:synthetic_mass_sampling_efficiency}, MASS attains the lowest mean on every error except the smoothness error on the block-covariance data, where the isotropic VP baseline is lower with a spread that covers MASS. On the stationary data, MASS lowers the power-spectrum, band-ratio and smoothness errors of Cosine by factors of $10$, $50$ and $15$.

Fig.~\ref{fig:spectral_power_matching} compares the frequency-wise power of MASS with that of the isotropic VP baseline, the strongest baseline. On the stationary data, the isotropic VP baseline overshoots the upper half of the graph frequencies by about $15\%$, whereas MASS stays within $8\%$ of the target everywhere. On the block-covariance data, MASS recovers the dominant GFT coefficient at $\sclambda_0=0$ within $9\%$, whereas the isotropic VP baseline recovers only half of it and overshoots the next two graph frequencies by factors of $8$ and $3.5$.

\smallskip
\noindent\textbf{Ablation of the forward-process parameters.}
MASS sets the target SNR and leaves two forward-process parameters free, the shift $\scgamma$ and the noise scale $\scsigma$. Table~\ref{tab:gamma_sigma_ablation} varies one at a time on the stationary data around the selected MASS setting. Every row recalibrates the endpoints from the same tolerances, so that every row satisfies the same attenuation tolerances at both ends of the horizon.

On these data, $g_{\min}=\scgamma p_0/\scsigma^{2}$ is attained at $\sclambda_0=0$, so the span $r_{\max}/r_{\min}$ grows in proportion to $\cond(\mtL_{\scgamma})$. By its definition, $M$ then grows by approximately $\tfrac{\pi}{2T}\log33$, i.e., $5.5$, over the upper block, which the table matches, whereas a shared diffusion rate would grow by a factor of $33$. SWD worsens by about $12\%$ at the two smallest shifts and is flat from $\scgamma=3{\times}10^{-3}$ on, so a larger shift improves MASS only marginally, while it moves every nonconstant eigenvalue by up to $\scgamma/\sclambda_1$, i.e., $7.7\%$ at $3{\times}10^{-3}$ and $25.7\%$ at $10^{-2}$.

The noise scale leaves the attenuation unchanged by construction. Both $g_k$ and the recalibrated endpoints scale as $\scsigma^{-2}$, so $a_{t,k}^{2}$, $M$ and $L_S^{(0)}$ do not depend on $\scsigma$, as the lower block shows. Only the noise level at which the denoiser trains changes, and the metrics move with it by about one standard deviation and without a monotone trend, which supports setting $\scsigma$ without tuning.

\begin{table}[t]
\centering
\caption{\textbf{Ablation of the forward-process parameters.} MASS on the stationary data around the selected setting ($^{*}$). Upper block: shift at $\scsigma=1$. Lower block: noise scale at $\scgamma=3{\times}10^{-3}$.}
\label{tab:gamma_sigma_ablation}
\scriptsize
\setlength{\tabcolsep}{2.5pt}
\renewcommand{\arraystretch}{1.05}
\resizebox{\columnwidth}{!}{%
\begin{tabular}{ccccccc}
\toprule
$\scgamma$ & $\scsigma$ & $\cond(\mtL_{\scgamma})$ & $M$
& $L_S^{(0)}\downarrow$ & SWD$\downarrow$ & Power-spectrum err.\ $\downarrow$ \\
\midrule
$3{\times}10^{-4}$ & $1$ & $25620$ & $33.53$ & $33.27$ & $0.0715\pm0.0074$ & $0.1038\pm0.0291$ \\
$10^{-3}$ & $1$ & $7687$ & $31.64$ & $31.25$ & $0.0699\pm0.0079$ & $0.1106\pm0.0241$ \\
$3{\times}10^{-3}{}^{*}$ & $1$ & $2563$ & $29.91$ & $29.36$ & $0.0630\pm0.0037$ & $0.0797\pm0.0128$ \\
$10^{-2}$ & $1$ & $770$ & $28.02$ & $27.24$ & $0.0630\pm0.0055$ & $0.0685\pm0.0158$ \\
\midrule
$3{\times}10^{-3}$ & $0.25$ & $2563$ & $29.91$ & $29.36$ & $0.0730\pm0.0157$ & $0.1360\pm0.0541$ \\
$3{\times}10^{-3}$ & $0.5$ & $2563$ & $29.91$ & $29.36$ & $0.0645\pm0.0095$ & $0.0724\pm0.0042$ \\
$3{\times}10^{-3}$ & $2$ & $2563$ & $29.91$ & $29.36$ & $0.0693\pm0.0065$ & $0.0946\pm0.0318$ \\
\bottomrule
\end{tabular}}
\end{table}

\subsection{Downstream Tasks}
\label{subsec:downstream_graph_diffusion_tasks}

We replace the isotropic forward process of three diffusion pipelines, for collaborative recommendation, traffic imputation and probabilistic spatio-temporal forecasting, with the MASS attenuation and noise variance \eqref{eq:mas_snr_channel}. We keep their denoiser, conditioning variables, data splits and evaluation code, apart from the task-specific changes described below. To separate the graph operator from the diffusion rate, we also run the shared FCPS diffusion rate~\cite{rozada2025graphawarediffusionsignalgeneration} on the same graph, denoiser and sampler, with its endpoints matched to the MASS attenuations. We call the two variants GAD-MASS and GAD-FCPS. Baselines are as reported in the original works, and the GAD rows report mean$\pm$std over five seeds.

\smallskip
\noindent\textbf{Collaborative recommendation.}
We follow the collaborative filtering setup of CDiff4Rec~\cite{lee2025cdiff4rec}. The history of a user is a binary vector $\vcx_0\in\{0,1\}^{n}$ over $n$ items, and the task is to rank the items absent from the history so that the held-out interactions appear at the top. CDiff4Rec corrupts the history with isotropic noise and recovers it with an MLP aided by the collaborative signals of similar users and review words, which it applies at $50$ diffusion levels. We view the history as a graph signal on an item graph that connects each item to the items with the most similar review profiles, so that the low graph frequencies are preferences shared by such items. We estimate $p_k$ from the training histories, apply the denoiser once, at the $r_{\max}$ end of the target SNR where $a_{t_0,k}^{2}\approx1$, and place $70\%$ of the training-time mass of $\pi_{\mathrm{tr}}$ on the four least corrupted levels to match. Both methods rank the items by the output of the trained denoiser, and we select one checkpoint per seed on validation Recall@20.

Table~\ref{tab:real_recommendation_results} reports Recall@20 and NDCG@20, averaged over the test users. Recall@20 is the fraction of the held-out items among the $20$ top-ranked items, and NDCG@20 also weights each hit by the inverse of $\log_2(1+\text{rank})$, normalized so that a perfect ranking scores $1$. GAD-MASS improves on CDiff4Rec on every dataset and metric, by $5\%$ to $19\%$, with one denoiser evaluation per user against $50$, whereas GAD-FCPS lies within $-4\%$ to $+3\%$ of CDiff4Rec.

\begin{table}[t]
\centering
\caption{\textbf{Collaborative recommendation.} Baselines as reported in~\cite{lee2025cdiff4rec}.}
\label{tab:real_recommendation_results}
\scriptsize
\setlength{\tabcolsep}{2.5pt}
\renewcommand{\arraystretch}{1.05}
\resizebox{\columnwidth}{!}{%
\begin{tabular}{lcccccc}
\toprule
\multirow{2}{*}{Method}
& \multicolumn{2}{c}{Amazon-Game}
& \multicolumn{2}{c}{Yelp}
& \multicolumn{2}{c}{Citeulike-t} \\
\cmidrule(lr){2-3}\cmidrule(lr){4-5}\cmidrule(lr){6-7}
& R@20$\uparrow$ & N@20$\uparrow$ & R@20$\uparrow$ & N@20$\uparrow$ & R@20$\uparrow$ & N@20$\uparrow$ \\
\midrule
BPRMF~\cite{rendle2009bpr}
& 0.1839 & 0.0844 & 0.0918 & 0.0485 & 0.1535 & 0.0792 \\
LightGCN~\cite{he2020lightgcn}
& 0.1899 & 0.0854 & 0.1047 & 0.0553 & 0.1574 & 0.0800 \\
EASE~\cite{steck2019embarrassingly}
& 0.2108 & 0.1012 & 0.1099 & 0.0601 & 0.1591 & 0.0852 \\
MultiVAE~\cite{liang2018variational}
& 0.2181 & 0.0995 & 0.1056 & 0.0548 & 0.1447 & 0.0770 \\
ConVAE~\cite{carraro2022conditioned}
& 0.2192 & 0.0997 & 0.1082 & 0.0564 & 0.1477 & 0.0775 \\
DiffRec~\cite{wang2023diffrec}
& 0.2193 & 0.1034 & 0.1045 & 0.0563 & 0.1491 & 0.0791 \\
CDiff4Rec~\cite{lee2025cdiff4rec}
& $\underline{0.2255}$ & 0.1052 & 0.1145 & 0.0622 & 0.1616 & $\underline{0.0885}$ \\
\midrule
CDiff4Rec + GAD-FCPS
& $0.2216 \pm 0.0058$ & $\underline{0.1056 \pm 0.0023}$
& $\underline{0.1175 \pm 0.0006}$ & $\underline{0.0640 \pm 0.0002}$
& $\underline{0.1618 \pm 0.0005}$ & $0.0849 \pm 0.0005$ \\
CDiff4Rec + GAD-MASS
& $\mathbf{0.2465 \pm 0.0019}$ & $\mathbf{0.1159 \pm 0.0009}$
& $\mathbf{0.1202 \pm 0.0005}$ & $\mathbf{0.0653 \pm 0.0001}$
& $\mathbf{0.1922 \pm 0.0019}$ & $\mathbf{0.1045 \pm 0.0014}$ \\
\bottomrule
\end{tabular}%
}
\end{table}

\smallskip
\noindent\textbf{Traffic imputation.}
We follow the traffic imputation setting of FENCE~\cite{mao2026fence} on PEMS04 and PEMS08, where $307$ and $170$ road sensors record traffic flow every five minutes, with its spatially clustered, temporally contiguous (SC--TC) pattern. This pattern hides communities of neighboring sensors together for one hour with probability $0.8$, so the hidden values must be inferred through the graph from sensors outside the community, and the imputation relies on the learned prior over graph signals. FENCE corrupts the readings with isotropic noise, trains an attention denoiser unconditionally and then conditioned on the observed entries, and samples with $50$ reverse steps under a guidance scale set per sensor cluster by a feedback rule. We view each time slice as a graph signal on the sensors. Since the reverse process regenerates only the hidden entries, $p_k$ blends the clean power with the residual power left on them by the mean over neighboring sensors and time. We evaluate the feedback rule in the graph Fourier basis, where the reverse-step covariance of MASS is diagonal, keep the denoiser, masks and evaluator of FENCE, and add a fine-tuning stage at small numbers of reverse steps. We sample with Euler--Maruyama at $16$ and $25$ steps, take the median of ten samples as the imputation, and select checkpoints on validation at the same step count.

Table~\ref{tab:traffic_fence_sctc_comparison} reports the mean absolute error (MAE), the root mean squared error (RMSE) and the mean absolute percentage error (MAPE, as a fraction) on the hidden test entries. On PEMS08, GAD-MASS improves all three metrics at $16$ steps and further at $25$, where MAE, RMSE and MAPE fall by $6.8\%$, $7.8\%$ and $10.9\%$. On PEMS04, it matches FENCE in RMSE and MAPE at $25$ steps and trails it in MAE by $1.9\%$, with half the reverse steps. GAD-FCPS gives two to four times the MAE of GAD-MASS at both step counts, worse than every baseline. This is consistent with Section~\ref{subsec:mass_effects}, since the largest reverse-time gain under a shared diffusion rate, attained at the highest graph frequency, may not be resolved at these step counts, whereas under MASS it does not grow with $\sclambda_{\scgamma,k}$.

\begin{table}[t]
\centering
\caption{\textbf{Traffic imputation under SC--TC missingness.} Baselines as reported in~\cite{mao2026fence}. NR: unreported number of steps.}
\label{tab:traffic_fence_sctc_comparison}
\scriptsize
\setlength{\tabcolsep}{2.5pt}
\renewcommand{\arraystretch}{1.05}
\resizebox{\columnwidth}{!}{%
\begin{tabular}{lccccccc}
\toprule
\multirow{2}{*}{Method} & \multirow{2}{*}{Steps}
& \multicolumn{3}{c}{PEMS04 SC--TC} & \multicolumn{3}{c}{PEMS08 SC--TC} \\
\cmidrule(lr){3-5}\cmidrule(lr){6-8}
& & MAE$\downarrow$ & RMSE$\downarrow$ & MAPE$\downarrow$ & MAE$\downarrow$ & RMSE$\downarrow$ & MAPE$\downarrow$ \\
\midrule
ASTGNN~\cite{guo2021learning} & -- & 31.70 & 47.76 & 0.212 & 31.82 & 50.76 & 0.220 \\
IGNNK~\cite{wu2021inductive} & -- & 33.32 & 49.12 & 0.214 & 32.25 & 51.35 & 0.231 \\
GCASTN~\cite{peng2023generative} & -- & 30.07 & 47.97 & 0.205 & 30.78 & 49.39 & 0.213 \\
LCR~\cite{chen2024laplacian} & -- & 28.98 & 46.96 & 0.197 & 30.51 & 50.03 & 0.218 \\
ImputeFormer~\cite{nie2024imputeformer} & -- & 29.35 & 46.71 & 0.206 & 29.64 & 48.37 & 0.216 \\
mTAN~\cite{shukla2021multi} & -- & 30.59 & 48.19 & 0.217 & 30.29 & 51.82 & 0.229 \\
CSDI~\cite{tashiro2021csdi} & NR & 28.26 & 45.39 & 0.189 & 28.23 & 48.54 & 0.190 \\
PriSTI~\cite{liu2023pristi} & NR & 28.47 & 45.31 & 0.190 & 35.90 & 48.80 & 0.193 \\
FENCE~\cite{mao2026fence} & 50 & $\mathbf{27.31}$ & $\underline{44.28}$ & $\underline{0.180}$ & 27.29 & 47.78 & 0.175 \\
\midrule
\multirow{2}{*}{FENCE + GAD-FCPS}
& 16 & $119.49 \pm 8.03$ & $159.44 \pm 16.20$ & $1.236 \pm 0.372$ & $74.21 \pm 16.96$ & $104.58 \pm 21.95$ & $0.417 \pm 0.088$ \\
& 25 & $72.99 \pm 3.38$  & $99.96 \pm 5.73$   & $0.639 \pm 0.238$ & $49.86 \pm 2.24$  & $72.23 \pm 3.07$   & $0.293 \pm 0.023$ \\
\midrule
\multirow{2}{*}{FENCE + GAD-MASS}
& 16 & $28.38 \pm 0.76$ & $44.71 \pm 0.91$ & $0.184 \pm 0.004$ & $\underline{25.94 \pm 0.63}$ & $\underline{44.49 \pm 0.72}$ & $\underline{0.161 \pm 0.006}$ \\
& 25 & $\underline{27.82 \pm 0.60}$ & $\mathbf{44.26 \pm 0.65}$ & $\mathbf{0.177 \pm 0.003}$ & $\mathbf{25.44 \pm 0.58}$ & $\mathbf{44.05 \pm 0.60}$ & $\mathbf{0.156 \pm 0.005}$ \\
\bottomrule
\end{tabular}%
}
\end{table}

\smallskip
\noindent\textbf{Probabilistic forecasting.}
We follow DiffSTG~\cite{wen2024diffstg} on PEMS08, where the task is to forecast the next $12$ traffic slices of $170$ road sensors from the past $12$, with their uncertainty. DiffSTG masks the horizon in a window of history and horizon, corrupts the window with isotropic noise, trains UGNet to predict the noise, and samples eight trajectories with $100$ DDIM steps. Their mean is scored by MAE and RMSE, and the ensemble by the continuous ranked probability score (CRPS), which rewards samples that are close to the reading and correctly spread. We view each time slice as a graph signal on the road graph of DiffSTG. Since the reverse process must generate what the history leaves undetermined, $p_k$ is the residual power at $\sclambda_k$ after a per-frequency ridge regression of the horizon on the history, fitted on the training split. Three further changes go beyond the forward process. The denoiser predicts the clean horizon in the graph Fourier basis, its time embedding takes the attenuation and the noise variance as input, and the history slices follow their own forward trajectory during sampling.

We sample with DDIM in the graph Fourier basis at $3$ and $8$ reverse steps, on a grid uniform in $\log r_t$, where MASS gives every step the same increment of log-SNR at every graph frequency, and select one checkpoint per seed on validation CRPS.
Table~\ref{tab:pems08_diffstg_comparison} reports the results. With $3$ steps against $100$, GAD-MASS improves the MAE and RMSE of DiffSTG, and at $8$ steps it stays within $1\%$ of them. It trails DiffSTG on CRPS by $8.5\%$ and $4.5\%$ at $3$ and $8$ steps, and lowers the CRPS of GAD-FCPS by $8$ to $9\%$. The ensemble mean is accurate, but its spread is too narrow, and finer grids beyond $8$ steps do not close the gap. A likely cause is that one clean power per graph frequency serves all twelve horizon steps, whereas the uncertainty grows with the step.

\begin{table}[t]
\centering
\caption{\textbf{Probabilistic forecasting on PEMS08} with $12$ history and $12$ horizon steps. Baselines as reported in~\cite{wen2024diffstg}.}
\label{tab:pems08_diffstg_comparison}
\scriptsize
\setlength{\tabcolsep}{2.5pt}
\renewcommand{\arraystretch}{1.05}
\resizebox{\columnwidth}{!}{%
\begin{tabular}{lcccc}
\toprule
Method & Steps & MAE$\downarrow$ & RMSE$\downarrow$ & CRPS$\downarrow$ \\
\midrule
Latent ODE~\cite{rubanova2019latent} & -- & 26.05 & 39.50 & 0.1100 \\
DeepAR~\cite{salinas2020deepar} & -- & 21.56 & 33.37 & 0.0700 \\
CSDI~\cite{tashiro2021csdi} & -- & 32.11 & 47.40 & 0.1100 \\
TimeGrad~\cite{rasul2021autoregressive} & -- & 24.46 & 38.06 & 0.0900 \\
MC Dropout~\cite{wu2021quantifying} & -- & 19.01 & 29.35 & 0.0700 \\
DiffSTG~\cite{wen2024diffstg} & 100 & \underline{17.68} & \underline{27.13} & $\mathbf{0.0600}$ \\
\midrule
\multirow{2}{*}{DiffSTG + GAD-FCPS}
& 3 & $17.70 \pm 0.18$ & $27.69 \pm 0.50$ & $0.0714 \pm 0.0008$ \\
& 8 & $17.85 \pm 0.17$ & $28.01 \pm 0.46$ & $0.0683 \pm 0.0005$ \\
\midrule
\multirow{2}{*}{DiffSTG + GAD-MASS}
& 3 & $\mathbf{17.27 \pm 0.10}$ & $\mathbf{26.83 \pm 0.38}$ & $0.0651 \pm 0.0005$ \\
& 8 & $17.78 \pm 0.17$ & $27.32 \pm 0.25$ & $\underline{0.0627 \pm 0.0008}$ \\
\bottomrule
\end{tabular}%
}
\end{table}

\section{Conclusion}
\label{sec:conclusion}

We have proposed and analyzed GAD, a graph-aware diffusion scheme that operates in the graph Fourier domain. The heat drift contracts each graph frequency at a rate proportional to its eigenvalue. With a shared diffusion rate, the resulting SNR imbalance affects denoising and can make the reverse SDE stiff. In our experiments, forward processes with a shared rate give no consistent gain over isotropic ones on the same graph and denoiser. MASS uses frequency-dependent diffusion rates to place all graph frequencies on one target SNR while keeping the GMRF prior. The analysis is exact for linear denoisers under graph stationarity, and MASS requires the eigendecomposition of the Laplacian and an estimate of the power spectrum of the data. Natural next steps are polynomial approximations for large graphs, clean powers that vary along the forecast horizon, and directed or time-varying graphs.

\section*{Acknowledgments}
The authors used ChatGPT (OpenAI) and Claude (Anthropic) to assist with language editing, implementation and debugging of experimental code, literature research, and minor drafting of manuscript text. AI-assisted contents were reviewed, verified, and validated by the authors, who take full responsibility for the final manuscript.

\bibliographystyle{IEEEtran}
\bibliography{references}

@IEEEtranBSTCTL{IEEEexample:BSTcontrol,
  CTLdash_repeated_names   = "no",
  CTLuse_url               = "no",
  CTLuse_article_number    = "yes",
  CTLuse_paper             = "yes",
  CTLuse_forced_etal       = "no",
  CTLmax_names_forced_etal = "10",
  CTLnames_show_etal       = "1",
  CTLuse_alt_spacing       = "yes",
  CTLalt_stretch_factor    = "4"
}

@article{Shuman_2013,
  author  = {Shuman, David I. and Narang, Sunil K. and Frossard, Pascal and
             Ortega, Antonio and Vandergheynst, Pierre},
  title   = {The Emerging Field of Signal Processing on Graphs: Extending
             High-Dimensional Data Analysis to Networks and Other Irregular
             Domains},
  journal = {IEEE Signal Process. Mag.},
  volume  = {30},
  number  = {3},
  pages   = {83--98},
  month   = may,
  year    = {2013}
}

@article{ortega2018graph,
  author  = {Ortega, Antonio and Frossard, Pascal and Kova{\v{c}}evi{\'c}, Jelena
             and Moura, Jos{\'e} M. F. and Vandergheynst, Pierre},
  title   = {Graph Signal Processing: Overview, Challenges, and Applications},
  journal = {Proc. IEEE},
  volume  = {106},
  number  = {5},
  pages   = {808--828},
  month   = may,
  year    = {2018}
}

@article{isufi2024graph,
  author  = {Isufi, Elvin and Gama, Fernando and Shuman, David I. and
             Segarra, Santiago},
  title   = {Graph Filters for Signal Processing and Machine Learning on Graphs},
  journal = {IEEE Trans. Signal Process.},
  volume  = {72},
  pages   = {4745--4781},
  year    = {2024}
}

@article{bronstein2017geometric,
  author  = {Bronstein, Michael M. and Bruna, Joan and LeCun, Yann and
             Szlam, Arthur and Vandergheynst, Pierre},
  title   = {Geometric Deep Learning: Going Beyond {E}uclidean Data},
  journal = {IEEE Signal Process. Mag.},
  volume  = {34},
  number  = {4},
  pages   = {18--42},
  month   = jul,
  year    = {2017}
}

@article{gama2020graphs,
  author  = {Gama, Fernando and Isufi, Elvin and Leus, Geert and
             Ribeiro, Alejandro},
  title   = {Graphs, Convolutions, and Neural Networks: From Graph Filters to
             Graph Neural Networks},
  journal = {IEEE Signal Process. Mag.},
  volume  = {37},
  number  = {6},
  pages   = {128--138},
  month   = nov,
  year    = {2020}
}

@article{corso2024graph,
  author    = {Corso, Gabriele and St{\"a}rk, Hannes and Jegelka, Stefanie and
               Jaakkola, Tommi and Barzilay, Regina},
  title     = {Graph Neural Networks},
  journal   = {Nature Rev. Methods Primers},
  volume    = {4},
  number    = {1},
  pages     = {17},
  year      = {2024},
  publisher = {Nature Publishing Group}
}

@article{chen2021graph,
  author  = {Chen, Siheng and Eldar, Yonina C. and Zhao, Lingxiao},
  title   = {Graph Unrolling Networks: Interpretable Neural Networks for Graph
             Signal Denoising},
  journal = {IEEE Trans. Signal Process.},
  volume  = {69},
  pages   = {3699--3713},
  year    = {2021}
}

@article{marques2017stationary,
  author  = {Marques, Antonio G. and Segarra, Santiago and Leus, Geert and
             Ribeiro, Alejandro},
  title   = {Stationary Graph Processes and Spectral Estimation},
  journal = {IEEE Trans. Signal Process.},
  volume  = {65},
  number  = {22},
  pages   = {5911--5926},
  month   = nov,
  year    = {2017}
}

@article{perraudin2017stationary,
  author  = {Perraudin, Nathana{\"e}l and Vandergheynst, Pierre},
  title   = {Stationary Signal Processing on Graphs},
  journal = {IEEE Trans. Signal Process.},
  volume  = {65},
  number  = {13},
  pages   = {3462--3477},
  month   = jul,
  year    = {2017}
}

@misc{schmidt2026stability,
  author       = {Schmidt, Martin and Mateos, Gonzalo},
  title        = {Stability of Flow Models for Graph Signals},
  howpublished = {arXiv:2607.07510},
  year         = {2026}
}

@book{chungspectral,
  author    = {Chung, Fan R. K.},
  title     = {Spectral Graph Theory},
  series    = {CBMS Regional Conference Series in Mathematics},
  number    = {92},
  publisher = {American Mathematical Society},
  address   = {Providence, RI, USA},
  year      = {1997}
}

@inproceedings{smola2003kernels,
  author    = {Smola, Alexander J. and Kondor, Risi},
  title     = {Kernels and Regularization on Graphs},
  booktitle = {Proc. Annu. Conf. Learn. Theory (COLT) and Kernel Workshop},
  series    = {Lecture Notes in Computer Science},
  volume    = {2777},
  pages     = {144--158},
  publisher = {Springer},
  year      = {2003}
}

@article{hammond2009waveletsgraphsspectralgraph,
  author  = {Hammond, David K. and Vandergheynst, Pierre and Gribonval, R{\'e}mi},
  title   = {Wavelets on Graphs via Spectral Graph Theory},
  journal = {Appl. Comput. Harmon. Anal.},
  volume  = {30},
  number  = {2},
  pages   = {129--150},
  year    = {2011}
}

@article{thanou2016learningheatdiffusiongraphs,
  author  = {Thanou, Dorina and Dong, Xiaowen and Kressner, Daniel and
             Frossard, Pascal},
  title   = {Learning Heat Diffusion Graphs},
  journal = {IEEE Trans. Signal Inf. Process. Netw.},
  volume  = {3},
  number  = {3},
  pages   = {484--499},
  month   = sep,
  year    = {2017}
}

@book{rue2005gaussian,
  author    = {Rue, H{\aa}vard and Held, Leonhard},
  title     = {Gaussian {M}arkov Random Fields: Theory and Applications},
  series    = {Monographs on Statistics and Applied Probability},
  volume    = {104},
  publisher = {Chapman and Hall/CRC},
  address   = {Boca Raton, FL, USA},
  year      = {2005}
}

@inproceedings{sohldickstein2015deepunsupervisedlearningusing,
  author    = {Sohl-Dickstein, Jascha and Weiss, Eric A. and
               Maheswaranathan, Niru and Ganguli, Surya},
  title     = {Deep Unsupervised Learning Using Nonequilibrium Thermodynamics},
  booktitle = {Proc. Int. Conf. Mach. Learn. (ICML)},
  pages     = {2256--2265},
  year      = {2015}
}

@inproceedings{ho2020denoisingdiffusionprobabilisticmodels,
  author    = {Ho, Jonathan and Jain, Ajay and Abbeel, Pieter},
  title     = {Denoising Diffusion Probabilistic Models},
  booktitle = {Adv. Neural Inf. Process. Syst. (NeurIPS)},
  volume    = {33},
  pages     = {6840--6851},
  year      = {2020}
}

@inproceedings{song2021scorebasedgenerativemodelingstochastic,
  author    = {Song, Yang and Sohl-Dickstein, Jascha and Kingma, Diederik P. and
               Kumar, Abhishek and Ermon, Stefano and Poole, Ben},
  title     = {Score-Based Generative Modeling Through Stochastic Differential
               Equations},
  booktitle = {Proc. Int. Conf. Learn. Represent. (ICLR)},
  year      = {2021}
}

@article{anderson1982reverse,
  author  = {Anderson, Brian D. O.},
  title   = {Reverse-Time Diffusion Equation Models},
  journal = {Stochastic Process. Appl.},
  volume  = {12},
  number  = {3},
  pages   = {313--326},
  year    = {1982}
}

@article{vincent2011connection,
  author  = {Vincent, Pascal},
  title   = {A Connection Between Score Matching and Denoising Autoencoders},
  journal = {Neural Comput.},
  volume  = {23},
  number  = {7},
  pages   = {1661--1674},
  year    = {2011}
}

@article{Efron01122011,
  author  = {Efron, Bradley},
  title   = {Tweedie's Formula and Selection Bias},
  journal = {J. Amer. Statist. Assoc.},
  volume  = {106},
  number  = {496},
  pages   = {1602--1614},
  year    = {2011}
}

@book{sarkka2019applied,
  author    = {S{\"a}rkk{\"a}, Simo and Solin, Arno},
  title     = {Applied Stochastic Differential Equations},
  series    = {Institute of Mathematical Statistics Textbooks},
  volume    = {10},
  publisher = {Cambridge Univ. Press},
  address   = {Cambridge, U.K.},
  year      = {2019}
}

@inproceedings{kingma2021variational,
  author    = {Kingma, Diederik P. and Salimans, Tim and Poole, Ben and
               Ho, Jonathan},
  title     = {Variational Diffusion Models},
  booktitle = {Adv. Neural Inf. Process. Syst. (NeurIPS)},
  volume    = {34},
  pages     = {21696--21707},
  year      = {2021}
}

@inproceedings{kingma2023understanding,
  author    = {Kingma, Diederik P. and Gao, Ruiqi},
  title     = {Understanding Diffusion Objectives as the {ELBO} with Simple Data
               Augmentation},
  booktitle = {Adv. Neural Inf. Process. Syst. (NeurIPS)},
  year      = {2023}
}

@inproceedings{karras2022elucidating,
  author    = {Karras, Tero and Aittala, Miika and Aila, Timo and Laine, Samuli},
  title     = {Elucidating the Design Space of Diffusion-Based Generative Models},
  booktitle = {Adv. Neural Inf. Process. Syst. (NeurIPS)},
  year      = {2022}
}

@inproceedings{nichol2021improved,
  author    = {Nichol, Alexander Quinn and Dhariwal, Prafulla},
  title     = {Improved Denoising Diffusion Probabilistic Models},
  booktitle = {Proc. Int. Conf. Mach. Learn. (ICML)},
  pages     = {8162--8171},
  year      = {2021}
}

@inproceedings{lin2024common,
  author    = {Lin, Shanchuan and Liu, Bingchen and Li, Jiashi and Yang, Xiao},
  title     = {Common Diffusion Noise Schedules and Sample Steps Are Flawed},
  booktitle = {Proc. IEEE/CVF Winter Conf. Appl. Comput. Vis. (WACV)},
  pages     = {5404--5411},
  year      = {2024}
}

@inproceedings{hang2023minsnr,
  author    = {Hang, Tiankai and Gu, Shuyang and Li, Chen and Bao, Jianmin and
               Chen, Dong and Hu, Han and Geng, Xin and Guo, Baining},
  title     = {Efficient Diffusion Training via Min-{SNR} Weighting Strategy},
  booktitle = {Proc. IEEE/CVF Int. Conf. Comput. Vis. (ICCV)},
  year      = {2023}
}

@inproceedings{sahoo2024mulan,
  author    = {Sahoo, Subham Sekhar and Gokaslan, Aaron and De Sa, Christopher
               and Kuleshov, Volodymyr},
  title     = {Diffusion Models with Learned Adaptive Noise},
  booktitle = {Adv. Neural Inf. Process. Syst. (NeurIPS)},
  year      = {2024}
}

@misc{chen2023importance,
  author       = {Chen, Ting},
  title        = {On the Importance of Noise Scheduling for Diffusion Models},
  howpublished = {arXiv:2301.10972},
  year         = {2023}
}

@inproceedings{voleti2022nonisotropic,
  author    = {Voleti, Vikram and Pal, Christopher and Oberman, Adam},
  title     = {Score-Based Denoising Diffusion with Non-Isotropic {G}aussian
               Noise Models},
  booktitle = {NeurIPS Workshop Score-Based Methods},
  year      = {2022}
}

@inproceedings{phillips2022spectral,
  author    = {Phillips, Angus and Seror, Thomas and Hutchinson, Michael and
               De Bortoli, Valentin and Doucet, Arnaud and Mathieu, Emile},
  title     = {Spectral Diffusion Processes},
  booktitle = {NeurIPS Workshop Score-Based Methods},
  year      = {2022}
}

@inproceedings{curreli2025nonisotropic,
  author    = {Curreli, Cecilia and Muhle, Dominik and Saroha, Abhishek and
               Ye, Zhenzhang and Marin, Riccardo and Cremers, Daniel},
  title     = {Nonisotropic {G}aussian Diffusion for Realistic {3D} Human Motion
               Prediction},
  booktitle = {Proc. IEEE/CVF Conf. Comput. Vis. Pattern Recognit. (CVPR)},
  pages     = {1871--1882},
  year      = {2025}
}

@inproceedings{rissanen2023generative,
  author    = {Rissanen, Severi and Heinonen, Markus and Solin, Arno},
  title     = {Generative Modelling with Inverse Heat Dissipation},
  booktitle = {Proc. Int. Conf. Learn. Represent. (ICLR)},
  year      = {2023}
}

@inproceedings{hoogeboom2023blurring,
  author    = {Hoogeboom, Emiel and Salimans, Tim},
  title     = {Blurring Diffusion Models},
  booktitle = {Proc. Int. Conf. Learn. Represent. (ICLR)},
  year      = {2023}
}

@article{daras2022soft,
  author  = {Daras, Giannis and Delbracio, Mauricio and Talebi, Hossein and
             Dimakis, Alexandros G. and Milanfar, Peyman},
  title   = {Soft Diffusion: Score Matching for General Corruptions},
  journal = {Trans. Mach. Learn. Res.},
  year    = {2023}
}

@misc{falck2024fourier,
  author       = {Falck, Fabian and Pandeva, Teodora and Zahirnia, Kiarash and
                  Lawrence, Rachel and Turner, Richard E. and Meeds, Edward and
                  Zazo, Javier and Karmalkar, Sushrut},
  title        = {A {F}ourier Space Perspective on Diffusion Models},
  howpublished = {arXiv:2505.11278},
  year         = {2025}
}

@misc{scimeca2024learning,
  author       = {Scimeca, Luca and Jiralerspong, Thomas and Earnshaw, Berton and
                  Hartford, Jason and Bengio, Yoshua},
  title        = {Learning What Matters: Steering Diffusion via Spectrally
                  Anisotropic Forward Noise},
  howpublished = {arXiv:2510.09660},
  year         = {2025}
}

@inproceedings{rozada2025graphawarediffusionsignalgeneration,
  author    = {Rozada, Sergio and K. B., Vimal and Cavallo, Andrea and
               Marques, Antonio G. and Jamali-Rad, Hadi and Isufi, Elvin},
  title     = {Graph-Aware Diffusion for Signal Generation},
  booktitle = {Proc. IEEE Int. Conf. Acoust., Speech, Signal Process. (ICASSP)},
  pages     = {461--465},
  year      = {2026}
}

@inproceedings{yang2025topological,
  author    = {Yang, Maosheng},
  title     = {Topological {S}chr{\"o}dinger Bridge Matching},
  booktitle = {Proc. Int. Conf. Learn. Represent. (ICLR)},
  year      = {2025}
}

@inproceedings{wyrwal2026topological,
  author    = {Wyrwal, Kacper and Ceylan, {\.I}smail {\.I}lkan and Tong, Alexander},
  title     = {Topological Flow Matching},
  booktitle = {Proc. Int. Conf. Learn. Represent. (ICLR)},
  year      = {2026}
}

@inproceedings{choi2023bspm,
  author    = {Choi, Jeongwhan and Hong, Seoyoung and Park, Noseong and
               Cho, Sung-Bae},
  title     = {Blurring-Sharpening Process Models for Collaborative Filtering},
  booktitle = {Proc. ACM SIGIR Conf. Res. Develop. Inf. Retr. (SIGIR)},
  pages     = {1096--1106},
  year      = {2023}
}

@inproceedings{zhu2024giffcf,
  author    = {Zhu, Yunqin and Wang, Chao and Zhang, Qi and Xiong, Hui},
  title     = {Graph Signal Diffusion Model for Collaborative Filtering},
  booktitle = {Proc. ACM SIGIR Conf. Res. Develop. Inf. Retr. (SIGIR)},
  pages     = {1380--1390},
  year      = {2024}
}

@inproceedings{xia2025sdiff,
  author    = {Xia, Rui and Cheng, Yanhua and Tang, Yongxiang and
               Liu, Xiaocheng and Liu, Xialong and Wang, Lisong and Jiang, Peng},
  title     = {{S-Diff}: An Anisotropic Diffusion Model for Collaborative
               Filtering in Spectral Domain},
  booktitle = {Proc. ACM Int. Conf. Web Search Data Mining (WSDM)},
  pages     = {70--78},
  year      = {2025}
}

@inproceedings{uslu2026graphgeneric,
  author    = {Uslu, Yi{\u{g}}it Berkay and Hadou, Samar and Rozada, Sergio and
               Saeedi Bidokhti, Shirin and Ribeiro, Alejandro},
  title     = {Graph Signal Generative Diffusion Models},
  booktitle = {Proc. IEEE Int. Conf. Acoust., Speech, Signal Process. (ICASSP)},
  pages     = {626--630},
  year      = {2026}
}

@inproceedings{uslu2025generative,
  author    = {Uslu, Yi{\u{g}}it Berkay and Hadou, Samar and
               Saeedi Bidokhti, Shirin and Ribeiro, Alejandro},
  title     = {Generative Diffusion Models for Resource Allocation in Wireless
               Networks},
  booktitle = {Proc. IEEE Int. Workshop Comput. Adv. Multi-Sensor Adaptive
               Process. (CAMSAP)},
  year      = {2025}
}

@misc{uslu2026graph,
  author       = {Uslu, Yi{\u{g}}it Berkay and Hadou, Samar and
                  Saeedi Bidokhti, Shirin and Ribeiro, Alejandro},
  title        = {Graph Signal Diffusion Models for Wireless Resource Allocation},
  howpublished = {arXiv:2604.05175},
  year         = {2026}
}

@misc{uslu2026generative,
  title={Generative Diffusion Models of Stochastic Graph Signals},
  author={Uslu, Yi{\u{g}}it Berkay and Hadou, Samar and Rozada, Sergio and Bidokhti, Shirin Saeedi and Ribeiro, Alejandro},
  howpublished = {arXiv:2607.06833},
  year         = {2026}
}

@inproceedings{park2024equivariant,
  author    = {Park, Jiwoong and Shen, Yang},
  title     = {Equivariant Blurring Diffusion for Hierarchical Molecular
               Conformer Generation},
  booktitle = {Adv. Neural Inf. Process. Syst. (NeurIPS)},
  year      = {2024}
}

@inproceedings{rahaman2019spectral,
  author    = {Rahaman, Nasim and Baratin, Aristide and Arpit, Devansh and
               Draxler, Felix and Lin, Min and Hamprecht, Fred A. and
               Bengio, Yoshua and Courville, Aaron},
  title     = {On the Spectral Bias of Neural Networks},
  booktitle = {Proc. Int. Conf. Mach. Learn. (ICML)},
  pages     = {5301--5310},
  year      = {2019}
}

@inproceedings{wang2024analytical,
  author    = {Wang, Binxu and Pehlevan, Cengiz},
  title     = {An Analytical Theory of Spectral Bias in the Learning Dynamics of
               Diffusion Models},
  booktitle = {Adv. Neural Inf. Process. Syst. (NeurIPS)},
  year      = {2025}
}

@inproceedings{kadkhodaie2024generalization,
  author    = {Kadkhodaie, Zahra and Guth, Florentin and Simoncelli, Eero P. and
               Mallat, St{\'e}phane},
  title     = {Generalization in Diffusion Models Arises from Geometry-Adaptive
               Harmonic Representations},
  booktitle = {Proc. Int. Conf. Learn. Represent. (ICLR)},
  year      = {2024}
}

@inproceedings{jacot2018neural,
  author    = {Jacot, Arthur and Gabriel, Franck and Hongler, Cl{\'e}ment},
  title     = {Neural Tangent Kernel: Convergence and Generalization in Neural
               Networks},
  booktitle = {Adv. Neural Inf. Process. Syst. (NeurIPS)},
  year      = {2018}
}

@inproceedings{du2019graph,
  author    = {Du, Simon S. and Hou, Kangcheng and P{\'o}czos, Barnab{\'a}s and
               Salakhutdinov, Ruslan and Wang, Ruosong and Xu, Keyulu},
  title     = {Graph Neural Tangent Kernel: Fusing Graph Neural Networks with
               Graph Kernels},
  booktitle = {Adv. Neural Inf. Process. Syst. (NeurIPS)},
  year      = {2019}
}

@inproceedings{liu2020linearity,
  author    = {Liu, Chaoyue and Zhu, Libin and Belkin, Mikhail},
  title     = {On the Linearity of Large Non-Linear Models: When and Why the
               Tangent Kernel Is Constant},
  booktitle = {Adv. Neural Inf. Process. Syst. (NeurIPS)},
  year      = {2020}
}

@inproceedings{song2021denoising,
  author    = {Song, Jiaming and Meng, Chenlin and Ermon, Stefano},
  title     = {Denoising Diffusion Implicit Models},
  booktitle = {Proc. Int. Conf. Learn. Represent. (ICLR)},
  year      = {2021}
}

@inproceedings{liu2022pseudo,
  author    = {Liu, Luping and Ren, Yi and Lin, Zhijie and Zhao, Zhou},
  title     = {Pseudo Numerical Methods for Diffusion Models on Manifolds},
  booktitle = {Proc. Int. Conf. Learn. Represent. (ICLR)},
  year      = {2022}
}

@inproceedings{lu2022dpmsolverfastodesolver,
  author    = {Lu, Cheng and Zhou, Yuhao and Bao, Fan and Chen, Jianfei and
               Li, Chongxuan and Zhu, Jun},
  title     = {{DPM-Solver}: A Fast {ODE} Solver for Diffusion Probabilistic
               Model Sampling in Around 10 Steps},
  booktitle = {Adv. Neural Inf. Process. Syst. (NeurIPS)},
  year      = {2022}
}

@article{lu2022dpmpp,
  author  = {Lu, Cheng and Zhou, Yuhao and Bao, Fan and Chen, Jianfei and
             Li, Chongxuan and Zhu, Jun},
  title   = {{DPM-Solver++}: Fast Solver for Guided Sampling of Diffusion
             Probabilistic Models},
  journal = {Mach. Intell. Res.},
  volume  = {22},
  number  = {4},
  pages   = {730--751},
  year    = {2025}
}

@inproceedings{zhao2023unipc,
  author    = {Zhao, Wenliang and Bai, Lujia and Rao, Yongming and Zhou, Jie and
               Lu, Jiwen},
  title     = {{UniPC}: A Unified Predictor-Corrector Framework for Fast Sampling
               of Diffusion Models},
  booktitle = {Adv. Neural Inf. Process. Syst. (NeurIPS)},
  year      = {2023}
}

@inproceedings{salimans2022progressivedistillationfastsampling,
  author    = {Salimans, Tim and Ho, Jonathan},
  title     = {Progressive Distillation for Fast Sampling of Diffusion Models},
  booktitle = {Proc. Int. Conf. Learn. Represent. (ICLR)},
  year      = {2022}
}

@inproceedings{song2023consistencymodels,
  author    = {Song, Yang and Dhariwal, Prafulla and Chen, Mark and
               Sutskever, Ilya},
  title     = {Consistency Models},
  booktitle = {Proc. Int. Conf. Mach. Learn. (ICML)},
  year      = {2023}
}

@article{dahlquist1963special,
  author  = {Dahlquist, Germund G.},
  title   = {A Special Stability Problem for Linear Multistep Methods},
  journal = {BIT Numer. Math.},
  volume  = {3},
  number  = {1},
  pages   = {27--43},
  year    = {1963}
}

@book{hairer1996solving,
  author    = {Hairer, Ernst and Wanner, Gerhard},
  title     = {Solving Ordinary Differential Equations {II}: Stiff and
               Differential-Algebraic Problems},
  series    = {Springer Series in Computational Mathematics},
  volume    = {14},
  edition   = {2nd},
  publisher = {Springer},
  address   = {Berlin, Germany},
  year      = {1996}
}

@inproceedings{zhang2023fast,
  author    = {Zhang, Qinsheng and Chen, Yongxin},
  title     = {Fast Sampling of Diffusion Models with Exponential Integrator},
  booktitle = {Proc. Int. Conf. Learn. Represent. (ICLR)},
  year      = {2023}
}

@inproceedings{rendle2009bpr,
  author    = {Rendle, Steffen and Freudenthaler, Christoph and Gantner, Zeno and
               Schmidt-Thieme, Lars},
  title     = {{BPR}: {B}ayesian Personalized Ranking from Implicit Feedback},
  booktitle = {Proc. Conf. Uncertainty Artif. Intell. (UAI)},
  pages     = {452--461},
  year      = {2009}
}

@inproceedings{he2020lightgcn,
  author    = {He, Xiangnan and Deng, Kuan and Wang, Xiang and Li, Yan and
               Zhang, Yongdong and Wang, Meng},
  title     = {{LightGCN}: Simplifying and Powering Graph Convolution Network for
               Recommendation},
  booktitle = {Proc. ACM SIGIR Conf. Res. Develop. Inf. Retr. (SIGIR)},
  pages     = {639--648},
  year      = {2020}
}

@inproceedings{steck2019embarrassingly,
  author    = {Steck, Harald},
  title     = {Embarrassingly Shallow Autoencoders for Sparse Data},
  booktitle = {Proc. World Wide Web Conf. (WWW)},
  pages     = {3251--3257},
  year      = {2019}
}

@inproceedings{liang2018variational,
  author    = {Liang, Dawen and Krishnan, Rahul G. and Hoffman, Matthew D. and
               Jebara, Tony},
  title     = {Variational Autoencoders for Collaborative Filtering},
  booktitle = {Proc. World Wide Web Conf. (WWW)},
  pages     = {689--698},
  year      = {2018}
}

@inproceedings{carraro2022conditioned,
  author    = {Carraro, Tommaso and Polato, Mirko and Bergamin, Luca and
               Aiolli, Fabio},
  title     = {Conditioned Variational Autoencoder for Top-{N} Item Recommendation},
  booktitle = {Proc. Int. Conf. Artif. Neural Netw. (ICANN)},
  series    = {Lecture Notes in Computer Science},
  volume    = {13530},
  pages     = {785--796},
  publisher = {Springer},
  year      = {2022}
}

@inproceedings{wang2023diffrec,
  author    = {Wang, Wenjie and Xu, Yiyan and Feng, Fuli and Lin, Xinyu and
               He, Xiangnan and Chua, Tat-Seng},
  title     = {Diffusion Recommender Model},
  booktitle = {Proc. ACM SIGIR Conf. Res. Develop. Inf. Retr. (SIGIR)},
  pages     = {832--841},
  year      = {2023}
}

@inproceedings{lee2025cdiff4rec,
  author    = {Lee, Gyuseok and Zhu, Yaochen and Yu, Hwanjo and Zhou, Yao and
               Li, Jundong},
  title     = {Collaborative Diffusion Model for Recommender System},
  booktitle = {Companion Proc. ACM Web Conf. (WWW)},
  pages     = {1091--1095},
  year      = {2025}
}

@article{guo2021learning,
  author  = {Guo, Shengnan and Lin, Youfang and Wan, Huaiyu and Li, Xiucheng and
             Cong, Gao},
  title   = {Learning Dynamics and Heterogeneity of Spatial-Temporal Graph Data
             for Traffic Forecasting},
  journal = {IEEE Trans. Knowl. Data Eng.},
  volume  = {34},
  number  = {11},
  pages   = {5415--5428},
  year    = {2022}
}

@inproceedings{wu2021inductive,
  author    = {Wu, Yuankai and Zhuang, Dingyi and Labbe, Aurelie and Sun, Lijun},
  title     = {Inductive Graph Neural Networks for Spatiotemporal Kriging},
  booktitle = {Proc. AAAI Conf. Artif. Intell. (AAAI)},
  volume    = {35},
  number    = {5},
  pages     = {4478--4485},
  year      = {2021}
}

@inproceedings{peng2023generative,
  author    = {Peng, Wenchuang and Lin, Youfang and Guo, Shengnan and
               Tang, Weiwen and Liu, Le and Wan, Huaiyu},
  title     = {Generative-Contrastive-Attentive Spatial-Temporal Network for
               Traffic Data Imputation},
  booktitle = {Proc. Pacific-Asia Conf. Knowl. Discov. Data Mining (PAKDD)},
  series    = {Lecture Notes in Computer Science},
  pages     = {45--56},
  publisher = {Springer},
  year      = {2023}
}

@article{chen2024laplacian,
  author  = {Chen, Xinyu and Cheng, Zhanhong and Cai, HanQin and Saunier, Nicolas
             and Sun, Lijun},
  title   = {{L}aplacian Convolutional Representation for Traffic Time Series
             Imputation},
  journal = {IEEE Trans. Knowl. Data Eng.},
  volume  = {36},
  number  = {11},
  pages   = {6490--6502},
  month   = nov,
  year    = {2024}
}

@inproceedings{nie2024imputeformer,
  author    = {Nie, Tong and Qin, Guoyang and Ma, Wei and Mei, Yuewen and
               Sun, Jian},
  title     = {{ImputeFormer}: Low Rankness-Induced Transformers for
               Generalizable Spatiotemporal Imputation},
  booktitle = {Proc. ACM SIGKDD Conf. Knowl. Discov. Data Mining (KDD)},
  pages     = {2260--2271},
  year      = {2024}
}

@inproceedings{shukla2021multi,
  author    = {Shukla, Satya Narayan and Marlin, Benjamin M.},
  title     = {Multi-Time Attention Networks for Irregularly Sampled Time Series},
  booktitle = {Proc. Int. Conf. Learn. Represent. (ICLR)},
  year      = {2021}
}

@inproceedings{tashiro2021csdi,
  author    = {Tashiro, Yusuke and Song, Jiaming and Song, Yang and
               Ermon, Stefano},
  title     = {{CSDI}: Conditional Score-Based Diffusion Models for Probabilistic
               Time Series Imputation},
  booktitle = {Adv. Neural Inf. Process. Syst. (NeurIPS)},
  volume    = {34},
  year      = {2021}
}

@inproceedings{liu2023pristi,
  author    = {Liu, Mingzhe and Huang, Han and Feng, Hao and Sun, Leilei and
               Du, Bowen and Fu, Yanjie},
  title     = {{PriSTI}: A Conditional Diffusion Framework for Spatiotemporal
               Imputation},
  booktitle = {Proc. IEEE Int. Conf. Data Eng. (ICDE)},
  pages     = {1927--1939},
  year      = {2023}
}

@inproceedings{mao2026fence,
  author    = {Mao, Xiaowei and Ding, Huihu and Lin, Yan and Wu, Tingrui and
               Guo, Shengnan and Qiu, Dazhuo and Fang, Feiling and Hu, Jilin and
               Wan, Huaiyu},
  title     = {Spatial-Temporal Feedback Diffusion Guidance for Controlled
               Traffic Imputation},
  booktitle = {Proc. AAAI Conf. Artif. Intell. (AAAI)},
  year      = {2026}
}

@inproceedings{wen2024diffstg,
  author    = {Wen, Haomin and Lin, Youfang and Xia, Yutong and Wan, Huaiyu and
               Wen, Qingsong and Zimmermann, Roger and Liang, Yuxuan},
  title     = {{DiffSTG}: Probabilistic Spatio-Temporal Graph Forecasting with
               Denoising Diffusion Models},
  booktitle = {Proc. ACM SIGSPATIAL Int. Conf. Adv. Geogr. Inf. Syst.
               (SIGSPATIAL)},
  note      = {Art. no. 60},
  year      = {2023}
}

@inproceedings{rubanova2019latent,
  author    = {Rubanova, Yulia and Chen, Ricky T. Q. and Duvenaud, David},
  title     = {Latent {ODE}s for Irregularly-Sampled Time Series},
  booktitle = {Adv. Neural Inf. Process. Syst. (NeurIPS)},
  year      = {2019}
}

@article{salinas2020deepar,
  author  = {Salinas, David and Flunkert, Valentin and Gasthaus, Jan and
             Januschowski, Tim},
  title   = {{DeepAR}: Probabilistic Forecasting with Autoregressive Recurrent
             Networks},
  journal = {Int. J. Forecasting},
  volume  = {36},
  number  = {3},
  pages   = {1181--1191},
  year    = {2020}
}

@inproceedings{rasul2021autoregressive,
  author    = {Rasul, Kashif and Seward, Calvin and Schuster, Ingmar and
               Vollgraf, Roland},
  title     = {Autoregressive Denoising Diffusion Models for Multivariate
               Probabilistic Time Series Forecasting},
  booktitle = {Proc. Int. Conf. Mach. Learn. (ICML)},
  year      = {2021}
}

@inproceedings{wu2021quantifying,
  author    = {Wu, Dongxia and Gao, Liyao and Chinazzi, Matteo and Xiong, Xinyue
               and Vespignani, Alessandro and Ma, Yi-An and Yu, Rose},
  title     = {Quantifying Uncertainty in Deep Spatiotemporal Forecasting},
  booktitle = {Proc. ACM SIGKDD Conf. Knowl. Discov. Data Mining (KDD)},
  pages     = {1841--1851},
  year      = {2021}
}

@inproceedings{lee2022priorgrad,
  title     = {{PriorGrad}: Improving Conditional Denoising Diffusion Models with
               Data-Dependent Adaptive Prior},
  author    = {Lee, Sang-gil and Kim, Heeseung and Shin, Chaehun and Tan, Xu and
               Liu, Chang and Meng, Qi and Qin, Tao and Chen, Wei and
               Yoon, Sungroh and Liu, Tie-Yan},
  booktitle = {International Conference on Learning Representations (ICLR)},
  year      = {2022}
}

@article{huang2024blue,
  title   = {Blue Noise for Diffusion Models},
  author  = {Huang, Xingchang and Sala{\"u}n, Corentin and Vasconcelos, Cristina and
             Theobalt, Christian and {\"O}ztireli, Cengiz and Singh, Gurprit},
  journal = {ACM SIGGRAPH 2024 Conference Papers},
  year    = {2024}
}

@inproceedings{zhang2023gddim,
  title     = {{gDDIM}: Generalized Denoising Diffusion Implicit Models},
  author    = {Zhang, Qinsheng and Tao, Molei and Chen, Yongxin},
  booktitle = {International Conference on Learning Representations (ICLR)},
  year      = {2023}
}

@inproceedings{guth2022wavelet,
  title     = {Wavelet Score-Based Generative Modeling},
  author    = {Guth, Florentin and Coste, Simon and De Bortoli, Valentin and
               Mallat, St{\'e}phane},
  booktitle = {Advances in Neural Information Processing Systems (NeurIPS)},
  year      = {2022}
}

@book{kay1993fundamentals,
   author    = {Kay, Steven M.},
   title     = {Fundamentals of Statistical Signal Processing: Estimation Theory},
   publisher = {Prentice Hall},
   address   = {Upper Saddle River, NJ, USA},
   year      = {1993}
}

@inproceedings{defferrard2016convolutional,
  author    = {Defferrard, Micha{\"e}l and Bresson, Xavier and
               Vandergheynst, Pierre},
  title     = {Convolutional Neural Networks on Graphs with Fast Localized
               Spectral Filtering},
  booktitle = {Adv. Neural Inf. Process. Syst. (NIPS)},
  volume    = {29},
  year      = {2016}
}

\appendices

\section{Proofs for Spectral Analysis of Denoising Regression}
\label{app:denoising_regression}

Throughout the appendices the diffusion time $t\in(0,T]$ is fixed, $\xp[\,\cdot\mid t]$ denotes the expectation over the pair $(\vcx_0,\vcx_t)$ at time $t$, $k\in\{0,\ldots,n-1\}$ indexes the graph frequencies, and $q_k=\scsigma^{2}/\sclambda_{\scgamma,k}$ is the prior power at $\sclambda_k$, so that the power ratio is $g_k=p_k/q_k$. Since $\scbc_t>0$ by \eqref{eq:integrated_schedule} and $\sclambda_{\scgamma,k}\geq\scgamma>0$ by \eqref{eq:shifted_laplacian}, the attenuation $a_{t,k}$ and the noise variance $\scnu_{t,k}$ satisfy
\begin{equation}
    a_{t,k}=e^{-\sclambda_{\scgamma,k}\scbc_t}\in(0,1),
    \quad
    \scrho_{t,k}
    \geq
    \scnu_{t,k}
    =
    q_k\bigl(1-a_{t,k}^{2}\bigr)
    >0 .
    \label{eq:app_channel_range}
\end{equation}
Gradient flow is the continuous-time limit of gradient descent as the step size tends to zero. A trainable vector $\vcxi$ that minimizes an objective $\mathcal{J}$ evolves from an initialization $\vcxi(0)$ as
\begin{equation}
    \frac{d\vcxi(\sctau)}{d\sctau}
    =
    -\eta\,\nabla_{\vcxi}\mathcal{J}\bigl(\vcxi(\sctau)\bigr),
    \qquad \sctau\geq0,
    \label{eq:gradient_flow}
\end{equation}
with optimization time $\sctau$ and learning rate $\eta>0$. Under a shared diffusion rate, $\snr_{t,k}$ depends on $t$ only through $\scbc_t$, and we write $\snr_k(\scbc)$ for this dependence.

To help exposition, each proof is preceded by an auxiliary lemma. The proofs of the lemmas are provided in Appendix \ref{app:lemma_proofs}. 
\begin{lemma}[Optimal linear denoiser]
\label{lem:optimal_spectral_response}
Let Assumptions~\ref{ass:simple_spectrum} and~\ref{ass:graph_wss} hold, and let $t\in(0,T]$. Then, the unique minimizer of the fixed-time error $\stL_t(f_{\mtW})$ over $\mtW\in\fdR^{n\times n}$ is the spectral graph filter denoiser \eqref{eq:spectral_filter_denoiser} with the frequency response \eqref{eq:optimal_modewise_gain}.
\end{lemma}

\begin{proof}[\textbf{Proof of Proposition~\ref{thm:modewise_denoising_difficulty}}]
By Lemma~\ref{lem:optimal_spectral_response}, the optimal linear denoiser is a spectral graph filter. Restricting the fixed-time error to $\mtW=\mtV\diag(\vch_t)\mtV^{\Tr}$ separates it across graph frequencies,
\begin{equation}
    \stL_t(f_{\vch_t})
    =
    \sum_{k=0}^{n-1}\ell_{t,k}(h_{t,k}),
    \quad
    \ell_{t,k}(h)
    =
    \xp\bigl[(\sctx_{0,k}-h\sctx_{t,k})^{2}\mid t\bigr] ,
    \label{eq:app_risk_separation}
\end{equation}
with $\ell_{t,k}$ the error at $\sclambda_k$. The noise $\scte_{t,k}$ is zero-mean and independent of $\vcx_0$, hence $\xp[\sctx_{0,k}\sctx_{t,k}\mid t]=a_{t,k}p_k$, $\xp[\sctx_{t,k}^{\,2}\mid t]=\scrho_{t,k}$, and $\ell_{t,k}(h)=p_k-2a_{t,k}p_k\,h+\scrho_{t,k}h^{2}$. Completing the square in $\ell_{t,k}(h)$ with $h_{t,k}^{\star}$ of \eqref{eq:optimal_modewise_gain} gives
\begin{equation}
    \ell_{t,k}(h)
    =
    \scrho_{t,k}
    \bigl(h-h_{t,k}^{\star}\bigr)^{2}
    +
    p_k-\frac{a_{t,k}^{2}p_k^{2}}{\scrho_{t,k}} .
    \label{eq:app_completed_square}
\end{equation}
Summing over $k$ gives the fixed-time error of the spectral graph filter as a separable quadratic,
\begin{equation}
    \stL_t(f_{\vch_t})
    =
    \stB_t
    +
    \sum_{k=0}^{n-1}
    \scrho_{t,k}
    \bigl(
        h_{t,k}-h_{t,k}^{\star}
    \bigr)^{2} ,
    \label{eq:spectral_loss_quadratic}
\end{equation}
with constant Hessian $2\diag(\scrho_{t,0},\ldots,\scrho_{t,n-1})$ and with $\stB_t$ the sum of the constant terms in \eqref{eq:app_completed_square}. At $h=h_{t,k}^{\star}$ only the constant term remains. Since $\scrho_{t,k}-a_{t,k}^{2}p_k=\scnu_{t,k}$ by \eqref{eq:effective_variance_convex_combination} and $\scrho_{t,k}/\scnu_{t,k}=1+\snr_{t,k}$ by Definition~\ref{def:modewise_snr},
\begin{equation}
    \ell_{t,k}(h_{t,k}^{\star})
    =
    \frac{p_k\scnu_{t,k}}{\scrho_{t,k}}
    =
    \frac{p_k}{1+\snr_{t,k}} .
    \label{eq:app_modal_floor}
\end{equation}
Dividing \eqref{eq:app_modal_floor} by $p_k>0$ gives \eqref{eq:modewise_difficulty}. Summing \eqref{eq:app_modal_floor} over $k$ gives \eqref{eq:benchmark_risk}, and Lemma~\ref{lem:optimal_spectral_response} makes this sum the minimum over the whole linear class.
\end{proof}

\begin{proof}[\textbf{Proof of Proposition~\ref{thm:optimization_spectrum}}]
By \eqref{eq:spectral_loss_quadratic}, $\partial\stL_t(f_{\vch_t})/\partial h_{t,k} =2\scrho_{t,k}(h_{t,k}-h_{t,k}^{\star})$ depends on $h_{t,k}$ alone. The gradient flow \eqref{eq:gradient_flow} with $\vcxi=\vch_t$ therefore splits into the $n$ scalar equations
\begin{equation}
    \frac{dh_{t,k}(\sctau)}{d\sctau}
    =
    -2\eta\,\scrho_{t,k}
    \bigl(
        h_{t,k}(\sctau)-h_{t,k}^{\star}
    \bigr),
    \quad
    k=0,\ldots,n-1 .
    \label{eq:spectral_gradient_flow_mode}
\end{equation}
Fix a graph frequency $\sclambda_k$ and set $\scch_{t,k}(\sctau)=h_{t,k}(\sctau)-h_{t,k}^{\star}$, where the optimal frequency response $h_{t,k}^{\star}$ does not depend on $\sctau$. Then
\begin{equation}
    \frac{d}{d\sctau}
    \Bigl[
        e^{2\eta\scrho_{t,k}\sctau}\scch_{t,k}(\sctau)
    \Bigr]
    =
    e^{2\eta\scrho_{t,k}\sctau}
    \Bigl[
        \frac{d\scch_{t,k}}{d\sctau}
        +
        2\eta\scrho_{t,k}\scch_{t,k}
    \Bigr]
    =
    0 ,
    \label{eq:app_integrating_factor}
\end{equation}
hence $\scch_{t,k}(\sctau)=e^{-2\eta\scrho_{t,k}\sctau}\scch_{t,k}(0)$, which is \eqref{eq:spectral_training_dynamics} with $\sctau_{t,k}=(2\eta\scrho_{t,k})^{-1}$.
\end{proof}

\begin{lemma}[Informative window]
\label{lem:informative_window}
Let the diffusion rate be shared, and let $p_k>0$. Then, the set of integrated-rate values at which graph frequency $\sclambda_k$ is $\sceps$-informative,
\begin{equation}
    \stI_k(\sceps)
    =
    \bigl\{
        \scbc\geq0
        \;:\;
        \sceps\leq\snr_k(\scbc)\leq\sceps^{-1}
    \bigr\} ,
    \label{eq:app_informative_set}
\end{equation}
is the interval
\begin{equation}
    \stI_k(\sceps)
    =
    \left[
        \frac{\log(1+g_k\sceps)}{2\sclambda_{\scgamma,k}},
        \;
        \frac{\log(1+g_k\sceps^{-1})}{2\sclambda_{\scgamma,k}}
    \right],
    \label{eq:app_informative_interval}
\end{equation}
and its length satisfies
\begin{equation}
    \bigl|\stI_k(\sceps)\bigr|
    =
    \frac{1}{2\sclambda_{\scgamma,k}}
    \log
    \frac{1+g_k\sceps^{-1}}
         {1+g_k\sceps}
    \;\leq\;
    \frac{\log(1/\sceps)}{\sclambda_{\scgamma,k}} .
    \label{eq:informative_window}
\end{equation}
\end{lemma}

\begin{proof}[\textbf{Proof of Proposition~\ref{prop:informative_band}}]
Evaluating \eqref{eq:app_informative_interval} at $\scbc=\scbc_t>0$, graph frequency $\sclambda_k$ is informative at time $t$ if and only if
\begin{equation}
    \frac{\log(1+g_k\sceps)}{2\scbc_t}
    \;\leq\;
    \sclambda_{\scgamma,k}
    \;\leq\;
    \frac{\log(1+g_k\sceps^{-1})}{2\scbc_t} .
    \label{eq:app_informative_fixed_time}
\end{equation}
Let $j$ and $k$ both be informative at time $t$. Take the upper bound in \eqref{eq:app_informative_fixed_time} for $j$ and the lower bound for $k$. Since $\log(1+x)$ increases, $g_j\leq g_{\max}$ and $g_k\geq g_{\min}>0$,
\begin{equation*}
    \frac{\sclambda_{\scgamma,j}}{\sclambda_{\scgamma,k}}
    \leq
    \frac{\log(1+g_j\sceps^{-1})}{\log(1+g_k\sceps)}
    \leq
    \frac{\log\bigl(1+g_{\max}\sceps^{-1}\bigr)}
         {\log\bigl(1+g_{\min}\sceps\bigr)} . \qedhere
\end{equation*}
\end{proof}

\begin{lemma}[Orthogonality of the posterior mean]
\label{lem:bayes_orthogonality}
Let $t\in(0,T]$. Then, for every random variable $\scz$
that is a function of $\vcx_t$,
\begin{equation}
    \xp\bigl[
        (\sctx_{0,k}-m_{t,k})\,\scz
        \,\big|\,t
    \bigr]
    =
    0 ,
    \qquad
    k=0,\ldots,n-1 .
    \label{eq:app_orthogonality}
\end{equation}
\end{lemma}

Let $m_{t,k}=[\mtV^{\Tr}\fnm_t(\vcx_t)]_k$ and $\widehat{\sctx}_{0,k}=[\mtV^{\Tr}f_{\Theta}(\vcx_t,\mtL,t)]_k$ be the $k$th GFT coefficients of the posterior mean and of the GNN output, as in Proposition~\ref{thm:gnn_risk_decomposition}. At graph frequency $\sclambda_k$, we denote the MMSE, the gain of the posterior mean over the optimal linear denoiser, and the error of the GNN with respect to the posterior mean by
\begin{equation}
\begin{aligned}
    \mmse_{t,k}
    &=
    \xp\!\left[
        (\sctx_{0,k}-m_{t,k})^{2}\,\middle|\,t
    \right],
    \\
    \scDelta_{t,k}
    &=
    \xp\!\left[
        (m_{t,k}-h_{t,k}^{\star}\sctx_{t,k})^{2}\,\middle|\,t
    \right],
    \\
    \scomega_{t,k}
    &=
    \xp\!\left[
        (m_{t,k}-\widehat{\sctx}_{0,k})^{2}\,\middle|\,t
    \right],
\end{aligned}
\label{eq:modewise_error_terms}
\end{equation}
with sums $\mmse_t=\sum_k\mmse_{t,k}$, $\scDelta_{\mathrm{PM},t}=\sum_k\scDelta_{t,k}$ and $\scOmega_{\Theta,t}=\sum_k\scomega_{t,k}$. The three estimates $h_{t,k}^{\star}\sctx_{t,k}$, $m_{t,k}$ and $\widehat{\sctx}_{0,k}$ are functions of $\vcx_t$, and Lemma~\ref{lem:bayes_orthogonality} splits the error of each one around the posterior mean. First, write $\sctx_{0,k}-\widehat{\sctx}_{0,k} =(\sctx_{0,k}-m_{t,k})+(m_{t,k}-\widehat{\sctx}_{0,k})$ and take the second moment. The cross term vanishes by \eqref{eq:app_orthogonality} with $\scz=m_{t,k}-\widehat{\sctx}_{0,k}$, and the two remaining terms are $\mmse_{t,k}$ and $\scomega_{t,k}$. Second, write $\sctx_{0,k}-h_{t,k}^{\star}\sctx_{t,k} =(\sctx_{0,k}-m_{t,k})+(m_{t,k}-h_{t,k}^{\star}\sctx_{t,k})$. The cross term vanishes with $\scz=m_{t,k}-h_{t,k}^{\star}\sctx_{t,k}$, and the second moment of the left side is $p_kd_{t,k}$ by \eqref{eq:modewise_difficulty}. Third, write $\sctx_{0,k}=(\sctx_{0,k}-m_{t,k})+m_{t,k}$ and take $\scz=m_{t,k}$. The three identities are
\begin{equation}
\begin{aligned}
    \xp\bigl[
        (\sctx_{0,k}-\widehat{\sctx}_{0,k})^{2}\mid t
    \bigr]
    &=
    \mmse_{t,k}+\scomega_{t,k},
    \\
    p_k\,d_{t,k}
    &=
    \mmse_{t,k}+\scDelta_{t,k},
    \\
    p_k
    &=
    \mmse_{t,k}+\xp\bigl[m_{t,k}^{2}\mid t\bigr] .
\end{aligned}
    \label{eq:app_modal_splits}
\end{equation}

\begin{proof}[\textbf{Proof of Proposition~\ref{thm:gnn_risk_decomposition}}]
Since $\mtV$ is orthogonal, every $\vcz\in\fdR^{n}$ satisfies $\|\vcz\|_2^{2}=\sum_k[\mtV^{\Tr}\vcz]_k^{2}$. Summing the first line of \eqref{eq:app_modal_splits} over $k$ therefore gives
\begin{equation}
    \stL_t(f_{\Theta})
    =
    \mmse_t+\scOmega_{\Theta,t} .
    \label{eq:app_risk_two_terms}
\end{equation}
Summing the second line over $k$, with $\stB_t=\sum_kp_kd_{t,k}$ from \eqref{eq:benchmark_risk}, gives
\begin{equation}
    \mmse_t
    =
    \stB_t-\scDelta_{\mathrm{PM},t} .
    \label{eq:app_bayes_sum}
\end{equation}
Substituting \eqref{eq:app_bayes_sum} into \eqref{eq:app_risk_two_terms} yields \eqref{eq:gnn_risk_decomposition}. Both $\scDelta_{\mathrm{PM},t}$ and $\scOmega_{\Theta,t}$ are sums of expected squares and hence nonnegative.
\end{proof}

\begin{proof}[\textbf{Proof of Corollary~\ref{prop:bayes_target_energy}}]
Eliminating $\mmse_{t,k}$ between the last two lines of \eqref{eq:app_modal_splits} and using $1-d_{t,k}=\snr_{t,k}/(1+\snr_{t,k})$ from \eqref{eq:modewise_difficulty},
\begin{equation*}
    \xp\bigl[m_{t,k}^{2}\mid t\bigr]
    =
    p_k\bigl(1-d_{t,k}\bigr)+\scDelta_{t,k}
    =
    p_k\frac{\snr_{t,k}}{1+\snr_{t,k}}+\scDelta_{t,k} . \qedhere
\end{equation*}
\end{proof}

\section{Tangent-Kernel Dynamics of the GNN Denoiser}
\label{app:gnn_ntk}

Let $\scP$ be the number of parameters of $f_{\Theta}$. Under the gradient flow \eqref{eq:gradient_flow}, the GNN moves through the Jacobian of its output with respect to the parameters and through the tangent kernel,
\begin{equation}
\begin{gathered}
    \mtJ_{\Theta}(\vcx,t)
    =
    \nabla_{\Theta}f_{\Theta}(\vcx,\mtL,t)
    \in\fdR^{n\times\scP},
    \\
    \mtK_{\Theta,t}(\vcx,\vcx')
    =
    \mtJ_{\Theta}(\vcx,t)\,
    \mtJ_{\Theta}(\vcx',t)^{\Tr}
    \in\fdR^{n\times n} .
\end{gathered}
\label{eq:gnn_tangent_kernel}
\end{equation}
The kernel records how far a gradient step taken at $\vcx'$ moves the output at $\vcx$. Recall that $\pi_t$ is the law of the corrupted signal $\vcx_t$, and let $\vcr_{\sctau}=f_{\Theta(\sctau)}(\cdot,\mtL,t)-\fnm_t$ be the residual with respect to the posterior mean at optimization time $\sctau$. Its squared norm under $\pi_t$ is the excess error $\scOmega_{\Theta(\sctau),t}$ in \eqref{eq:gnn_risk_decomposition}. We write $\mtJ_{\sctau}(\vcx)=\mtJ_{\Theta(\sctau)}(\vcx,t)$, $\mtK_{\sctau}=\mtK_{\Theta(\sctau),t}$ and $\xp'[\,\cdot\,]=\xp_{\vcx'\sim\pi_t}[\,\cdot\,]$. Differentiating $\stL_t(f_{\Theta})$ under the expectation and conditioning on $\vcx_t=\vcx'$, with $\xp[\vcx_0\mid\vcx_t=\vcx',t]=\fnm_t(\vcx')$, gives
\begin{equation}
\begin{aligned}
    \nabla_{\Theta}\stL_t(f_{\Theta})
    &=
    -2\,
    \xp\bigl[
        \mtJ_{\sctau}(\vcx_t)^{\Tr}
        \bigl(
            \vcx_0-f_{\Theta(\sctau)}(\vcx_t,\mtL,t)
        \bigr)
        \,\big|\,t
    \bigr]
    \\
    &=
    2\,
    \xp'\bigl[
        \mtJ_{\sctau}(\vcx')^{\Tr}\vcr_{\sctau}(\vcx')
    \bigr].
\end{aligned}
    \label{eq:app_param_gradient}
\end{equation}
The posterior mean $\fnm_t$ does not depend on $\sctau$. The chain rule with \eqref{eq:gradient_flow} and \eqref{eq:gnn_tangent_kernel} gives the residual flow
\begin{equation}
    \frac{\partial\vcr_{\sctau}(\vcx)}{\partial\sctau}
    =
    -\eta\,\mtJ_{\sctau}(\vcx)\nabla_{\Theta}\stL_t(f_{\Theta})
    =
    -2\eta\,
    \xp'\bigl[
        \mtK_{\sctau}(\vcx,\vcx')\vcr_{\sctau}(\vcx')
    \bigr].
    \label{eq:gnn_ntk_dynamics}
\end{equation}
Resolving the residual and the kernel in the graph Fourier basis, $\sctr_{\sctau,k}(\vcx)=[\mtV^{\Tr}\vcr_{\sctau}(\vcx)]_k$ and $\sctK_{\sctau,jk}(\vcx,\vcx')=\vcv_j^{\Tr}\mtK_{\sctau}(\vcx,\vcx')\vcv_k$, and inserting $\mtV\mtV^{\Tr}=\mtI$ before $\vcr_{\sctau}(\vcx')$ in \eqref{eq:gnn_ntk_dynamics} gives the frequency-wise form
\begin{equation}
    \frac{\partial\sctr_{\sctau,j}(\vcx)}{\partial\sctau}
    =
    -2\eta
    \sum_{k=0}^{n-1}
    \xp'\bigl[
        \sctK_{\sctau,jk}(\vcx,\vcx')\,
        \sctr_{\sctau,k}(\vcx')
    \bigr] .
    \label{eq:gnn_modal_ntk_dynamics}
\end{equation}
The off-diagonal entries $\sctK_{\sctau,jk}$ with $j\neq k$ are what couples the graph frequencies. For the spectral graph filter \eqref{eq:spectral_filter_denoiser}
with $\Theta=\vch_t$, the kernel in the graph Fourier basis is
$\mtV^{\Tr}\mtK_{\sctau}(\vcx,\vcx')\mtV
=\diag(\vctx)\diag(\vctx')$.
It is diagonal, so \eqref{eq:gnn_modal_ntk_dynamics} reduces to
\eqref{eq:spectral_gradient_flow_mode}. The linear operations in \eqref{eq:message_passing_gnn} act separately across graph Fourier modes, but shared parameters can still couple modes in the tangent kernel. For example, the linear model $f_\theta(\vcx)=\theta\vcx$ has $\sctK_{\sctau,jk}(\vcx,\vcx')=\sctx_j\sctx'_k$, which is generally nonzero for $j\neq k$.

For the GNN the kernel moves with $\Theta(\sctau)$, and \eqref{eq:gnn_ntk_dynamics} is nonlinear in the residual. Freezing the kernel at $\mtK_{t}$ is the standard tangent-kernel linearization~\cite{jacot2018neural,du2019graph,liu2020linearity}, which we use only to identify which quantities depend on the diffusion rate. Under this linearization, the flow is generated by the linear integral operator
\begin{equation}
    \bigl(\stT_{t}\fnu\bigr)(\vcx)
    =
    \xp'\bigl[
        \mtK_{t}(\vcx,\vcx')\,\fnu(\vcx')
    \bigr] ,
    \quad
    \fnu\in L^{2}(\pi_t;\fdR^{n}),
    \label{eq:gntk_operator}
\end{equation}
which plays the role of the Hessian in \eqref{eq:spectral_loss_quadratic}. The flow becomes linear and solvable,
\begin{equation}
    \frac{\partial\vcr_{\sctau}}{\partial\sctau}
    =
    -2\eta\,\stT_{t}\vcr_{\sctau},
    \qquad
    \vcr_{\sctau}
    =
    e^{-2\eta\stT_{t}\sctau}\vcr_{0} ,
    \label{eq:gntk_linear_flow}
\end{equation}
whose exponential rates are given by the eigenvalues of $\stT_{t}$. For a fixed kernel function and initialization, the flow \eqref{eq:gntk_linear_flow} depends on the diffusion rate only through the law $\pi_t$ and the posterior mean $\fnm_t$, which at a fixed time are determined by the data law, $a_{t,k}$ and $\scnu_{t,k}$.

\section{Proofs for Mode-Adaptive SNR Scheduling}
\label{app:mas_snr_sampling_theory}

\begin{proof}[\textbf{Proof of Theorem~\ref{thm:mass_forward_channel}}]
Since $\mtC_t$ and $\mtL_{\scgamma}$ share the eigenbasis $\mtV$, projecting the forward SDE \eqref{eq:mass_forward_sde} onto $\vcv_k$ gives the Ornstein--Uhlenbeck equation \eqref{eq:gad_spectral_forward_mode} with $c_t$ replaced by $c_{t,k}$. Its solution is \eqref{eq:gad_channel_view} with attenuation and noise variance
\begin{equation}
    a_{t,k}
    =
    \exp\!\left(
        -\sclambda_{\scgamma,k}\scbc_{t,k}
    \right),
    \qquad
    \scnu_{t,k}
    =
    q_k\left(1-a_{t,k}^{2}\right),
    \label{eq:mas_mode_channel}
\end{equation}
and Definition~\ref{def:modewise_snr} gives
\begin{equation}
    \snr_{t,k}
    =
    \frac{a_{t,k}^{2}p_k}{q_k\bigl(1-a_{t,k}^{2}\bigr)}
    =
    g_k\,\frac{a_{t,k}^{2}}{1-a_{t,k}^{2}} .
    \label{eq:app_snr_attenuation_form}
\end{equation}
Since $p_k>0$, the power ratio $g_k>0$, and $s\mapsto g_ks/(1-s)$ is a strictly increasing bijection from $[0,1)$ onto $[0,\infty)$. The constraint \eqref{eq:mas_snr_constraint} has the unique solution $a_{t,k}^{2}=r_t/(r_t+g_k)$, which is the attenuation in \eqref{eq:mas_snr_channel}. Inverting $a_{t,k}^{2}=e^{-2\sclambda_{\scgamma,k}\scbc_{t,k}}$ gives \eqref{eq:mas_snr_cumulative_schedule} and carries the uniqueness to $\scbc_{t,k}$. The integrated rate is nonnegative because $g_k/r_t>0$. It is nondecreasing in $t$ because $r_t$ decreases, and it tends to zero as $r_t\to\infty$. Differentiating $\scbc_{t,k}=(2\sclambda_{\scgamma,k})^{-1}[\log(r_t+g_k)-\log r_t]$ in $t$ gives the diffusion rate \eqref{eq:mas_snr_rate_schedule}, which is nonnegative because $d\log r_t/dt\leq0$.
\end{proof}

\begin{proof}[\textbf{Proof of Theorem~\ref{thm:mas_snr_stiffnessRe}}]
We first prove the general bound \eqref{eq:mas_stiffness_bound} and then make the log-SNR slope $M$ explicit.

\smallskip
\noindent\emph{General bound.}
By \eqref{eq:mas_snr_channel}, $1-a_{t,k}^{2}=g_k/(r_t+g_k)$. Together with the diffusion rate \eqref{eq:mas_snr_rate_schedule}, the reverse-time gain in \eqref{eq:generic_reverse_drift} follows
\begin{equation}
    \frac{c_{t,k}\sclambda_{\scgamma,k}}{1-a_{t,k}^{2}}
    =
    -\frac{1}{2}\frac{d}{dt}\log r_t ,
    \label{eq:app_mass_rate_factor}
\end{equation}
which does not depend on $k$. Substituting \eqref{eq:app_mass_rate_factor} into \eqref{eq:generic_reverse_drift}, stacking over $k$ and differentiating in $\vctx$ gives the Jacobian of the MASS drift,
\begin{equation*}
\begin{aligned}
    \nabla_{\vctx}\vctb_t(\vctx)
    ={}&
    -\frac{1}{2}\frac{d}{dt}\log r_t
    \\
    &\times
    \left[
        \mtI+\diag(\vca_t)^{2}
        -
        2\diag(\vca_t)\nabla_{\vctx}\tilde{f}_{\Theta}(\vctx,t)
    \right].
\end{aligned}
\end{equation*}
Since $a_{t,k}^{2}\in(0,1)$, the first term in the bracket has norm $1+\max_ka_{t,k}^{2}\leq2$. Assumption~\ref{ass:local_bounded_jacobian} bounds the norm of the second term by $2B$ for all $\vctx\in\stR_t$. The triangle inequality then gives
\begin{equation*}
    \bigl\|\nabla_{\vctx}\vctb_t(\vctx)\bigr\|_2
    \leq
    \left|\frac{d}{dt}\log r_t\right|(1+B)
    \leq
    M(1+B) .
\end{equation*}
Taking the suprema in \eqref{eq:reverse_stiffness_proxy} gives \eqref{eq:mas_stiffness_bound}.

\smallskip
\noindent\emph{Explicit log-SNR slope.}
Let $\sclambda_{\scgamma,\max}=\sclambda_{\scgamma,n-1}$ be the largest shifted eigenvalue. The smallest one is $\sclambda_{\scgamma,0}=\scgamma$, hence $\cond(\mtL_{\scgamma})=\sclambda_{\scgamma,\max}/\scgamma$. Since $g_k=\sclambda_{\scgamma,k}p_k/\scsigma^{2}$ and $\scgamma\leq\sclambda_{\scgamma,k}\leq\sclambda_{\scgamma,\max}$, the extreme power ratios satisfy $g_{\max}\leq\sclambda_{\scgamma,\max}\max_kp_k/\scsigma^{2}$ and $g_{\min}\geq\scgamma\min_kp_k/\scsigma^{2}$. The condition $\max_kp_k\leq\kappa\min_kp_k$ then gives
\begin{equation}
    \frac{g_{\max}}{g_{\min}}
    \leq
    \kappa\,\cond(\mtL_{\scgamma}) .
    \label{eq:app_power_ratio_spread}
\end{equation}
For the linear ramp $u_t=(t-t_0)/(T-t_0)$, the target SNR \eqref{eq:mas_snr_log_trajectory} has the constant slope $d\log r_t/dt=-\log(r_{\max}/r_{\min})/(T-t_0)$, hence $M=\log(r_{\max}/r_{\min})/(T-t_0)$. With the span \eqref{eq:mass_dynamic_range} at equality and $\scdelta_0=\scdelta$,
\begin{equation*}
    M
    =
    \frac{1}{T-t_0}
    \left[
        2\log\frac{1-\scdelta}{\scdelta}
        +\log\frac{g_{\max}}{g_{\min}}
    \right].
\end{equation*}
Substituting \eqref{eq:app_power_ratio_spread} into this expression, and the result into \eqref{eq:mas_stiffness_bound}, gives \eqref{eq:mass_rate_explicit}.
\end{proof}

To conclude this appendix, we provide the proposition that shows the bound in \eqref{eq:reverse_stiffness_proxy}.
\begin{proposition}[Stiffness under a shared diffusion rate]
\label{prop:scalar_stiffness}
Let Assumption~\ref{ass:graph_wss} hold, let the diffusion rate $c_t$ be shared and nondecreasing, and let the horizon be $\scdelta$-complete with $\snr_{T,n-1}\leq 1/3$. Then, for the optimal linear denoiser \eqref{eq:optimal_modewise_gain}, the reverse-time stiffness \eqref{eq:reverse_stiffness_proxy} satisfies
\begin{equation}
    L_S^{\mathrm{shared}}
    \geq
    \frac{\log(1/\scdelta)}{4T}\,
    \cond(\mtL_{\scgamma}) .
    \label{eq:scalar_stiffness_cond_bound}
\end{equation}
\end{proposition}

\begin{proof}[\textbf{Proof of Proposition~\ref{prop:scalar_stiffness}}]
At the highest graph frequency, the assumption gives $\snr_{T,n-1}\leq 1/3$.
By \eqref{eq:optimal_modewise_gain}, $2a_{T,n-1}h_{T,n-1}^{\star}=2\snr_{T,n-1}/(1+\snr_{T,n-1})$, which increases with the SNR and is therefore at most $\tfrac12$. The numerator of the local stiffness $\sczeta_{T,n-1}$ at time $T$ is then at least $\tfrac12$, and its denominator $1-a_{T,n-1}^{2}$ is at most one. Since $L_S^{\mathrm{shared}}=\sup_t\max_k|\sczeta_{t,k}|$ for a spectral graph filter,
\begin{equation}
    L_S^{\mathrm{shared}}
    \geq
    |\sczeta_{T,n-1}|
    \geq
    \frac12\,c_T\sclambda_{\scgamma,\max} .
    \label{eq:scalar_stiffness_lower_bound}
\end{equation}
It remains to bound the terminal diffusion rate $c_T$. A nondecreasing diffusion rate gives $\scbc_T=\int_0^Tc_s\,ds\leq Tc_T$, and $\scdelta$-completeness gives $\scbc_T\geq\log(1/\scdelta)/(2\scgamma)$. Hence $c_T\geq\log(1/\scdelta)/(2\scgamma T)$. Substituting this bound into \eqref{eq:scalar_stiffness_lower_bound} and using $\sclambda_{\scgamma,\max}/\scgamma=\cond(\mtL_{\scgamma})$ gives \eqref{eq:scalar_stiffness_cond_bound}.
\end{proof}

\section{Proofs of the Lemmas}
\label{app:lemma_proofs}

\begin{proof}[\textbf{Proof of Lemma~\ref{lem:optimal_spectral_response}}]
Under Assumption~\ref{ass:graph_wss}, the clean covariance is $\mtP=\mtV\diag(p_0,\ldots,p_{n-1})\mtV^{\Tr}$. The attenuation $\mtH_t=\exp(-\scbc_t\mtL_{\scgamma})$ and the covariance $\mtSigma_t=\scsigma^{2}(\mtI-\mtH_t^{2})\mtL_{\scgamma}^{-1}$ of the transition \eqref{eq:gad_forward_marginal} are functions of $\mtL_{\scgamma}$ and share the eigenbasis $\mtV$ with $\mtP$. Let $\mtR_t=\xp[\vcx_t\vcx_t^{\Tr}\mid t]$ be the second-moment matrix of the corrupted signal and $\mtM_t=\xp[\vcx_0\vcx_t^{\Tr}\mid t]$ the cross-moment matrix between the clean and the corrupted signal. The forward noise in \eqref{eq:gad_forward_marginal} is zero mean and independent of $\vcx_0$, hence
\begin{equation}
    \mtM_t=\mtP\mtH_t,
    \qquad
    \mtR_t=\mtH_t\mtP\mtH_t+\mtSigma_t .
    \label{eq:app_linear_moments}
\end{equation}
Both matrices are diagonal in $\mtV$. The $k$th diagonal entry of $\mtM_t$ is $a_{t,k}p_k$ and that of $\mtR_t$ is the corrupted power $\scrho_{t,k}$. The fixed-time error of a linear denoiser expands as~\cite{kay1993fundamentals}
\begin{equation}
    \stL_t(f_{\mtW})
    =
    \trace(\mtP)
    -2\trace(\mtW\mtM_t^{\Tr})
    +\trace(\mtW\mtR_t\mtW^{\Tr}) .
    \label{eq:app_linear_risk}
\end{equation}
Since $\mtR_t\succ\mtZeros$ by \eqref{eq:app_channel_range}, the loss is strictly convex in $\mtW$. Its gradient
\begin{equation}
    \nabla_{\mtW}\stL_t(f_{\mtW})
    =
    -2\mtM_t+2\mtW\mtR_t
    =
    2\bigl(\mtW-\mtM_t\mtR_t^{-1}\bigr)\mtR_t
    \label{eq:app_linear_gradient}
\end{equation}
vanishes only at $\mtW^{\star}=\mtM_t\mtR_t^{-1}$. In graph Fourier coordinates,
$\mtV^{\Tr}\mtW^{\star}\mtV=\diag(\vch_t^{\star})$, with $h_{t,k}^{\star}=a_{t,k}p_k/\scrho_{t,k}$, which is the denoiser \eqref{eq:spectral_filter_denoiser} with the first form of \eqref{eq:optimal_modewise_gain}. Under Assumption~\ref{ass:simple_spectrum} the eigenvalues are distinct, and the assignment $\sclambda_{\scgamma,k}\mapsto h_{t,k}^{\star}$ defines a scalar function of the eigenvalue, which makes $\mtW^{\star}$ a spectral graph filter \eqref{eq:spectral_graph_filter}. The second form of \eqref{eq:optimal_modewise_gain} follows from $\scrho_{t,k}=a_{t,k}^{2}p_k+\scnu_{t,k}$ and $\snr_{t,k}=a_{t,k}^{2}p_k/\scnu_{t,k}$.
\end{proof}

\begin{proof}[\textbf{Proof of Lemma~\ref{lem:informative_window}}]
Since $p_k>0$, the power ratio $g_k=\sclambda_{\scgamma,k}\,p_k/\scsigma^{2}$ is positive. Since $\sclambda_{\scgamma,k}>0$, the map
\begin{equation}
    \varphi_k(\scbc)
    =
    e^{2\sclambda_{\scgamma,k}\scbc}-1 ,
    \qquad
    \varphi_k^{-1}(u)
    =
    \frac{1}{2\sclambda_{\scgamma,k}}\log(1+u),
    \label{eq:app_exponential_inverse}
\end{equation}
is a continuous and strictly increasing bijection of $[0,\infty)$ onto itself, and $\snr_k(\scbc)=g_k/\varphi_k(\scbc)$ by \eqref{eq:modewise_snr_definition}. Hence
\begin{equation}
    \sceps\leq\snr_k(\scbc)\leq\sceps^{-1}
    \iff
    g_k\sceps\leq\varphi_k(\scbc)\leq g_k\sceps^{-1} ,
    \label{eq:app_informative_equivalence}
\end{equation}
and applying $\varphi_k^{-1}$ to both sides gives \eqref{eq:app_informative_interval}. Its length is the equality in \eqref{eq:informative_window}. For $\sceps\in(0,1)$ we have $\sceps^{2}(1+g_k\sceps^{-1})=\sceps^{2}+g_k\sceps\leq1+g_k\sceps$, hence the argument of the logarithm is at most $\sceps^{-2}$, which gives the bound.
\end{proof}

\begin{proof}[\textbf{Proof of Lemma~\ref{lem:bayes_orthogonality}}]
Since $\mtV$ is deterministic, $m_{t,k}=\vcv_k^{\Tr}\xp[\vcx_0\mid\vcx_t,t]=\xp[\sctx_{0,k}\mid\vcx_t,t]$. Conditioning first on $\vcx_t$ lets us take $\scz$ outside the inner expectation. The conditional mean of the residual is zero, so
\begin{equation*}
    \xp[(\sctx_{0,k}-m_{t,k})\,\scz\mid t]
    =
    \xp\bigl[
        \xp[\sctx_{0,k}-m_{t,k}\mid\vcx_t,t]\,\scz
        \mid t
    \bigr]
    =0. \qedhere
\end{equation*}
\end{proof}

\end{document}